%% file: main.tex
\documentclass{article}
\input{preamble/setup.tex}

\input{preamble/title.tex}
\iclrfinalcopy
\usepackage{placeins}
\begin{document}
\raggedbottom
\maketitle
\lhead{Preprint}
\vspace{-22pt}

\input{sections/abstract.tex}

\input{sections/introduction.tex}

\input{sections/method.tex}

\input{sections/experiments.tex}

\input{sections/conclusion.tex}

\clearpage
\bibliographystyle{iclr2027_conference}
\bibliography{bibliography/references,bibliography/additional}

\newpage
\appendix
\input{appendices/appendix.tex}

\end{document}

%% file: preamble/setup.tex
\ifdefined\pdfminorversion\pdfminorversion=7\fi
\usepackage{iclr2027_conference,times}

\input{preamble/math_commands.tex}

\usepackage{wrapfig}
\usepackage[utf8]{inputenc}
\usepackage[T1]{fontenc}
\usepackage{amsmath,amssymb}
\usepackage{booktabs}
\usepackage{graphicx}
\graphicspath{{figures/}}
\usepackage{subcaption}
\usepackage{multirow}
\usepackage{tabularx}
\usepackage{float}
\usepackage{enumitem}
\usepackage{etoolbox}
\usepackage{pifont}
\usepackage[most]{tcolorbox}
\usepackage{xcolor}
\definecolor{MyBlue}{RGB}{0,172,238}
\definecolor{MyPink}{RGB}{237,0,139}
\usepackage{colortbl}
\usepackage{microtype}
\usepackage{fancyvrb}
\usepackage[colorlinks=true,
            linkcolor=red,
            citecolor=MyBlue,
            urlcolor=MyPink]{hyperref}
\usepackage{url}

\newcommand{\modelname}{RotaFuse}

\newcommand{\DINO}{\textnormal{DINOv3}}
\newcommand{\SIGLIP}{\textnormal{SigLIP}}
\newcommand{\GRANULON}{\textnormal{Granulon}}

\definecolor{GranulonHeader}{HTML}{DED9E6}
\definecolor{GranulonStripe}{HTML}{F2F2F2}
\definecolor{GranulonOurs}{HTML}{FFFCCC}
\definecolor{CaseGrounded}{HTML}{18843F}
\definecolor{CaseError}{HTML}{C4252C}
\definecolor{CasePAIQ}{HTML}{315D9B}
\definecolor{CaseRule}{HTML}{B8B8B8}
\definecolor{CaseOutputBG}{HTML}{F7F7F6}

\usepackage{amsthm}
\usepackage{needspace}
\usepackage{listings,upquote}
\tcbuselibrary{skins,breakable,listings,theorems}

\input{preamble/method_style.tex}
\input{preamble/appendix_style.tex}
\input{preamble/table_style.tex}

%% file: preamble/math_commands.tex
\usepackage{amsmath,amsfonts,bm}

\def\eqref#1{equation~\ref{#1}}

\def\1{\bm{1}}

\def\vd{{\bm{d}}}

\def\vr{{\bm{r}}}
\def\vs{{\bm{s}}}

\def\vz{{\bm{z}}}

\def\mA{{\bm{A}}}

\def\mC{{\bm{C}}}
\def\mD{{\bm{D}}}

\def\mI{{\bm{I}}}

\def\mQ{{\bm{Q}}}
\def\mR{{\bm{R}}}
\def\mS{{\bm{S}}}

\def\mX{{\bm{X}}}

\def\mZ{{\bm{Z}}}

\DeclareMathAlphabet{\mathsfit}{\encodingdefault}{\sfdefault}{m}{sl}
\SetMathAlphabet{\mathsfit}{bold}{\encodingdefault}{\sfdefault}{bx}{n}



%% file: preamble/method_style.tex
\providecommand{\vd}{{\bm d}}
\providecommand{\vr}{{\bm r}}
\providecommand{\vs}{{\bm s}}
\providecommand{\vz}{{\bm z}}
\providecommand{\mA}{{\bm A}}
\providecommand{\mC}{{\bm C}}
\providecommand{\mD}{{\bm D}}
\providecommand{\mI}{{\bm I}}
\providecommand{\mQ}{{\bm Q}}
\providecommand{\mR}{{\bm R}}
\providecommand{\mS}{{\bm S}}
\providecommand{\mX}{{\bm X}}
\providecommand{\mZ}{{\bm Z}}
\providecommand{\PAIQeqref}[1]{\textup{(\ref{#1})}}

\makeatletter
\@ifundefined{lemma}{%
  \edef\PAIQsavedtheoremstyle{\the\thm@style}%
  \theoremstyle{plain}%
  \@ifundefined{c@theorem}{\newtheorem{lemma}{Lemma}}%
    {\newtheorem{lemma}[theorem]{Lemma}}%
  \expandafter\theoremstyle\expandafter{\PAIQsavedtheoremstyle}%
}{}
\makeatother

\definecolor{PAIQMatch}{HTML}{1D4ED8}
\definecolor{PAIQRotate}{HTML}{E66100}
\definecolor{PAIQTheoryFill}{HTML}{F4F5FC}
\newif\ifPAIQcolor
\PAIQcolortrue
\newcommand{\PAIQmatch}[1]{{\ifPAIQcolor\color{PAIQMatch}\fi#1}}
\newcommand{\PAIQrotate}[1]{{\ifPAIQcolor\color{PAIQRotate}\fi#1}}

\newenvironment{paiqtheorybox}{%
  \ifPAIQcolor
    \colorlet{PAIQBoxFill}{PAIQTheoryFill}%
  \else
    \colorlet{PAIQBoxFill}{black!3}%
  \fi
  \begin{tcolorbox}[enhanced,breakable,
    sharp corners,arc=0pt,outer arc=0pt,
    frame hidden,boxrule=0pt,
    colback=PAIQBoxFill,coltext=black,
    boxsep=0pt,left=8pt,right=8pt,top=3pt,bottom=3pt,
    before skip=7pt,after skip=7pt]
}{\end{tcolorbox}}

%% file: preamble/appendix_style.tex
\definecolor{APXInk}{HTML}{17263D}
\definecolor{APXBlue}{HTML}{254BC7}
\definecolor{APXTeal}{HTML}{007A74}
\definecolor{APXAmber}{HTML}{B65A14}
\definecolor{APXMuted}{HTML}{526174}
\definecolor{APXRule}{HTML}{CED6E2}
\definecolor{APXBlueTint}{HTML}{EFF3FF}
\definecolor{APXBlueHead}{HTML}{E2E9FF}
\definecolor{APXTealTint}{HTML}{EDF7F4}
\definecolor{APXTealHead}{HTML}{D9EEE8}
\definecolor{APXAmberTint}{HTML}{FFF5E8}
\definecolor{APXPaper}{HTML}{F6F8FB}

\newif\ifAPXMonochrome
\APXMonochromefalse
\AtBeginDocument{%
  \ifAPXMonochrome
    \colorlet{APXInk}{black!90}\colorlet{APXBlue}{black!80}%
    \colorlet{APXTeal}{black!70}\colorlet{APXAmber}{black!70}%
    \colorlet{APXMuted}{black!60}\colorlet{APXRule}{black!25}%
    \colorlet{APXBlueTint}{black!4}\colorlet{APXBlueHead}{black!9}%
    \colorlet{APXTealTint}{black!4}\colorlet{APXTealHead}{black!9}%
    \colorlet{APXAmberTint}{black!4}\colorlet{APXPaper}{black!3}%
  \fi
}

\newif\ifAPXGalleryLayout
\APXGalleryLayoutfalse

\tcbset{
  apx/base/.style={
    enhanced jigsaw,breakable,sharp corners,
    boxrule=0pt,boxsep=0pt,
    left=10pt,right=10pt,top=8pt,bottom=8pt,
    before skip=10pt,after skip=10pt,
    colback=white,colframe=APXRule,coltext=black,
    fontupper=\normalsize,fontlower=\normalsize,
    fonttitle=\normalsize\bfseries,coltitle=APXInk,
    toptitle=6pt,bottomtitle=6pt,
    pad at break*=2mm,lines before break=3,
    before upper={\setlength{\parindent}{0pt}},
  },
  apx/theory/.style={
    apx/base,colback=APXBlueTint,colbacktitle=APXBlueHead,
    coltitle=APXBlue,
    theorem style=standard,separator sign={\enspace\textbar\enspace},
    description font=\normalfont\bfseries,
    fontupper=\normalsize\normalfont,
  },
  apx/theory white/.style={
    colback=white,colbacktitle=white,
  },
  apx/proof/.style={
    apx/base,colback=white,colbacktitle=white,
    coltitle=APXInk,borderline west={1.1pt}{0pt}{APXRule},
    top=4pt,bottom=4pt,toptitle=2pt,bottomtitle=5pt,
  },
  apx/note/.style={
    apx/base,colback=APXAmberTint,colbacktitle=APXAmberTint,
    coltitle=APXAmber,borderline west={2pt}{0pt}{APXAmber},
    toptitle=6pt,bottomtitle=1pt,
  },
  apx/code/.style={
    apx/base,colback=APXPaper,colbacktitle=APXInk,coltitle=white,
    left=32pt,right=10pt,top=8pt,bottom=8pt,
    lefttitle=10pt,righttitle=10pt,
  },
  apx/prompt/.style={
    apx/base,colback=APXTealTint,colbacktitle=APXTealHead,
    coltitle=APXTeal,
  },
  apx/case/.style={
    apx/base,colback=white,colbacktitle=APXPaper,
  },
}

\newtcbtheorem[number within=section]{APXTheorem}{Theorem}
  {apx/theory,title after break={Theorem~\thetcbcounter\enspace\textnormal{(continued)}}}{apxthm}
\newtcbtheorem[use counter from=APXTheorem]{APXLemma}{Lemma}
  {apx/theory,title after break={Lemma~\thetcbcounter\enspace\textnormal{(continued)}}}{apxlem}
\newtcbtheorem[use counter from=APXTheorem]{APXProposition}{Proposition}
  {apx/theory,title after break={Proposition~\thetcbcounter\enspace\textnormal{(continued)}}}{apxprop}
\newtcbtheorem[use counter from=APXTheorem]{APXDefinition}{Definition}
  {apx/theory,colback=APXPaper,colbacktitle=APXPaper,
   coltitle=APXInk,
   title after break={Definition~\thetcbcounter\enspace\textnormal{(continued)}}}{apxdef}
\newtcbtheorem[use counter from=APXTheorem]{APXAssumption}{Assumption}
  {apx/theory,colback=APXPaper,colbacktitle=APXPaper,
   coltitle=APXInk,
   title after break={Assumption~\thetcbcounter\enspace\textnormal{(continued)}}}{apxass}

\NewDocumentEnvironment{APXProof}{O{} m}{%
  \begin{tcolorbox}[apx/proof,title={#2},
    title after break={#2\enspace\textnormal{(continued)}},#1]%
  \pushQED{\qed}\normalfont\ignorespaces
}{%
  \popQED\end{tcolorbox}\ignorespacesafterend
}

\NewDocumentEnvironment{APXRestatement}{O{} m}{%
  \begin{tcolorbox}[apx/theory,title={#2},
    title after break={#2\enspace\textnormal{(continued)}},#1]
}{\end{tcolorbox}}
\NewDocumentEnvironment{APXNote}{O{} m}{%
  \begin{tcolorbox}[apx/note,title={#2},
    title after break={#2\enspace\textnormal{(continued)}},#1]
}{\end{tcolorbox}}

\lstdefinestyle{APXPython}{
  language=Python,
  basicstyle=\normalsize\fontencoding{T1}\selectfont\ttfamily\color{APXInk},
  keywordstyle=\color{APXBlue}\bfseries,
  commentstyle=\color{APXMuted}\itshape,
  stringstyle=\color{APXTeal},
  numbers=left,numberstyle=\normalsize\ttfamily\color{APXMuted},
  numbersep=9pt,stepnumber=1,
  columns=fullflexible,keepspaces=true,showstringspaces=false,
  showtabs=false,tabsize=4,upquote=true,
  breaklines=true,breakatwhitespace=false,breakautoindent=true,
  breakindent=1em,aboveskip=0pt,belowskip=0pt,
  mathescape=false,texcl=false,
}
\lstdefinestyle{APXLiteral}{
  style=APXPython,language={},numbers=none,
  keywordstyle={},commentstyle={},stringstyle={},
  basicstyle=\normalsize\fontencoding{T1}\selectfont\ttfamily\color{APXInk},
}
\newtcblisting[auto counter,number within=section]{APXCode}[3][]{
  apx/code,listing only,
  listing options={style=APXPython},
  title={Listing~\thetcbcounter\enspace\textbar\enspace #2},
  title after break={Listing~\thetcbcounter\enspace\textbar\enspace #2
    \enspace\textnormal{(continued)}},
  nameref={#2},label={#3},#1,
}
\newtcbinputlisting[use counter from=APXCode]{\APXCodeFile}[4][]{
  apx/code,listing only,listing options={style=APXPython},
  title={Listing~\thetcbcounter\enspace\textbar\enspace #2},
  title after break={Listing~\thetcbcounter\enspace\textbar\enspace #2
    \enspace\textnormal{(continued)}},
  nameref={#2},label={#3},listing file={#4},#1,
}

\newtcolorbox[auto counter,number within=section]{APXPrompt}[3][]{
  apx/prompt,
  title={Prompt~\thetcbcounter\enspace\textbar\enspace #2},
  title after break={Prompt~\thetcbcounter\enspace\textbar\enspace #2
    \enspace\textnormal{(continued)}},
  nameref={#2},label={#3},#1,
}
\newtcblisting[use counter from=APXPrompt]{APXRawPrompt}[3][]{
  apx/prompt,listing only,listing options={style=APXLiteral},
  title={Prompt~\thetcbcounter\enspace\textbar\enspace #2},
  title after break={Prompt~\thetcbcounter\enspace\textbar\enspace #2
    \enspace\textnormal{(continued)}},
  nameref={#2},label={#3},#1,
}

\NewDocumentCommand{\APXFigurePlaceholder}{O{32mm} m m}{%
  \begin{tcolorbox}[
    enhanced,sharp corners,boxrule=0.45pt,colframe=APXRule,
    colback=APXPaper,coltext=APXMuted,
    boxsep=0pt,left=6pt,right=6pt,top=5pt,bottom=5pt,
    height=#1,valign=center,halign=center,
    before skip=0pt,after skip=0pt,fontupper=\normalsize]
    {\color{APXInk}\bfseries #2}\par
    \vspace{4pt}#3
  \end{tcolorbox}%
}

\newtcolorbox[auto counter,number within=section]{APXCase}[3][]{
  apx/case,
  title={Case~\thetcbcounter\enspace\textbar\enspace #2},
  title after break={Case~\thetcbcounter\enspace\textbar\enspace #2
    \enspace\textnormal{(continued)}},
  nameref={#2},label={#3},#1,
}

\newcommand{\APXSetupCleveref}{%
    \crefname{tcb@cnt@APXTheorem}{statement}{statements}%
    \Crefname{tcb@cnt@APXTheorem}{Statement}{Statements}%
    \crefname{tcb@cnt@APXCode}{listing}{listings}%
    \Crefname{tcb@cnt@APXCode}{Listing}{Listings}%
    \crefname{tcb@cnt@APXPrompt}{prompt}{prompts}%
    \Crefname{tcb@cnt@APXPrompt}{Prompt}{Prompts}%
    \crefname{tcb@cnt@APXCase}{case}{cases}%
    \Crefname{tcb@cnt@APXCase}{Case}{Cases}%
}
\makeatletter
\@ifpackageloaded{cleveref}{\APXSetupCleveref}{}
\makeatother

\newcommand{\AppendixContentsLine}[2]{%
  \noindent
  \hyperref[#1]{%
    \makebox[\linewidth][l]{%
      \textbf{\underline{Appendix~\textcolor{red}{\ref*{#1}}}}.~%
      #2\nobreak\dotfill\nobreak\pageref*{#1}%
    }%
  }\par
}

\definecolor{PAIQCaseSupported}{HTML}{18794E}
\definecolor{PAIQCaseUnsupported}{HTML}{C0262D}
\newcommand{\PAIQcasegood}[1]{{\ifPAIQcolor\color{PAIQCaseSupported}\fi#1}}
\newcommand{\PAIQcasebad}[1]{{\ifPAIQcolor\color{PAIQCaseUnsupported}\fi#1}}
\newenvironment{paiqcaseoutput}[1]{%
  \begin{tcolorbox}[enhanced,breakable,sharp corners,frame hidden,boxrule=0pt,
    colback=#1,coltext=black,boxsep=0pt,left=7pt,right=7pt,top=5pt,bottom=5pt,
    before skip=5pt,after skip=5pt,fontupper=\footnotesize]
  \raggedright
}{\end{tcolorbox}}

%% file: preamble/table_style.tex
\definecolor{PAIQInk}{HTML}{25262B}
\definecolor{PAIQLavender}{HTML}{E5E1EC}
\definecolor{PAIQGray}{HTML}{ECEEF1}
\definecolor{PAIQYellow}{HTML}{FFF8D8}
\definecolor{PAIQGreen}{HTML}{147344}
\definecolor{PAIQRed}{HTML}{B6313C}

\newcommand{\PAIQAcc}{Acc.\,$\uparrow$}
\newcommand{\PAIQHall}{Hall.\,$\downarrow$}
\newcommand{\PAIQDelta}[3]{\hspace{0.35pt}\makebox[18.5pt][l]{\raisebox{-1.5pt}{%
  {\fontsize{5.5}{6.5}\selectfont\textcolor{#1}{\ensuremath{#2}#3}}}}}
\newcommand{\PAIQValue}[1]{\makebox[21pt][c]{#1}}
\newcommand{\PAIQGain}[4]{\PAIQValue{#1}\makebox[0pt][l]{\PAIQDelta{#2}{#3}{#4}}}
\newcommand{\PAIQOurs}[1]{\PAIQValue{#1}}
\newcommand{\PAIQBand}[1]{\rule[-4.4pt]{0pt}{14pt}\hspace{5pt}{\fontfamily{ptm}\bfseries\itshape #1}}
\newcommand{\PAIQSetup}{%
  \centering\fontsize{8}{9.4}\selectfont
  \setlength{\tabcolsep}{3pt}\renewcommand{\arraystretch}{1.13}%
  \setlength{\aboverulesep}{2pt}\setlength{\belowrulesep}{2pt}%
  \setlength{\heavyrulewidth}{0.8pt}\setlength{\lightrulewidth}{0.35pt}%
  \arrayrulecolor{PAIQInk}\color{PAIQInk}}

%% file: preamble/title.tex
\title{Patch-wise Semantic Enrichment via Residual Rotation for Vision-Language Models}

\makeatletter
\patchcmd{\@maketitle}
  {\rule{\z@}{24pt}\@author}
  {\rule{\z@}{10pt}\@author}
  {}{\PackageError{PAIQ}{Author spacing patch failed}{Check the title macro.}}
\makeatother

\newlength{\PAIQauthorwidth}
\newcommand{\PAIQauthorblock}[3]{%
  \begin{minipage}[t]{\PAIQauthorwidth}
    \raggedright\normalfont\fontsize{9}{9.5}\selectfont
    \microtypesetup{protrusion=false}
    \setlength{\parskip}{0pt}
    \textbf{\normalsize #1}\\
    #2\\
    #3
  \end{minipage}%
}
\author{%
\hspace*{-\tabcolsep}%
\begin{tabular}[t]{@{}l@{\hspace{8pt}}l@{\hspace{8pt}}l@{}}
\PAIQauthorblock{Pinze Ren\thanks{Equal contribution.}}
  {Tsinghua University}{rpz21@mails.tsinghua.edu.cn}
&
\PAIQauthorblock{Yuwei Zhang\footnotemark[1]}
  {Beijing University of\\Posts and Telecommunications}{2023213843@bupt.cn}
&
\PAIQauthorblock{Hao Chen\footnotemark[1]}
  {Tsinghua University}{chenhao26@mails.tsinghua.edu.cn}
\\[4pt]
\PAIQauthorblock{Linghao Meng}
  {National University of Singapore}{e1583364@u.nus.edu}
&
\PAIQauthorblock{Chang Li}
  {Tsinghua University}{lichang24@mails.tsinghua.edu.cn}
&
\PAIQauthorblock{Qiankun Li\thanks{Corresponding author.}}
  {Nanyang Technological University}{cs-qiankun.li@ntu.edu.sg}
\end{tabular}%
\hspace*{-\tabcolsep}%
}

\hypersetup{
  pdfauthor={Pinze Ren, Yuwei Zhang, Hao Chen, Linghao Meng, Chang Li, Qiankun Li},
  pdftitle={Patch-wise Semantic Enrichment via Residual Rotation for Vision-Language Models}
}

%% file: sections/abstract.tex
\begin{abstract}
Language-aligned and self-supervised visual encoders offer complementary semantic and spatial cues. Integrating these representations poses a geometric challenge: direct interpolation toward a shared aggregate contracts distances between local features. We introduce \textbf{\modelname{}}, a patch-wise fusion framework that separates cross-encoder feature aggregation from geometry-constrained residual updates. Using DINOv3 patch features as the spatial base, \modelname{} jointly allocates and aggregates SigLIP features at each base position. A shared, learnable orthogonal transform rotates each aggregate--base residual before it is scaled and added back, preserving residual geometry up to uniform scaling. For fixed base and aggregate features, we show that under shared aggregates, rotation adds a nonnegative term to the squared feature distance obtained by direct interpolation, and derive a rotation-aware separability bound for general aggregates. We further establish function-class equivalence to direct interpolation under unrestricted affine content projections and linear matching maps with the allocation rule and residual scale fixed, identifying explicit rotation as a geometry-constrained training parameterization. With both visual encoders and the language model frozen, \modelname{} learns a 196-token visual interface through answer generation supervision. Across five language backbones (2B--27B) and four datasets spanning image description and visual question answering, it improves judge-assessed correctness over both matched single-encoder interfaces in 19 of 20 settings and reduces judge-assessed hallucination severity in all 20. On Qwen3.5-2B and 9B, \modelname{} outperforms the strongest evaluated fusion or token-compression baselines by 2.94 correctness score points on average, and leads the included comparators in judge-independent multiple-choice evaluation at 2B.
\end{abstract}

%% file: sections/introduction.tex
\providecommand{\modelname}{RotaFuse}

\section{Introduction}
\label{sec:intro}

\begin{figure}[!h]
  \centering
  \includegraphics[width=\linewidth]{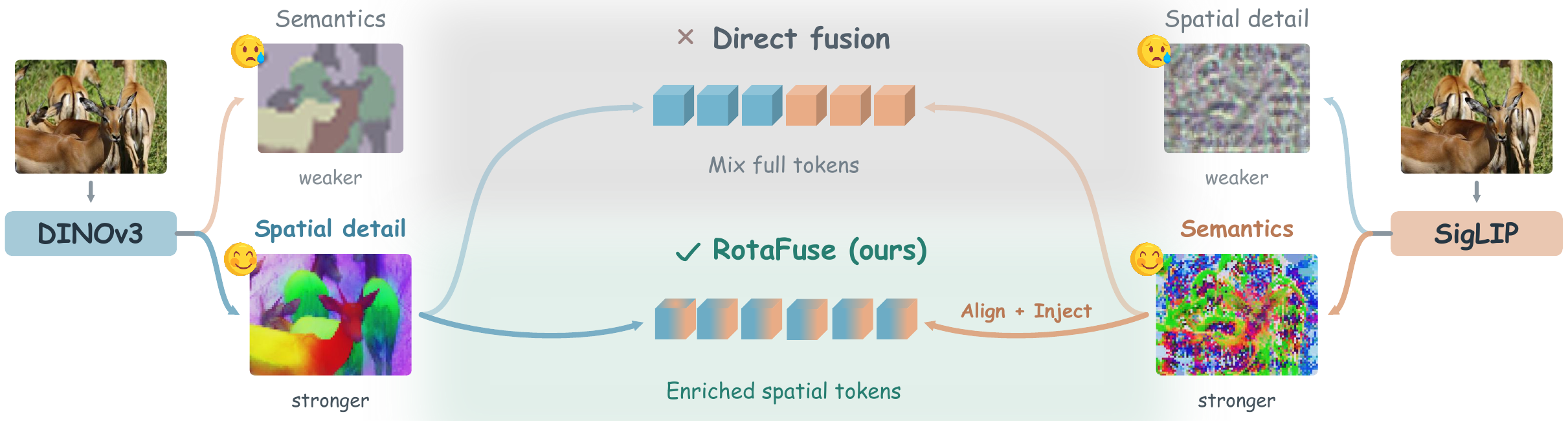}
  \caption{Conceptual illustration of complementary visual representations and the \modelname{} fusion principle. DINOv3 and SigLIP differ in their emphasis on local structure and language-aligned semantics. \modelname{} uses DINOv3 patch features as a spatial base, aggregates complementary SigLIP features through content matching, and incorporates them through orthogonal residual updates.}
  \label{fig:teaser}
\end{figure}

Visual understanding requires representations that capture both the semantics shared by related image regions and the local distinctions between them. Different pretraining objectives emphasize these properties to different degrees. Language-aligned encoders such as SigLIP~\citep{SigLIP} associate visual content with linguistic concepts, while self-supervised encoders such as DINOv2~\citep{DINOv2} and DINOv3~\citep{DINOv3} learn strong local correspondence and dense representations. Their complementary strengths motivate integrating both types of features into a single visual interface for vision-language models (Figure~\ref{fig:teaser}).

Existing multi-encoder interfaces bring heterogeneous features together through different alignment and aggregation mechanisms. COMM~\citep{COMM} and Eagle~\citep{Eagle} fuse features on a common grid, while BRAVE~\citep{BRAVE}, Cambrian-1~\citep{Cambrian1}, and CoME-VL~\citep{CoMEVL} aggregate features through queries or cross-encoder attention. Beyond cross-encoder correspondence, a key question is how fusion changes the geometry of local representations. Consider two patches on the same garment: they may draw on a common semantic feature while differing in texture, pattern, or boundary structure. Direct interpolation toward a shared aggregate contracts the distance between their base features. This observation motivates our central question: \emph{how can local visual representations share complementary semantics while retaining their separability?}

\begin{figure}[t]
  \centering
  \includegraphics[width=\linewidth]{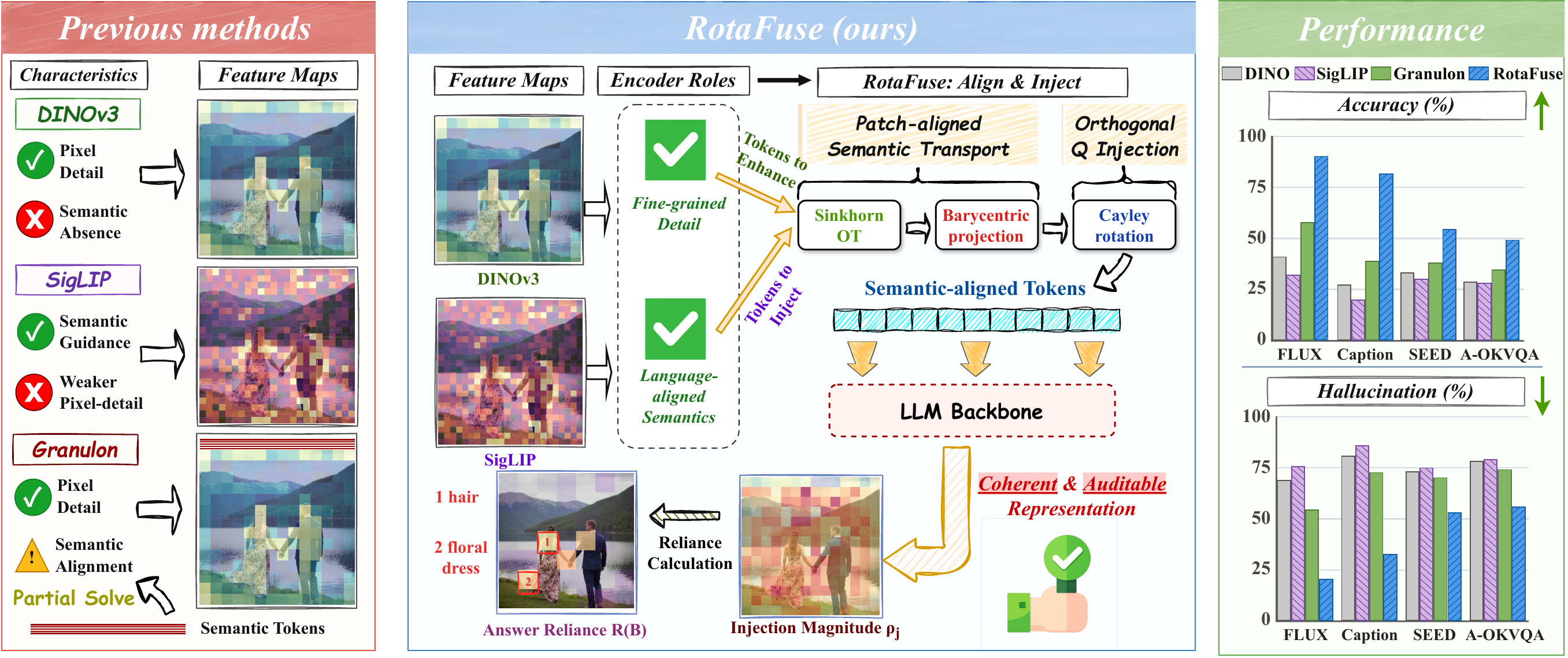}
  \caption{Overview of \modelname{}. Cross-encoder matching aggregates SigLIP features at DINOv3 patch positions. A shared orthogonal rotation transforms the aggregate--base residual before it is added back to the base, yielding 196 fused visual tokens. Post-training diagnostics characterize source allocation, relative update magnitude, and the sensitivity of a fixed answer to local residual removal. The performance panels report judge scores on a $[0,100]$ scale.}
  \label{fig:intro_overview}
\end{figure}

We introduce \textbf{\modelname{}}, a multi-encoder interface that fuses complementary visual features through residual rotation, separating feature aggregation from geometry-constrained representation updates (Figure~\ref{fig:intro_overview}). DINOv3 patch features define the spatial base, and SigLIP features supply complementary semantics. Content-based matching and finite-step row--column normalization jointly allocate SigLIP features across the base positions, yielding an aggregate for each patch. We subtract the base feature from its aggregate, rotate the resulting residual with a shared, learnable orthogonal matrix, and add the scaled residual back to the base. The fused sequence retains the 196 base positions. We learn the projections and fusion parameters through autoregressive answer supervision, with both visual encoders and the language model frozen.

Orthogonal rotation adapts the update directions while preserving residual norms and pairwise angles. For fixed projected bases and aggregates, we show that when two positions receive the same aggregate, the squared distance between their fused features equals that under direct interpolation plus a nonnegative rotation term. We extend this characterization to similar aggregates through a rotation-aware lower bound that gives a sufficient condition for local separability. In the full trainable interface, unrestricted affine content projections and linear matching maps allow the rotation to be absorbed into the preceding layers while preserving the matching weights. With the aggregation rule and residual scale fixed, the learned-rotation and direct-interpolation models therefore share the same function class. Explicit rotation thus defines a geometry-constrained training parameterization.

Across five language-backbone configurations spanning 2B to 27B parameters and four datasets covering image description and visual question answering, \modelname{} improves judge-assessed correctness over both matched single-encoder interfaces in 19 of 20 settings and reduces judge-assessed hallucination severity in all 20. With the Qwen3.5-2B and 9B backbones, it outperforms the strongest evaluated alternative among CoME-VL~\citep{CoMEVL}, MERV~\citep{MERV}, LLaVA-Mini~\citep{LLaVAMini}, and a Granulon-style comparator~\citep{Granulon} on every dataset, by 2.94 correctness score points on average. Under benchmark-aligned multiple-choice evaluation on the same VQA slices, the 2B interface also surpasses the strongest included comparators by 1.5 and 1.3 percentage points on SEED-Bench and A-OKVQA, respectively. Component ablations support both joint allocation and orthogonally constrained updates. At 2B, learning the residual rotation instead of fixing it to the identity improves judge-assessed correctness on these two VQA datasets by 21.24 and 16.58 points, respectively.

Our contributions are threefold:
\begin{itemize}
    \item \textbf{Geometry-constrained semantic enrichment.} We introduce a patch-wise fusion interface that couples joint cross-encoder feature allocation with learnable orthogonal residual updates, integrating complementary representations within the original base-token budget.
    \item \textbf{Local separability and parameterization.} We characterize the contribution of residual rotation to local feature separation under shared or similar aggregates and establish function-class equivalence to direct interpolation under unrestricted affine projections and linear matching maps.
    \item \textbf{Empirical validation across tasks and backbones.} We demonstrate gains in visual-language understanding across language backbones and scales, supported by judge-independent VQA evaluation and ablations that distinguish joint allocation, orthogonal constraints, and learned residual rotation.
\end{itemize}

%% file: sections/method.tex
\section{Method}
\label{sec:method}

\begin{figure}[t]
  \centering
  \includegraphics[width=\linewidth]{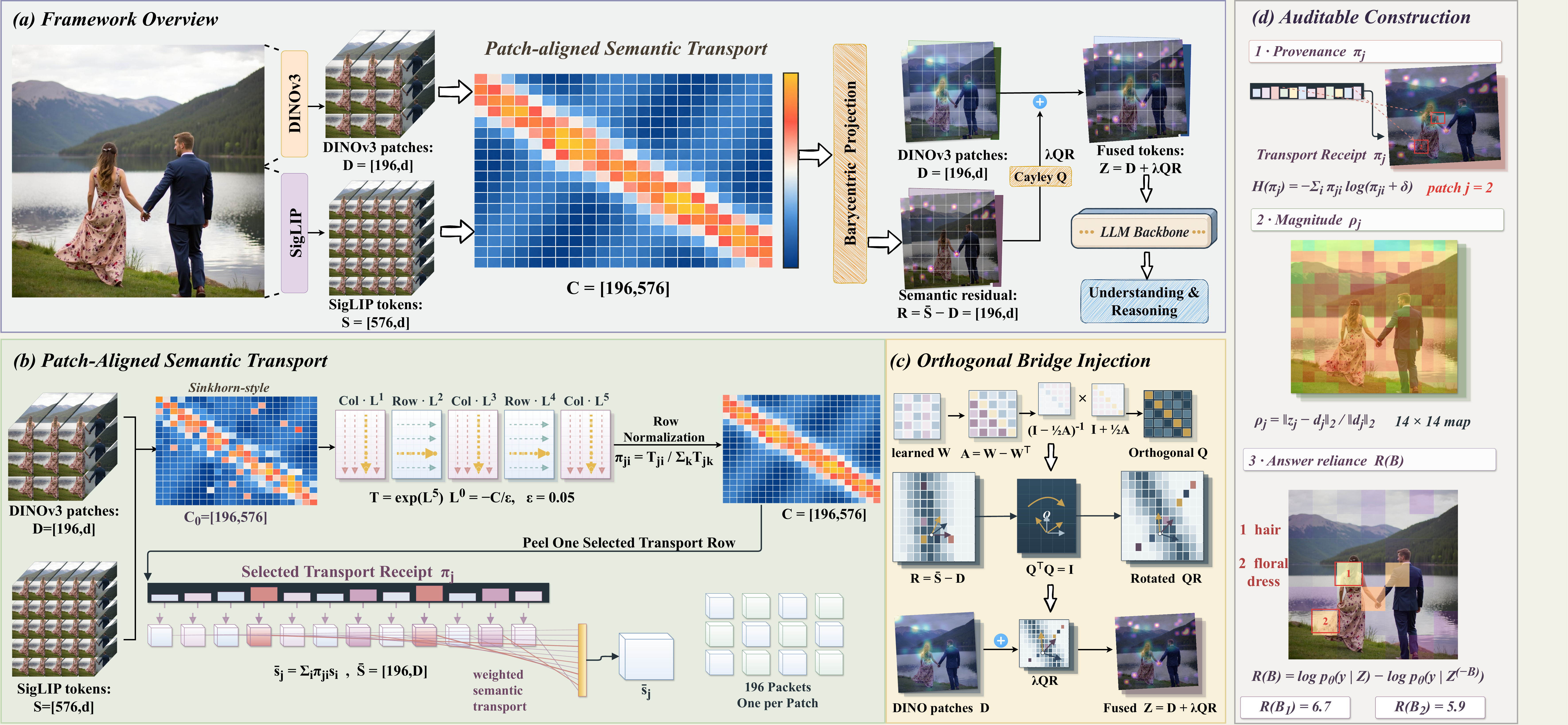}
  \caption[Overview of \modelname{}.]{\textbf{Overview of \modelname{}.}
  (a) The two frozen visual encoders supply complementary features that are fused into $196$ tokens for a frozen language model.
  (b) Content matching and finite-step row--column scaling allocate \SIGLIP{} features jointly across \DINO{} positions.
  (c) A shared orthogonal rotation transforms the difference between each aggregate and its base feature, yielding $\mZ=\mD+\lambda(\bar{\mS}-\mD)\mQ^\top$ for row-stacked tokens.
  (d) Post-training diagnostics characterize the aggregation weights, update magnitudes, and sensitivity of a fixed answer to local residual removal, the latter requiring additional teacher-forced evaluations (Appendix~\ref{app:diagnostics}).}
  \label{fig:method}
\end{figure}

\modelname{} is designed to enrich local visual representations with complementary semantics while retaining their ability to distinguish different image regions. As illustrated in Figure~\ref{fig:method}, we use \DINO{} features as a spatial base and separate semantic aggregation from representation updates: joint cross-encoder allocation gathers complementary features from \SIGLIP{} at each base position (Section~\ref{sec:semantic_transport}), and a geometry-constrained residual update learns how to incorporate them (Section~\ref{sec:orthogonal_injection}). We characterize the resulting local separability under shared aggregates (Section~\ref{sec:local_separability}) and train the visual interface through answer generation (Section~\ref{sec:visual_learning}).

\subsection{Joint cross-encoder feature allocation}
\label{sec:semantic_transport}
\label{sec:representation}

Given an image $I$, the frozen \DINO{} and \SIGLIP{} encoders produce contextualized patch features $\mX_d\in\mathbb R^{n\times1024}$ and $\mX_s\in\mathbb R^{m\times1024}$, with $n=196$ and $m=576$ corresponding to their $14\times14$ and $24\times24$ grids. We use the final patch outputs of both encoders and map them to the language-model dimension $d$ through independent affine projections:
\begin{equation}
\mD=\mX_dP_d^\top+\mathbf1b_d^\top,\qquad
\mS=\mX_sP_s^\top+\mathbf1b_s^\top.
\label{eq:content_project}
\end{equation}
Here $P_d,P_s\in\mathbb R^{d\times1024}$ and $b_d,b_s\in\mathbb R^d$ are trainable, and the projections remain affine (Appendix~\ref{app:implementation}). Feature matrices stack tokens as rows; individual features are column vectors. We denote the base feature at position $j$ by $\vd_j$ and the source feature at position $i$ by $\vs_i$.

The encoders construct representations on different grids, so we establish correspondence through feature content. Two trainable, bias-free maps $W_d,W_s\in\mathbb R^{d\times d}$ produce normalized queries and keys, from which we compute a cosine cost:
\begin{equation}
\begin{aligned}
\hat{\vd}_j&=\operatorname{norm}(W_d\vd_j),\qquad
\hat{\vs}_i=\operatorname{norm}(W_s\vs_i),\\
C_{ji}&=[1-\hat{\vd}_j^\top\hat{\vs}_i]_+.
\end{aligned}
\label{eq:cost}
\end{equation}
Here $\operatorname{norm}$ denotes $\ell_2$ normalization with near-zero protection and $[a]_+=\max\{a,0\}$. The cost matrix $\mC\in\mathbb R^{n\times m}$ relates every base position to every source token using image features alone.

A row-wise softmax would allocate source features independently for each query. To account for how the same sources are used across queries, we instead couple their allocation through finite-step Sinkhorn-style scaling~\citep{cuturi2013sinkhorn}. Starting from the log-kernel $K_{ji}=-C_{ji}/\varepsilon$ and zero-initialized log-scaling factors $\alpha_j$ and $\beta_i$, each iteration updates the row factors and then the column factors. Row scaling regulates the total weight assigned to each query, whereas column scaling adjusts each source's weight jointly across all queries. After $L$ iterations, the allocation weights are
\begin{equation}
\begin{aligned}
T_{ji}&=\exp\!\left[\min\!\left(K_{ji}+\alpha_j^{(L)}+\PAIQmatch{\beta_i^{(L)}},0\right)\right],\\
\PAIQmatch{\pi_{ji}}&=\frac{T_{ji}}{\sum_{k=1}^{m}T_{jk}+\delta}.
\end{aligned}
\label{eq:allocation_weights}
\end{equation}
We use $\varepsilon=0.05$, $L=5$, and $\delta=10^{-6}$, with the clipped log-domain updates specified in Appendix~\ref{app:matching}.

The effect of column scaling becomes clear in the idealized case without clipping or the stabilizer ($\delta=0$), where the weights reduce to
\begin{equation}
\PAIQmatch{\pi^\circ_{ji}}=\operatorname{softmax}_i\!\left(-C_{ji}/\varepsilon+\PAIQmatch{\beta_i^{(L)}}\right).
\label{eq:source_bias}
\end{equation}
The source bias $\PAIQmatch{\beta_i^{(L)}}$ is computed jointly from all queries and shared across them, coupling content-based matching with image-level source allocation. This interpretation applies to the idealized scaling; the implemented finite, clipped iterations need not satisfy exact transport marginals (Appendix~\ref{app:matching}).

We use the resulting weights to aggregate the projected source features at each base position:
\begin{equation}
\bar{\vs}_j=\sum_{i=1}^{m}\PAIQmatch{\pi_{ji}}\vs_i,\qquad
\bar{\mS}=\PAIQmatch{\boldsymbol\Pi}\mS,
\label{eq:barycentric}
\end{equation}
where $\boldsymbol\Pi\in\mathbb R^{n\times m}$. The aggregate sequence $\bar{\mS}\in\mathbb R^{n\times d}$ follows the base grid while allowing multiple positions to draw on shared source features.

\subsection{Orthogonal residual enrichment}
\label{sec:orthogonal_injection}

Once complementary features have been aggregated, the remaining question is how they should update the local representation. For each position, we form a cross-encoder residual $\vr_j=\bar{\vs}_j-\vd_j$. Directly adding a scaled residual to the base gives the interpolation
\begin{equation}
\vz_j^{\mathrm{interp}}=\vd_j+\lambda\vr_j
=(1-\lambda)\vd_j+\lambda\bar{\vs}_j,
\label{eq:direct_interpolation}
\end{equation}
where $0<\lambda<1$. If two positions receive the same aggregate, this update contracts their original feature distance by a factor of $1-\lambda$. We therefore introduce a learnable residual transform to adapt the direction of enrichment while controlling its geometry:
\begin{equation}
\vz_j=\vd_j+\lambda\PAIQrotate{\mQ}\vr_j,\qquad
\PAIQrotate{\mQ^\top}\PAIQrotate{\mQ}=\mI.
\label{eq:orthogonal_update}
\end{equation}
The orthogonal constraint allows the update to change direction relative to the base while preserving residual lengths before scaling. A single $\mQ$ acts on the feature channels and is shared across all positions and images.

We enforce orthogonality through the Cayley parameterization. Given a trainable matrix $W_Q\in\mathbb R^{d\times d}$, we construct a skew-symmetric matrix $\mA$ and its associated rotation:
\begin{equation}
\mA=W_Q-W_Q^\top,\qquad
\PAIQrotate{\mQ}=(\mI-\mA/2)^{-1}(\mI+\mA/2).
\label{eq:cayley}
\end{equation}
Skew-symmetry ensures that $\mQ^\top\mQ=\mI$ (Appendix~\ref{app:geometry}). Combining the rotation with the aggregates in~\PAIQeqref{eq:barycentric}, the fused visual sequence is
\begin{equation}
\mZ=\mD+\lambda(\PAIQmatch{\boldsymbol\Pi}\mS-\mD)\PAIQrotate{\mQ^\top},
\label{eq:inject}
\end{equation}
where $\mZ\in\mathbb R^{n\times d}$. We fix $\lambda=0.6$ and initialize $W_Q$ to zero, so that fusion begins with direct interpolation and learns the residual rotation through training.

\paragraph{Residual geometry.}
Let $\mR=\bar{\mS}-\mD$ denote the residual matrix and $\boldsymbol\Delta=\mZ-\mD=\lambda\mR\mQ^\top$ the corresponding updates. Orthogonality yields
\begin{equation}
\boldsymbol\Delta\boldsymbol\Delta^\top=\lambda^2\mR\mR^\top,\qquad
\|\vz_j-\vd_j\|_2=\lambda\|\vr_j\|_2.
\label{eq:residual_gram}
\end{equation}
The rotation preserves angles between nonzero residuals, while $\lambda$ scales their lengths uniformly. Thus, for fixed residuals, direction adaptation is separated from scale control. This constraint governs the updates themselves; the geometry of the fused features also depends on their relation to the base (Appendix~\ref{app:geometry}).

\subsection{Local separability under shared semantics}
\label{sec:local_separability}

Different positions may receive similar aggregates while having distinct base features. To study the geometry of semantic sharing, we examine the Euclidean distance between their fused features for fixed projected bases and aggregates. Let $x_{jk}=\vd_j-\vd_k$ and $e_{jk}=\bar{\vs}_j-\bar{\vs}_k$ denote their base and aggregate differences, respectively. By~\PAIQeqref{eq:orthogonal_update},
\begin{equation}
\vz_j-\vz_k=(\mI-\lambda\mQ)x_{jk}+\lambda\mQ e_{jk}.
\label{eq:pair_decomposition}
\end{equation}
The following result characterizes how the rotation contributes to separation under shared or similar aggregates.

\begin{paiqtheorybox}
\begin{lemma}[Local separability under shared aggregates]
\label{lem:separation}
\leavevmode\par\nobreak
Let $0<\lambda<1$ and $\mQ^\top\mQ=\mI$. For fixed base and aggregate features, define
\begin{equation}
g_Q(x)=\sqrt{(1-\lambda)^2\|x\|_2^2+\PAIQrotate{\lambda\|(\mI-\mQ)x\|_2^2}}.
\label{eq:rotation_margin}
\end{equation}
\par\noindent
\textnormal{\textbf{(i) Shared aggregates.}} If $e_{jk}=0$, then
\begin{equation}
\|\vz_j-\vz_k\|_2^2=g_Q(x_{jk})^2.
\label{eq:contrast_identity}
\end{equation}
\par\noindent
\textnormal{\textbf{(ii) General aggregates.}} For any $j,k$,
\begin{equation}
\|\vz_j-\vz_k\|_2\geq\big[g_Q(x_{jk})-\lambda\|e_{jk}\|_2\big]_+.
\label{eq:rotation_aware_bound}
\end{equation}
\end{lemma}
\end{paiqtheorybox}

When two positions share an aggregate, the squared distance decomposes into an interpolation term and a nonnegative rotation term, as given by~\PAIQeqref{eq:rotation_margin}. Direct interpolation ($\mQ=\mI$) retains only the first term. For the same base and aggregate features, residual rotation yields a strictly larger distance whenever $\mQ x_{jk}\ne x_{jk}$. The general bound further shows that distinct base features remain distinct if $\|e_{jk}\|_2<g_Q(x_{jk})/\lambda$. Since $g_Q(x)\geq(1-\lambda)\|x\|_2$, the rotation can enlarge the range of aggregate differences covered by this sufficient condition.

The role of the orthogonal constraint becomes evident for an unrestricted residual map, which can eliminate all base differences under a shared aggregate: choosing $G=\lambda^{-1}\mI$ gives $\vd_j+\lambda G(\bar{\vs}_j-\vd_j)=\bar{\vs}_j$. Orthogonality excludes this collapse for distinct fixed bases and shared aggregates when $0<\lambda<1$. These results concern feature-level distances rather than lossless preservation of spatial information. Proofs and extensions are provided in Appendix~\ref{app:separation}.

\paragraph{Parameterization and expressivity.}
The separation analysis holds for fixed intermediate features, whereas the full model learns the content projections jointly with the fusion parameters. Under unrestricted affine content projections and linear matching maps, with fixed $\lambda$ and the same allocation rule, the rotated update and direct interpolation have identical input--output function classes in exact arithmetic (Lemma~\ref{lem:equivalence}, Appendix~\ref{app:equivalence}). The maps $(\mI-\lambda\mQ)/(1-\lambda)$ and $\mQ$ can be absorbed into the base and source projections, respectively, with compensating changes to the matching maps that preserve $\boldsymbol\Pi$. The explicit rotation therefore defines a geometry-constrained training parameterization of the fusion mechanism; function-class equivalence alone does not determine its learning behavior.

\subsection{Learning the visual interface}
\label{sec:visual_learning}

The fused sequence retains the $n=196$ positions of the \DINO{} grid. We use $\mZ(I)$ to replace the corresponding image-placeholder embeddings in the language-model input. Both visual encoders and the language model, including its output head, remain frozen; only the projections, matching maps, and residual-transform parameters $\theta=\{P_d,b_d,P_s,b_s,W_d,W_s,W_Q\}$ are learned.

Given a text instruction $T$ and a target answer $y$ including the end-of-sequence (EOS) token, we optimize the autoregressive answer-generation objective
\begin{equation}
\mathcal L_{\mathrm{NLL}}(\theta)
=-\sum_{t=1}^{|y|}\log p_\theta(y_t\mid y_{<t},T,\mZ(I)).
\label{eq:training}
\end{equation}
Supervision is applied only to the answer tokens and EOS. Gradients pass through the frozen language model to the visual interface, allowing the answer loss alone to jointly optimize the projections, source allocation, and residual updates. Orthogonality is enforced directly by the parameterization in~\PAIQeqref{eq:cayley}.

Post-training diagnostics of aggregation, update magnitude, and fixed-answer sensitivity are described in Appendix~\ref{app:diagnostics}.

%% file: sections/experiments.tex
\section{Experiments}
\label{sec:experiments}

\subsection{Experimental setup}

\paragraph{Tasks and baselines.}
We evaluate image description on FLUX-Reason~\citep{fang2025fluxreason} and a CC3M slice (Caption)~\citep{sharma2018conceptual}, and visual question answering (VQA) on SEED-Bench~\citep{li2023seedbench} and A-OKVQA~\citep{schwenk2022aokvqa}. FLUX shares its source dataset with training and serves as an in-domain description diagnostic; the remaining datasets assess cross-source transfer. We compare \modelname{} with independently trained \DINO{}-only and \SIGLIP{}-only interfaces using the same training data, optimization recipe, and fixed 1,000 evaluation records per dataset within each language backbone. Table~\ref{tab:main} additionally includes \GRANULON{}, CoME-VL, MERV, and LLaVA-Mini, spanning alternative approaches to visual feature fusion and token compression. Training and evaluation settings are detailed in Appendix~\ref{app:experiment_settings}.

\paragraph{Evaluation protocol.}
Following Granulon~\citep{Granulon}, we use an automated judge to score Accuracy (response correctness) and Hallucination (severity of unsupported or contradictory content) independently on a $[0,100]$ scale. Image descriptions are judged against the image, while VQA responses are evaluated with the image, question, answer options, and reference answer. Higher Accuracy and lower Hallucination are preferred; these scores differ from benchmark accuracy and hallucination rates. We complement this evaluation with closed-ended VQA independent of the judge. Appendix~\ref{app:evaluation} details scoring, Appendix~\ref{app:mcq_evaluation} specifies the closed-ended protocols, and Appendix~\ref{app:prompts} documents the prompts.

\subsection{Effectiveness across visual-language tasks}

\paragraph{Gains from complementary representations.}
Across the four language backbones in Table~\ref{tab:main}, \modelname{} exceeds both matched single-encoder interfaces in Accuracy in 15 of 16 backbone--dataset settings and lowers Hallucination in all 16. The gains extend from open-ended description to question-conditioned understanding: for Qwen3.5-9B, Accuracy improves over the stronger single-encoder baseline by 7.34 and 11.00 score points on SEED and A-OKVQA, respectively.

\begin{table}[H]
\setlength{\belowcaptionskip}{6pt}
\caption{Comparison across four language backbones and four datasets. Accuracy ($\uparrow$) and Hallucination ($\downarrow$) are mean judge scores on $[0,100]$. The matched DINOv3, SigLIP, and \modelname{} interfaces use 1,000 examples per setting. \textbf{Bold} and \underline{underlining} mark the best and second-best displayed values, respectively, within each backbone and task metric. Colored arrows beside each baseline give the direction and size of \modelname{}'s difference from that baseline in score points; green favors \modelname{} and red favors the baseline.}
\label{tab:main}
\PAIQSetup
\setlength{\tabcolsep}{1.8pt}
\begin{tabularx}{\linewidth}{@{}>{\columncolor{white}[0pt][\dimexpr\tabcolsep+8.5pt\relax]\hspace{5pt}\raggedright\arraybackslash}p{57pt}*{7}{>{\columncolor{white}[\dimexpr\tabcolsep-8.5pt\relax][\dimexpr\tabcolsep+8.5pt\relax]\centering\arraybackslash}X}>{\columncolor{white}[\dimexpr\tabcolsep-8.5pt\relax][11pt]\centering\arraybackslash}X@{\hspace{11pt}}}
\toprule
\textbf{Method} & \multicolumn{2}{c}{\textbf{FLUX-Reason}} & \multicolumn{2}{c}{\textbf{Caption}} & \multicolumn{2}{c}{\textbf{SEED}} & \multicolumn{2}{c@{\hspace{11pt}}}{\textbf{A-OKVQA}} \\
\cmidrule(lr){2-3}\cmidrule(lr){4-5}\cmidrule(lr){6-7}\cmidrule(l){8-9}
 & \PAIQAcc & \PAIQHall & \PAIQAcc & \PAIQHall & \PAIQAcc & \PAIQHall & \PAIQAcc & \PAIQHall \\
\rowcolor{PAIQLavender}[0pt][0pt]
\multicolumn{9}{@{}l@{}}{\PAIQBand{Qwen3.5-2B}} \\
\rowcolor{PAIQGray}
DINOv3 & \PAIQGain{40.59}{PAIQGreen}{\uparrow}{49.38} & \PAIQGain{68.66}{PAIQGreen}{\downarrow}{48.50} & \PAIQGain{26.80}{PAIQGreen}{\uparrow}{54.47} & \PAIQGain{80.60}{PAIQGreen}{\downarrow}{48.36} & \PAIQGain{32.70}{PAIQGreen}{\uparrow}{21.32} & \PAIQGain{72.80}{PAIQGreen}{\downarrow}{20.17} & \PAIQGain{28.29}{PAIQGreen}{\uparrow}{20.56} & \PAIQGain{77.94}{PAIQGreen}{\downarrow}{22.46} \\
\rowcolor{white}
SigLIP & \PAIQGain{31.72}{PAIQGreen}{\uparrow}{58.25} & \PAIQGain{75.42}{PAIQGreen}{\downarrow}{55.26} & \PAIQGain{19.40}{PAIQGreen}{\uparrow}{61.87} & \PAIQGain{85.59}{PAIQGreen}{\downarrow}{53.35} & \PAIQGain{29.77}{PAIQGreen}{\uparrow}{24.25} & \PAIQGain{74.71}{PAIQGreen}{\downarrow}{22.08} & \PAIQGain{27.68}{PAIQGreen}{\uparrow}{21.17} & \PAIQGain{78.86}{PAIQGreen}{\downarrow}{23.38} \\
\rowcolor{PAIQGray}
Granulon & \PAIQGain{57.52}{PAIQGreen}{\uparrow}{32.45} & \PAIQGain{54.13}{PAIQGreen}{\downarrow}{33.97} & \PAIQGain{38.57}{PAIQGreen}{\uparrow}{42.70} & \PAIQGain{72.43}{PAIQGreen}{\downarrow}{40.19} & \PAIQGain{37.66}{PAIQGreen}{\uparrow}{16.36} & \PAIQGain{69.85}{PAIQGreen}{\downarrow}{17.22} & \PAIQGain{34.26}{PAIQGreen}{\uparrow}{14.59} & \PAIQGain{73.85}{PAIQGreen}{\downarrow}{18.37} \\
\rowcolor{white}
CoME-VL & \PAIQGain{80.83}{PAIQGreen}{\uparrow}{9.14} & \PAIQGain{29.31}{PAIQGreen}{\downarrow}{9.15} & \PAIQGain{69.76}{PAIQGreen}{\uparrow}{11.51} & \PAIQGain{41.10}{PAIQGreen}{\downarrow}{8.86} & \PAIQGain{47.28}{PAIQGreen}{\uparrow}{6.74} & \PAIQGain{55.43}{PAIQGreen}{\downarrow}{2.80} & \PAIQGain{\underline{44.25}}{PAIQGreen}{\uparrow}{4.60} & \PAIQGain{59.30}{PAIQGreen}{\downarrow}{3.82} \\
\rowcolor{PAIQGray}
MERV & \PAIQGain{84.35}{PAIQGreen}{\uparrow}{5.62} & \PAIQGain{24.94}{PAIQGreen}{\downarrow}{4.78} & \PAIQGain{72.93}{PAIQGreen}{\uparrow}{8.34} & \PAIQGain{37.29}{PAIQGreen}{\downarrow}{5.05} & \PAIQGain{\underline{49.24}}{PAIQGreen}{\uparrow}{4.78} & \PAIQGain{\textbf{51.92}}{PAIQRed}{\uparrow}{0.71} & \PAIQGain{43.49}{PAIQGreen}{\uparrow}{5.36} & \PAIQGain{\underline{57.16}}{PAIQGreen}{\downarrow}{1.68} \\
\rowcolor{white}
LLaVA-Mini & \PAIQGain{\underline{86.91}}{PAIQGreen}{\uparrow}{3.06} & \PAIQGain{\underline{21.28}}{PAIQGreen}{\downarrow}{1.12} & \PAIQGain{\underline{77.33}}{PAIQGreen}{\uparrow}{3.94} & \PAIQGain{\underline{32.44}}{PAIQGreen}{\downarrow}{0.20} & \PAIQGain{44.38}{PAIQGreen}{\uparrow}{9.64} & \PAIQGain{58.90}{PAIQGreen}{\downarrow}{6.27} & \PAIQGain{33.77}{PAIQGreen}{\uparrow}{15.08} & \PAIQGain{66.34}{PAIQGreen}{\downarrow}{10.86} \\
\rowcolor{PAIQYellow}
\textbf{\modelname{}} & \PAIQOurs{\textbf{89.97}} & \PAIQOurs{\textbf{20.16}} & \PAIQOurs{\textbf{81.27}} & \PAIQOurs{\textbf{32.24}} & \PAIQOurs{\textbf{54.02}} & \PAIQOurs{\underline{52.63}} & \PAIQOurs{\textbf{48.85}} & \PAIQOurs{\textbf{55.48}} \\
\addlinespace[4pt]
\rowcolor{PAIQLavender}[0pt][0pt]
\multicolumn{9}{@{}l@{}}{\PAIQBand{Qwen3.5-9B}} \\
\rowcolor{PAIQGray}
DINOv3 & \PAIQGain{47.06}{PAIQGreen}{\uparrow}{46.14} & \PAIQGain{62.43}{PAIQGreen}{\downarrow}{46.94} & \PAIQGain{22.08}{PAIQGreen}{\uparrow}{62.94} & \PAIQGain{69.17}{PAIQGreen}{\downarrow}{43.77} & \PAIQGain{35.64}{PAIQGreen}{\uparrow}{25.14} & \PAIQGain{65.80}{PAIQGreen}{\downarrow}{22.79} & \PAIQGain{32.05}{PAIQGreen}{\uparrow}{29.06} & \PAIQGain{68.90}{PAIQGreen}{\downarrow}{27.24} \\
\rowcolor{white}
SigLIP & \PAIQGain{80.51}{PAIQGreen}{\uparrow}{12.69} & \PAIQGain{31.02}{PAIQGreen}{\downarrow}{15.53} & \PAIQGain{58.62}{PAIQGreen}{\uparrow}{26.40} & \PAIQGain{50.97}{PAIQGreen}{\downarrow}{25.57} & \PAIQGain{53.44}{PAIQGreen}{\uparrow}{7.34} & \PAIQGain{50.98}{PAIQGreen}{\downarrow}{7.97} & \PAIQGain{50.11}{PAIQGreen}{\uparrow}{11.00} & \PAIQGain{53.77}{PAIQGreen}{\downarrow}{12.11} \\
\rowcolor{PAIQGray}
Granulon & \PAIQGain{84.32}{PAIQGreen}{\uparrow}{8.88} & \PAIQGain{27.51}{PAIQGreen}{\downarrow}{12.02} & \PAIQGain{69.27}{PAIQGreen}{\uparrow}{15.75} & \PAIQGain{41.97}{PAIQGreen}{\downarrow}{16.57} & \PAIQGain{54.63}{PAIQGreen}{\uparrow}{6.15} & \PAIQGain{46.20}{PAIQGreen}{\downarrow}{3.19} & \PAIQGain{56.30}{PAIQGreen}{\uparrow}{4.81} & \PAIQGain{45.17}{PAIQGreen}{\downarrow}{3.51} \\
\rowcolor{white}
CoME-VL & \PAIQGain{90.34}{PAIQGreen}{\uparrow}{2.86} & \PAIQGain{\underline{17.69}}{PAIQGreen}{\downarrow}{2.20} & \PAIQGain{\underline{83.91}}{PAIQGreen}{\uparrow}{1.11} & \PAIQGain{\textbf{23.72}}{PAIQRed}{\uparrow}{1.68} & \PAIQGain{\underline{59.65}}{PAIQGreen}{\uparrow}{1.13} & \PAIQGain{43.80}{PAIQGreen}{\downarrow}{0.79} & \PAIQGain{\underline{58.89}}{PAIQGreen}{\uparrow}{2.22} & \PAIQGain{\underline{42.72}}{PAIQGreen}{\downarrow}{1.06} \\
\rowcolor{PAIQGray}
MERV & \PAIQGain{\underline{90.54}}{PAIQGreen}{\uparrow}{2.66} & \PAIQGain{20.78}{PAIQGreen}{\downarrow}{5.29} & \PAIQGain{77.35}{PAIQGreen}{\uparrow}{7.67} & \PAIQGain{31.67}{PAIQGreen}{\downarrow}{6.27} & \PAIQGain{59.14}{PAIQGreen}{\uparrow}{1.64} & \PAIQGain{\textbf{42.94}}{PAIQRed}{\uparrow}{0.07} & \PAIQGain{55.68}{PAIQGreen}{\uparrow}{5.43} & \PAIQGain{45.61}{PAIQGreen}{\downarrow}{3.95} \\
\rowcolor{white}
LLaVA-Mini & \PAIQGain{89.29}{PAIQGreen}{\uparrow}{3.91} & \PAIQGain{22.63}{PAIQGreen}{\downarrow}{7.14} & \PAIQGain{74.15}{PAIQGreen}{\uparrow}{10.87} & \PAIQGain{41.15}{PAIQGreen}{\downarrow}{15.75} & \PAIQGain{39.02}{PAIQGreen}{\uparrow}{21.76} & \PAIQGain{61.56}{PAIQGreen}{\downarrow}{18.55} & \PAIQGain{31.41}{PAIQGreen}{\uparrow}{29.70} & \PAIQGain{69.55}{PAIQGreen}{\downarrow}{27.89} \\
\rowcolor{PAIQYellow}
\textbf{\modelname{}} & \PAIQOurs{\textbf{93.20}} & \PAIQOurs{\textbf{15.49}} & \PAIQOurs{\textbf{85.02}} & \PAIQOurs{\underline{25.40}} & \PAIQOurs{\textbf{60.78}} & \PAIQOurs{\underline{43.01}} & \PAIQOurs{\textbf{61.11}} & \PAIQOurs{\textbf{41.66}} \\
\addlinespace[4pt]
\rowcolor{PAIQLavender}[0pt][0pt]
\multicolumn{9}{@{}l@{}}{\PAIQBand{InternVL3.5-8B}} \\
\rowcolor{PAIQGray}
DINOv3 & \PAIQGain{82.05}{PAIQGreen}{\uparrow}{9.14} & \PAIQGain{29.13}{PAIQGreen}{\downarrow}{10.81} & \PAIQGain{71.99}{PAIQGreen}{\uparrow}{15.16} & \PAIQGain{41.11}{PAIQGreen}{\downarrow}{17.90} & \PAIQGain{\underline{48.24}}{PAIQGreen}{\uparrow}{1.45} & \PAIQGain{52.67}{PAIQGreen}{\downarrow}{2.59} & \PAIQGain{\textbf{52.45}}{PAIQRed}{\downarrow}{0.32} & \PAIQGain{50.36}{PAIQGreen}{\downarrow}{1.41} \\
\rowcolor{white}
SigLIP & \PAIQGain{87.91}{PAIQGreen}{\uparrow}{3.28} & \PAIQGain{22.46}{PAIQGreen}{\downarrow}{4.14} & \PAIQGain{78.71}{PAIQGreen}{\uparrow}{8.44} & \PAIQGain{33.84}{PAIQGreen}{\downarrow}{10.63} & \PAIQGain{44.49}{PAIQGreen}{\uparrow}{5.20} & \PAIQGain{56.71}{PAIQGreen}{\downarrow}{6.63} & \PAIQGain{48.49}{PAIQGreen}{\uparrow}{3.64} & \PAIQGain{55.47}{PAIQGreen}{\downarrow}{6.52} \\
\rowcolor{PAIQGray}
Granulon & \PAIQGain{85.98}{PAIQGreen}{\uparrow}{5.21} & \PAIQGain{24.99}{PAIQGreen}{\downarrow}{6.67} & \PAIQGain{77.96}{PAIQGreen}{\uparrow}{9.19} & \PAIQGain{36.15}{PAIQGreen}{\downarrow}{12.94} & \PAIQGain{44.46}{PAIQGreen}{\uparrow}{5.23} & \PAIQGain{\underline{49.51}}{PAIQRed}{\uparrow}{0.57} & \PAIQGain{49.30}{PAIQGreen}{\uparrow}{2.83} & \PAIQGain{\textbf{47.20}}{PAIQRed}{\uparrow}{1.75} \\
\rowcolor{white}
CoME-VL & \PAIQGain{88.76}{PAIQGreen}{\uparrow}{2.43} & \PAIQGain{\underline{18.69}}{PAIQGreen}{\downarrow}{0.37} & \PAIQGain{84.65}{PAIQGreen}{\uparrow}{2.50} & \PAIQGain{24.88}{PAIQGreen}{\downarrow}{1.67} & \PAIQGain{44.05}{PAIQGreen}{\uparrow}{5.64} & \PAIQGain{54.14}{PAIQGreen}{\downarrow}{4.06} & \PAIQGain{49.04}{PAIQGreen}{\uparrow}{3.09} & \PAIQGain{50.87}{PAIQGreen}{\downarrow}{1.92} \\
\rowcolor{PAIQGray}
MERV & \PAIQGain{\underline{90.19}}{PAIQGreen}{\uparrow}{1.00} & \PAIQGain{20.41}{PAIQGreen}{\downarrow}{2.09} & \PAIQGain{\underline{86.80}}{PAIQGreen}{\uparrow}{0.35} & \PAIQGain{\underline{24.61}}{PAIQGreen}{\downarrow}{1.40} & \PAIQGain{47.80}{PAIQGreen}{\uparrow}{1.89} & \PAIQGain{\textbf{49.44}}{PAIQRed}{\uparrow}{0.64} & \PAIQGain{50.16}{PAIQGreen}{\uparrow}{1.97} & \PAIQGain{\underline{47.76}}{PAIQRed}{\uparrow}{1.19} \\
\rowcolor{white}
LLaVA-Mini & \PAIQGain{89.99}{PAIQGreen}{\uparrow}{1.20} & \PAIQGain{22.22}{PAIQGreen}{\downarrow}{3.90} & \PAIQGain{74.25}{PAIQGreen}{\uparrow}{12.90} & \PAIQGain{39.89}{PAIQGreen}{\downarrow}{16.68} & \PAIQGain{29.65}{PAIQGreen}{\uparrow}{20.04} & \PAIQGain{72.88}{PAIQGreen}{\downarrow}{22.80} & \PAIQGain{23.81}{PAIQGreen}{\uparrow}{28.32} & \PAIQGain{76.15}{PAIQGreen}{\downarrow}{27.20} \\
\rowcolor{PAIQYellow}
\textbf{\modelname{}} & \PAIQOurs{\textbf{91.19}} & \PAIQOurs{\textbf{18.32}} & \PAIQOurs{\textbf{87.15}} & \PAIQOurs{\textbf{23.21}} & \PAIQOurs{\textbf{49.69}} & \PAIQOurs{50.08} & \PAIQOurs{\underline{52.13}} & \PAIQOurs{48.95} \\
\addlinespace[4pt]
\rowcolor{PAIQLavender}[0pt][0pt]
\multicolumn{9}{@{}l@{}}{\PAIQBand{Llama-3.1-8B}} \\
\rowcolor{PAIQGray}
DINOv3 & \PAIQGain{34.14}{PAIQGreen}{\uparrow}{54.20} & \PAIQGain{70.95}{PAIQGreen}{\downarrow}{46.97} & \PAIQGain{18.67}{PAIQGreen}{\uparrow}{58.71} & \PAIQGain{82.15}{PAIQGreen}{\downarrow}{45.40} & \PAIQGain{29.89}{PAIQGreen}{\uparrow}{9.98} & \PAIQGain{74.52}{PAIQGreen}{\downarrow}{8.01} & \PAIQGain{28.57}{PAIQGreen}{\uparrow}{7.41} & \PAIQGain{77.67}{PAIQGreen}{\downarrow}{7.41} \\
\rowcolor{white}
SigLIP & \PAIQGain{41.58}{PAIQGreen}{\uparrow}{46.76} & \PAIQGain{68.79}{PAIQGreen}{\downarrow}{44.81} & \PAIQGain{20.44}{PAIQGreen}{\uparrow}{56.94} & \PAIQGain{83.09}{PAIQGreen}{\downarrow}{46.34} & \PAIQGain{18.95}{PAIQGreen}{\uparrow}{20.92} & \PAIQGain{80.35}{PAIQGreen}{\downarrow}{13.84} & \PAIQGain{19.12}{PAIQGreen}{\uparrow}{16.86} & \PAIQGain{82.02}{PAIQGreen}{\downarrow}{11.76} \\
\rowcolor{PAIQGray}
Granulon & \PAIQGain{73.48}{PAIQGreen}{\uparrow}{14.86} & \PAIQGain{39.94}{PAIQGreen}{\downarrow}{15.96} & \PAIQGain{56.33}{PAIQGreen}{\uparrow}{21.05} & \PAIQGain{58.17}{PAIQGreen}{\downarrow}{21.42} & \PAIQGain{30.48}{PAIQGreen}{\uparrow}{9.39} & \PAIQGain{68.36}{PAIQGreen}{\downarrow}{1.85} & \PAIQGain{33.68}{PAIQGreen}{\uparrow}{2.30} & \PAIQGain{\underline{64.95}}{PAIQRed}{\uparrow}{5.31} \\
\rowcolor{white}
CoME-VL & \PAIQGain{51.26}{PAIQGreen}{\uparrow}{37.08} & \PAIQGain{58.90}{PAIQGreen}{\downarrow}{34.92} & \PAIQGain{27.87}{PAIQGreen}{\uparrow}{49.51} & \PAIQGain{78.28}{PAIQGreen}{\downarrow}{41.53} & \PAIQGain{13.06}{PAIQGreen}{\uparrow}{26.81} & \PAIQGain{84.59}{PAIQGreen}{\downarrow}{18.08} & \PAIQGain{12.87}{PAIQGreen}{\uparrow}{23.11} & \PAIQGain{87.35}{PAIQGreen}{\downarrow}{17.09} \\
\rowcolor{PAIQGray}
MERV & \PAIQGain{85.30}{PAIQGreen}{\uparrow}{3.04} & \PAIQGain{27.65}{PAIQGreen}{\downarrow}{3.67} & \PAIQGain{73.53}{PAIQGreen}{\uparrow}{3.85} & \PAIQGain{40.48}{PAIQGreen}{\downarrow}{3.73} & \PAIQGain{\textbf{44.78}}{PAIQRed}{\downarrow}{4.91} & \PAIQGain{\textbf{55.88}}{PAIQRed}{\uparrow}{10.63} & \PAIQGain{\textbf{43.70}}{PAIQRed}{\downarrow}{7.72} & \PAIQGain{\textbf{58.51}}{PAIQRed}{\uparrow}{11.75} \\
\rowcolor{white}
LLaVA-Mini & \PAIQGain{\textbf{89.38}}{PAIQRed}{\downarrow}{1.04} & \PAIQGain{\textbf{23.18}}{PAIQRed}{\uparrow}{0.80} & \PAIQGain{\textbf{78.55}}{PAIQRed}{\downarrow}{1.17} & \PAIQGain{\textbf{36.65}}{PAIQRed}{\uparrow}{0.10} & \PAIQGain{5.28}{PAIQGreen}{\uparrow}{34.59} & \PAIQGain{91.08}{PAIQGreen}{\downarrow}{24.57} & \PAIQGain{3.22}{PAIQGreen}{\uparrow}{32.76} & \PAIQGain{92.52}{PAIQGreen}{\downarrow}{22.26} \\
\rowcolor{PAIQYellow}
\textbf{\modelname{}} & \PAIQOurs{\underline{88.34}} & \PAIQOurs{\underline{23.98}} & \PAIQOurs{\underline{77.38}} & \PAIQOurs{\underline{36.75}} & \PAIQOurs{\underline{39.87}} & \PAIQOurs{\underline{66.51}} & \PAIQOurs{\underline{35.98}} & \PAIQOurs{70.26} \\
\bottomrule
\end{tabularx}
\end{table}

\paragraph{Comparison with alternative visual interfaces.}
On Qwen3.5-2B and 9B, \modelname{} achieves the highest Accuracy on all four datasets, outperforming the strongest evaluated alternative by 1.11--4.78 score points (2.94 points on average). On InternVL3.5-8B, it leads on three of four datasets; on Llama-3.1-8B, LLaVA-Mini remains stronger on image description and MERV on VQA.

\subsection{Scaling and visual-token budget}
\label{sec:scaling}

\paragraph{Scaling the language backbone.}
The advantage of complementary representations persists at 27B (Table~\ref{tab:qwen_scaling}): relative to the stronger single-encoder interface, \modelname{} improves Accuracy by 3.09--6.16 points and reduces Hallucination by 4.70--11.54 points across the four datasets. All interfaces at 27B use a 512-token generation limit, compared with 128 at smaller scales; the gains are thus interpreted within each scale.

\begin{table}[!t]
\setlength{\belowcaptionskip}{6pt}
\caption{Comparison with matched single-encoder interfaces across Qwen3.5 scales (1,000 examples per setting). \textbf{Bold} and \underline{underlining} mark the best and second-best means, respectively, for each metric at each scale. Colored arrows beside baseline means show the direction and size of \modelname{}'s difference in score points; green favors \modelname{} and red favors the baseline.}
\label{tab:qwen_scaling}
\PAIQSetup
\setlength{\tabcolsep}{1.8pt}
\begin{tabularx}{\linewidth}{@{}>{\columncolor{white}[0pt][\dimexpr\tabcolsep+8.5pt\relax]\hspace{5pt}\raggedright\arraybackslash}p{39pt}*{7}{>{\columncolor{white}[\dimexpr\tabcolsep-8.5pt\relax][\dimexpr\tabcolsep+8.5pt\relax]\centering\arraybackslash}X}>{\columncolor{white}[\dimexpr\tabcolsep-8.5pt\relax][11pt]\centering\arraybackslash}X@{\hspace{11pt}}}
\toprule
\textbf{Method} & \multicolumn{2}{c}{\textbf{FLUX-Reason}} & \multicolumn{2}{c}{\textbf{Caption}} & \multicolumn{2}{c}{\textbf{SEED}} & \multicolumn{2}{c@{\hspace{11pt}}}{\textbf{A-OKVQA}} \\
\cmidrule(lr){2-3}\cmidrule(lr){4-5}\cmidrule(lr){6-7}\cmidrule(l){8-9}
 & \PAIQAcc & \PAIQHall & \PAIQAcc & \PAIQHall & \PAIQAcc & \PAIQHall & \PAIQAcc & \PAIQHall \\
\rowcolor{PAIQLavender}
\multicolumn{9}{@{}>{\columncolor{PAIQLavender}[0pt][0pt]}l@{}}{\PAIQBand{Qwen3.5-2B}} \\
\rowcolor{PAIQGray}
DINOv3 & \PAIQGain{\underline{40.59}}{PAIQGreen}{\uparrow}{49.38} & \PAIQGain{\underline{68.66}}{PAIQGreen}{\downarrow}{48.50} & \PAIQGain{\underline{26.80}}{PAIQGreen}{\uparrow}{54.47} & \PAIQGain{\underline{80.60}}{PAIQGreen}{\downarrow}{48.36} & \PAIQGain{\underline{32.70}}{PAIQGreen}{\uparrow}{21.32} & \PAIQGain{\underline{72.80}}{PAIQGreen}{\downarrow}{20.17} & \PAIQGain{\underline{28.29}}{PAIQGreen}{\uparrow}{20.56} & \PAIQGain{\underline{77.94}}{PAIQGreen}{\downarrow}{22.46} \\
\rowcolor{white}
SigLIP & \PAIQGain{31.72}{PAIQGreen}{\uparrow}{58.25} & \PAIQGain{75.42}{PAIQGreen}{\downarrow}{55.26} & \PAIQGain{19.40}{PAIQGreen}{\uparrow}{61.87} & \PAIQGain{85.59}{PAIQGreen}{\downarrow}{53.35} & \PAIQGain{29.77}{PAIQGreen}{\uparrow}{24.25} & \PAIQGain{74.71}{PAIQGreen}{\downarrow}{22.08} & \PAIQGain{27.68}{PAIQGreen}{\uparrow}{21.17} & \PAIQGain{78.86}{PAIQGreen}{\downarrow}{23.38} \\
\rowcolor{PAIQYellow}
\textbf{\modelname{}} & \PAIQOurs{\textbf{89.97}} & \PAIQOurs{\textbf{20.16}} & \PAIQOurs{\textbf{81.27}} & \PAIQOurs{\textbf{32.24}} & \PAIQOurs{\textbf{54.02}} & \PAIQOurs{\textbf{52.63}} & \PAIQOurs{\textbf{48.85}} & \PAIQOurs{\textbf{55.48}} \\
\addlinespace[4pt]
\rowcolor{PAIQLavender}
\multicolumn{9}{@{}>{\columncolor{PAIQLavender}[0pt][0pt]}l@{}}{\PAIQBand{Qwen3.5-9B}} \\
\rowcolor{PAIQGray}
DINOv3 & \PAIQGain{47.06}{PAIQGreen}{\uparrow}{46.14} & \PAIQGain{62.43}{PAIQGreen}{\downarrow}{46.94} & \PAIQGain{22.08}{PAIQGreen}{\uparrow}{62.94} & \PAIQGain{69.17}{PAIQGreen}{\downarrow}{43.77} & \PAIQGain{35.64}{PAIQGreen}{\uparrow}{25.14} & \PAIQGain{65.80}{PAIQGreen}{\downarrow}{22.79} & \PAIQGain{32.05}{PAIQGreen}{\uparrow}{29.06} & \PAIQGain{68.90}{PAIQGreen}{\downarrow}{27.24} \\
\rowcolor{white}
SigLIP & \PAIQGain{\underline{80.51}}{PAIQGreen}{\uparrow}{12.69} & \PAIQGain{\underline{31.02}}{PAIQGreen}{\downarrow}{15.53} & \PAIQGain{\underline{58.62}}{PAIQGreen}{\uparrow}{26.40} & \PAIQGain{\underline{50.97}}{PAIQGreen}{\downarrow}{25.57} & \PAIQGain{\underline{53.44}}{PAIQGreen}{\uparrow}{7.34} & \PAIQGain{\underline{50.98}}{PAIQGreen}{\downarrow}{7.97} & \PAIQGain{\underline{50.11}}{PAIQGreen}{\uparrow}{11.00} & \PAIQGain{\underline{53.77}}{PAIQGreen}{\downarrow}{12.11} \\
\rowcolor{PAIQYellow}
\textbf{\modelname{}} & \PAIQOurs{\textbf{93.20}} & \PAIQOurs{\textbf{15.49}} & \PAIQOurs{\textbf{85.02}} & \PAIQOurs{\textbf{25.40}} & \PAIQOurs{\textbf{60.78}} & \PAIQOurs{\textbf{43.01}} & \PAIQOurs{\textbf{61.11}} & \PAIQOurs{\textbf{41.66}} \\
\addlinespace[4pt]
\rowcolor{PAIQLavender}
\multicolumn{9}{@{}>{\columncolor{PAIQLavender}[0pt][0pt]}l@{}}{\PAIQBand{Qwen3.5-27B}} \\
\rowcolor{PAIQGray}
DINOv3 & \PAIQGain{70.30}{PAIQGreen}{\uparrow}{22.86} & \PAIQGain{42.60}{PAIQGreen}{\downarrow}{25.27} & \PAIQGain{41.24}{PAIQGreen}{\uparrow}{44.42} & \PAIQGain{56.90}{PAIQGreen}{\downarrow}{32.07} & \PAIQGain{47.31}{PAIQGreen}{\uparrow}{27.94} & \PAIQGain{53.98}{PAIQGreen}{\downarrow}{28.77} & \PAIQGain{56.13}{PAIQGreen}{\uparrow}{25.71} & \PAIQGain{49.97}{PAIQGreen}{\downarrow}{31.32} \\
\rowcolor{white}
SigLIP & \PAIQGain{\underline{89.70}}{PAIQGreen}{\uparrow}{3.46} & \PAIQGain{\underline{22.03}}{PAIQGreen}{\downarrow}{4.70} & \PAIQGain{\underline{82.57}}{PAIQGreen}{\uparrow}{3.09} & \PAIQGain{\underline{31.57}}{PAIQGreen}{\downarrow}{6.74} & \PAIQGain{\underline{69.09}}{PAIQGreen}{\uparrow}{6.16} & \PAIQGain{\underline{36.75}}{PAIQGreen}{\downarrow}{11.54} & \PAIQGain{\underline{76.16}}{PAIQGreen}{\uparrow}{5.68} & \PAIQGain{\underline{28.67}}{PAIQGreen}{\downarrow}{10.02} \\
\rowcolor{PAIQYellow}
\textbf{\modelname{}} & \PAIQOurs{\textbf{93.16}} & \PAIQOurs{\textbf{17.33}} & \PAIQOurs{\textbf{85.66}} & \PAIQOurs{\textbf{24.83}} & \PAIQOurs{\textbf{75.25}} & \PAIQOurs{\textbf{25.21}} & \PAIQOurs{\textbf{81.84}} & \PAIQOurs{\textbf{18.65}} \\
\bottomrule
\end{tabularx}
\end{table}

\begin{figure}[!t]
  \vspace{-7pt}
  \centering
  \begin{minipage}[t]{0.49\textwidth}
    \vspace{0pt}
    \centering
    \includegraphics[width=\linewidth]{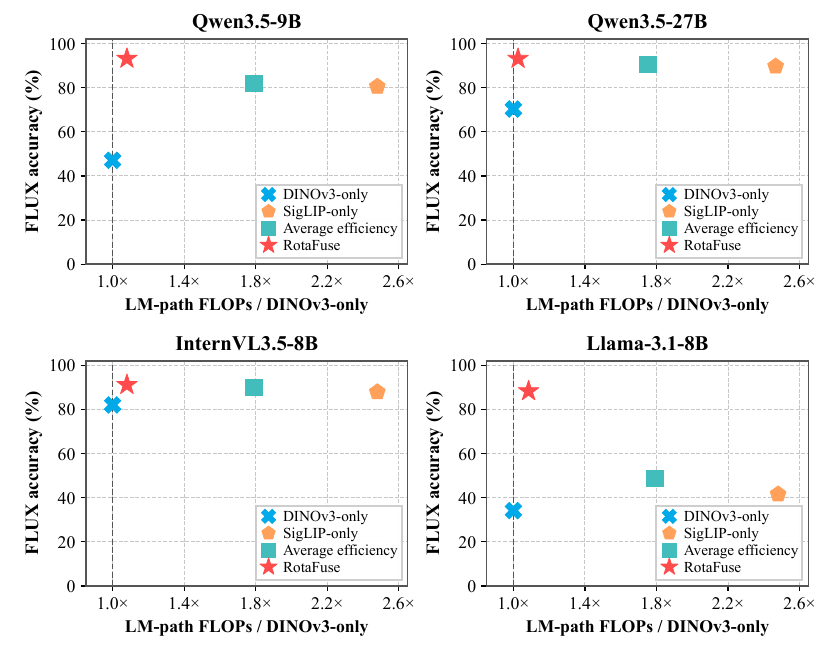}\vspace{-3pt}
    \caption{FLUX Accuracy versus normalized analytical compute. The best-of-two Accuracy is plotted at the average-efficiency coordinate; the proxy excludes fusion (Appendix~\ref{app:compute}).}
    \label{fig:oracle_efficiency}
  \end{minipage}\hfill
  \begin{minipage}[t]{0.48\textwidth}
    \vspace{0pt}
    \paragraph{Feature-level complementarity.}
    On FLUX, \modelname{} surpasses the post-hoc best-of-two Accuracy reference across all four backbones shown in Figure~\ref{fig:oracle_efficiency}. This reference selects the single-encoder response with higher Accuracy for each example (Appendix~\ref{app:oracle}), showing a benefit beyond selecting between the two observed outputs.

\paragraph{Compact visual interface.}
\modelname{} combines both feature streams in 196 visual tokens, compared with 576 for \mbox{\SIGLIP{}-only} and 772 for concatenation. The analytical compute proxy is 133\% of the \DINO{}-only reference at 2B and 103--108\% at larger scales (Table~\ref{tab:efficiency}). The proxy excludes fusion, while the fixed token budget directly limits the visual prefix of the language model.
  \end{minipage}
  \vspace{-8pt}
\end{figure}

\begin{table}[!t]
\setlength{\belowcaptionskip}{6pt}
\caption{Visual-token budgets and analytical compute relative to \DINO{}-only, excluding fusion (Appendix~\ref{app:compute}). Average efficiency uses the mean prefix length of 386 as an analytical reference; concatenation would expose 772 tokens.}
\label{tab:efficiency}
\centering
\small
\setlength{\tabcolsep}{5pt}
\renewcommand{\arraystretch}{1.18}
\arrayrulecolor{black}
\begin{tabularx}{\textwidth}{@{\hspace{\tabcolsep}}lc|>{\centering\arraybackslash}X>{\centering\arraybackslash}X>{\centering\arraybackslash}X>{\centering\arraybackslash}X>{\centering\arraybackslash}X@{\hspace{\tabcolsep}}}
\toprule
\multirow{2}{*}{\textbf{Method}} & \multirow{2}{*}{\textbf{LM visual tokens}}
& \multicolumn{5}{c}{\textbf{Relative analytical compute (\%) $\downarrow$}} \\
\cmidrule(l){3-7}
& & \textbf{2B}
& \textbf{9B}
& \textbf{27B}
& \textbf{InternVL}
& \textbf{Llama} \\
\midrule
\rowcolor{GranulonStripe} \DINO-only & 196 & 100 & 100 & 100 & 100 & 100 \\
\SIGLIP-only & 576 & 252 & 248 & 247 & 248 & 248 \\
\rowcolor{GranulonStripe} Average efficiency & \underline{386} & \underline{198} & \underline{179} & \underline{175} & \underline{179} & \underline{179} \\
\rowcolor{GranulonOurs} \textbf{\modelname{}} & \PAIQOurs{\textbf{196}} & \PAIQOurs{\textbf{133}} & \PAIQOurs{\textbf{108}} & \PAIQOurs{\textbf{103}} & \PAIQOurs{\textbf{108}} & \PAIQOurs{\textbf{108}} \\
\bottomrule
\end{tabularx}
\end{table}

\subsection{Consistency across samples and evaluation protocols}

\begin{figure}[!t]
  \centering
  \includegraphics[width=\textwidth]{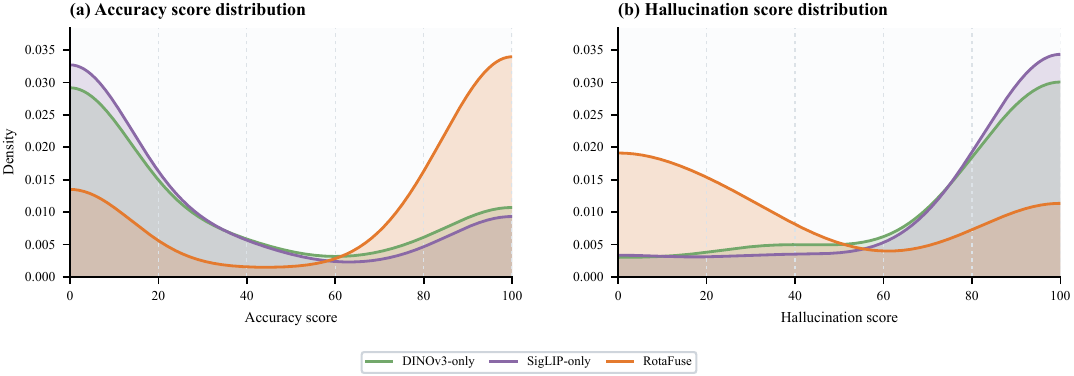}\vspace{-3pt}
  \caption{Qwen3.5-2B sample-level score distributions pooled over four equally sized datasets ($n=4{,}000$ per method). \modelname{} shifts Accuracy upward and Hallucination downward relative to both single-encoder interfaces.}
  \label{fig:score_distributions}
\end{figure}

\paragraph{Sample-level score distributions.}
On Qwen3.5-2B, 67.85\% of responses pooled over the four datasets receive Accuracy scores of at least 80, compared with under 21\% for either single-encoder interface; the Hallucination distribution also shifts toward lower scores (Figure~\ref{fig:score_distributions}).

\FloatBarrier 

\paragraph{Paired uncertainty.}
Against the single encoder with higher mean Accuracy in each backbone--dataset setting, both score improvements have paired 95\% bootstrap intervals above zero in 18 of 20 settings. The two exceptions are the InternVL VQA datasets (Appendix~\ref{app:bootstrap}). These intervals quantify record-level uncertainty conditional on the trained checkpoints and recorded judge scores.

\paragraph{Closed-ended VQA.}
To complement the judge-based results, we evaluate Qwen3.5-2B on the same VQA slices under benchmark-aligned multiple-choice protocols. \modelname{} achieves 42.1\% on SEED-Bench and 43.6\% on A-OKVQA, exceeding the strongest included comparator by 1.5 and 1.3 percentage points, respectively (Appendix~\ref{app:mcq_evaluation}, Table~\ref{tab:closed_ended_mcq}).

\subsection{Component ablations and residual diagnostics}

\begin{figure}[H]
  \centering
  \begin{subfigure}[t]{0.49\textwidth}
    \centering
    \includegraphics[width=\linewidth]{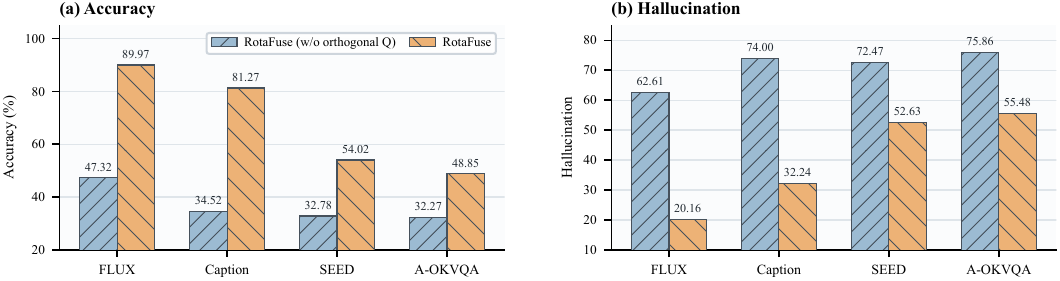}
    \caption{Fixed-$\mQ$ ablation.}
    \label{fig:qwen2b_ablation}
  \end{subfigure}\hfill
  \begin{subfigure}[t]{0.49\textwidth}
    \centering
    \includegraphics[width=\linewidth]{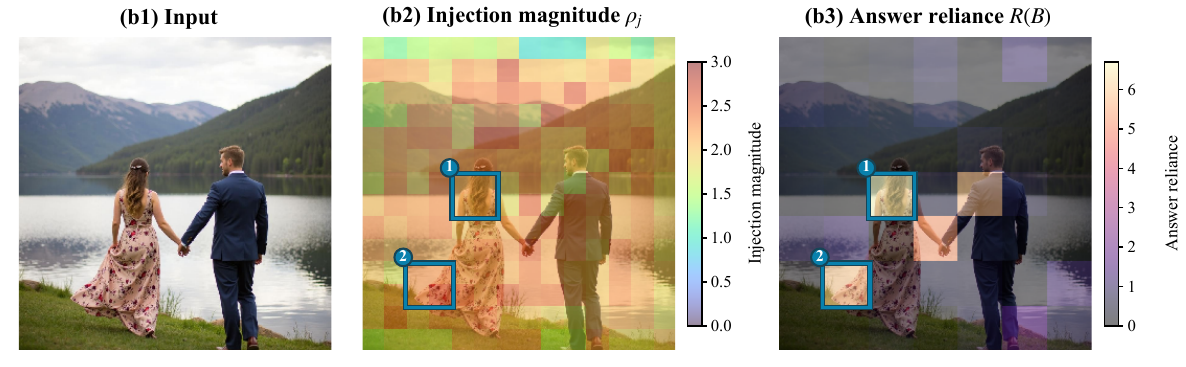}
    \caption{Qualitative FLUX case.}
    \label{fig:reliance_trace}
  \end{subfigure}
  \caption{Qwen3.5-2B analyses. (a) Learned rotation versus fixed-$\mQ=\mI$ interpolation, using judge scores on $[0,100]$ (Appendix~\ref{app:description_prompts} documents the description-template distinction). (b) A FLUX example showing the input (b1), relative update magnitude $\rho_j$ (b2), and fixed-answer likelihood sensitivity $R(B)$ (b3). Regions 1 and 2 mark the woman's hair and floral dress at corresponding locations in both maps. The sensitivity map clips negative $R(B)$ values; the highlighted regions illustrate answer-level rather than phrase-level sensitivity.}
  \label{fig:ablation_reliance}
\end{figure}

\paragraph{Learning the residual rotation.}
We further compare the learned rotation with direct interpolation ($\mQ=\mI$ throughout training), retaining the same cross-encoder allocation rule. On SEED and A-OKVQA, learned rotation improves Accuracy by 21.24 and 16.58 score points, respectively (Figure~\ref{fig:ablation_reliance}\subref{fig:qwen2b_ablation}; Appendix~\ref{app:qwen2b_final}). The archived difference in task-macro Accuracy is 31.81 points, with the qualification that the fixed-rotation ablation uses a separate image-description template (Appendix~\ref{app:description_prompts}). Given the function-class equivalence in Lemma~\ref{lem:equivalence}, the VQA gains support explicit rotation as a learning parameterization under the reported training recipe, without requiring greater expressive power.

\paragraph{Sensitivity to local residual updates.}
In the FLUX example of Figure~\ref{fig:ablation_reliance}\subref{fig:reliance_trace}, removing updates near the woman's hair and floral dress produces large decreases in the likelihood of the original answer, whereas large relative updates occur across a broader region. This case illustrates the distinction between update magnitude and a trained model's sensitivity to residual removal. We detail the intervention and its interpretation in Appendices~\ref{app:diagnostics} and~\ref{app:case_record}.

\paragraph{Joint allocation and orthogonal constraint.}
We evaluate the two core design choices independently on Qwen3.5-2B: replacing joint source allocation with row-wise softmax, and replacing the orthogonal rotation with a learnable unconstrained matrix.

\begin{table}[H]
\centering
\setlength{\belowcaptionskip}{4pt}
\caption{Independent ablations of the orthogonal constraint and Sinkhorn-style normalization on Qwen3.5-2B. Accuracy ($\uparrow$) and Hallucination ($\downarrow$) are mean judge scores on $[0,100]$.}
\label{tab:independent_ablations}
\small
\setlength{\tabcolsep}{4pt}
\renewcommand{\arraystretch}{1.12}
\begin{tabularx}{\linewidth}{l*{4}{>{\centering\arraybackslash}X}}
\toprule
\textbf{Method}
& \textbf{FLUX (Acc./Hall.)}
& \textbf{Caption (Acc./Hall.)}
& \textbf{SEED (Acc./Hall.)}
& \textbf{A-OKVQA (Acc./Hall.)} \\
\midrule
\rowcolor{PAIQGray}
w/o orthogonal constraint
& 87.19 / 21.68 & 76.77 / 34.33 & 52.09 / 54.85 & 47.74 / 57.29 \\
\rowcolor{white}
w/o Sinkhorn-style normalization
& 87.40 / 21.05 & 78.60 / 31.65 & 50.42 / 55.43 & 46.79 / 57.87 \\
\rowcolor{PAIQYellow}
\textbf{\modelname{} (Ours)}
& \textbf{89.97 / 20.16} & \textbf{81.27 / 32.24}
& \textbf{54.02 / 52.63} & \textbf{48.85 / 55.48} \\
\bottomrule
\end{tabularx}
\end{table}
In Table~\ref{tab:independent_ablations}, the orthogonal constraint improves Accuracy by 1.11--4.50 score points over the unconstrained transform, while joint allocation improves it by 2.06--3.60 points over independent matching, with gains on every dataset. Hallucination is lower with the full model in seven of the eight comparisons. Together, the ablations support the use of joint allocation and orthogonal residual updates (Sections~\ref{sec:semantic_transport} and~\ref{sec:orthogonal_injection}).

%% file: sections/conclusion.tex
\section{Conclusion \& Limitations}
\label{sec:conclusion}

We introduced \modelname{}, a compact visual interface that separates cross-encoder feature aggregation from geometry-constrained residual updates. By enriching spatially structured features through learnable orthogonal rotation, \modelname{} integrates complementary visual representations within the original base-token budget. Our analysis characterizes local separability under shared aggregates and identifies explicit rotation as a geometry-constrained training parameterization. Across language backbones and visual-language tasks, the resulting 196-token interface consistently improves response quality over matched single-encoder interfaces, with component ablations supporting both joint feature allocation and orthogonal residual updates.

\textbf{Limitations.} Our study focuses on one pair of complementary visual encoders and fixed-resolution image inputs. Extending geometry-constrained enrichment to additional encoder families, higher-resolution representations, and video remains an important direction for future work.

%% file: appendices/appendix.tex
\section*{Appendix Contents}
\label{app:contents}
\begingroup
\hypersetup{linkcolor=black}
\setlength{\parindent}{0pt}
\setlength{\parskip}{2pt}

\AppendixContentsLine{app:related_work}
{Related Work}

\AppendixContentsLine{app:theory}
{Supplementary Method Analysis}

\AppendixContentsLine{app:method_details}
{Implementation and Internal Diagnostics}

\AppendixContentsLine{app:experiment_settings}
{Experimental Settings and Evaluation Scope}

\AppendixContentsLine{app:additional_results}
{Additional Results and Statistical Analysis}

\AppendixContentsLine{app:compute}
{Analytical Compute Accounting}

\AppendixContentsLine{app:case_record}
{Qualitative Diagnostics and Case Interpretation}

\AppendixContentsLine{app:extended-cases}
{Extended Qualitative Cases with Decoder Attention}

\AppendixContentsLine{app:prompts}
{Evaluator Prompts}
\endgroup

\input{appendices/related_work.tex}

\section{Supplementary Method Analysis}
\label{sec:supplement}
\label{app:theory}

This appendix develops the properties of joint source allocation and orthogonal residual enrichment introduced in Section~\ref{sec:method}. We first distinguish the implemented allocation rule from its idealized transport interpretation (Appendix~\ref{app:matching}), establish the geometry of residual updates (Appendix~\ref{app:geometry}), and prove the separation result in Lemma~\ref{lem:separation} (Appendix~\ref{app:separation}). Finally, we characterize the function-class equivalence between residual rotation and direct interpolation when the content and matching projections are freely trainable (Appendix~\ref{app:equivalence}). Throughout, feature matrices store tokens as rows, individual features are column vectors, and geometric identities are stated in exact arithmetic.

\subsection{Matching normalization and source allocation}
\label{app:matching}

\paragraph{Finite-step computation.}
The content cost in~\PAIQeqref{eq:cost} defines the log-kernel
\begin{equation}
K_{ji}=-C_{ji}/\varepsilon,\qquad \varepsilon=0.05.
\label{eq:sinkhorn_kernel}
\end{equation}
We initialize the log-scaling factors $\alpha^{(0)}=\beta^{(0)}=0$ and perform $L=5$ iterations of alternating row and column scaling:
\begin{equation}
\begin{aligned}
\alpha_j^{(t)}&=\operatorname{clip}_{[-30,30]}\!\left[-\operatorname{LSE}_i\!\left(K_{ji}+\PAIQmatch{\beta_i^{(t-1)}}\right)\right],\\
\PAIQmatch{\beta_i^{(t)}}&=\operatorname{clip}_{[-30,30]}\!\left[-\operatorname{LSE}_j\!\left(K_{ji}+\alpha_j^{(t)}\right)\right].
\end{aligned}
\label{eq:sinkhorn_updates}
\end{equation}
\begingroup
\makeatletter
\protected@edef\@currentlabel{\p@equation\theequation}
\label{eq:sinkhorn_main}
\makeatother
\endgroup
Here $\operatorname{LSE}_i(a_i)=\log\sum_i\exp(a_i)$. Each column update uses the newly computed row factor, giving five row updates and five column updates in total. The forward pass then constructs the allocation weights in~\PAIQeqref{eq:allocation_weights} and the aggregates in~\PAIQeqref{eq:barycentric}:
\begin{equation}
\begin{aligned}
T_{ji}&=\exp\!\left[\min\!\left(K_{ji}+\alpha_j^{(L)}+\PAIQmatch{\beta_i^{(L)}},0\right)\right],\\
q_j&=\sum_iT_{ji},\qquad
\PAIQmatch{\pi_{ji}}=\frac{T_{ji}}{q_j+\delta},\qquad \delta=10^{-6},\\
\bar{\mS}&=\PAIQmatch{\boldsymbol\Pi}\mS.
\end{aligned}
\label{eq:app_aggregation}
\end{equation}
The normalized matching keys determine the allocation, while the projected content features $\mS$ supply its values. The iteration count is fixed, and the scaling procedure is differentiated through the answer-generation objective in~\PAIQeqref{eq:training}.

The stabilizer $\delta$ introduces a small departure from exact row normalization when $q_j\gg\delta$. For $q_j>0$, define a normalized weight $\widetilde\pi_{ji}$ and a row-dependent scale $\eta_j$:
\begin{equation}
\begin{aligned}
\widetilde\pi_{ji}&=\frac{T_{ji}}{q_j},\qquad
\eta_j=\frac{q_j}{q_j+\delta},\\
\sum_i\pi_{ji}&=\eta_j,\qquad
\bar{\vs}_j=\eta_j\sum_i\widetilde\pi_{ji}\vs_i.
\end{aligned}
\label{eq:row_mass}
\end{equation}
Thus the implemented coefficients have row sums below one, and the aggregate is a scaled barycentric average. The idealized case $\delta=0$ recovers a conventional row-conditioned barycentric average.

\paragraph{Why the queries are coupled.}
To isolate the role of column scaling, consider the idealized computation obtained by removing both clipping operations and setting $\delta=0$. After $L$ iterations, the resulting allocation matrix factorizes as
\begin{equation}
T^{\circ}_{ji}=\exp(\alpha_j^{(L)})\exp\!\left(K_{ji}+\PAIQmatch{\beta_i^{(L)}}\right).
\label{eq:bias_factorization}
\end{equation}
The row factor cancels in the final row conditioning, giving
\begin{equation}
\PAIQmatch{\pi^{\circ}_{ji}}
=\frac{\exp(K_{ji}+\beta_i^{(L)})}{\sum_k\exp(K_{jk}+\beta_k^{(L)})}
=\operatorname{softmax}_i\!\left(-C_{ji}/\varepsilon+\PAIQmatch{\beta_i^{(L)}}\right).
\label{eq:bias_proof}
\end{equation}
This establishes~\PAIQeqref{eq:source_bias}. Each column update depends on all queries through the log-sum-exp over $j$, and each subsequent row update depends on the resulting column factors. Although the final row factor cancels in~\PAIQeqref{eq:bias_proof}, the preceding row updates influence the shared bias through this alternation. The bias $\beta_i^{(L)}$ is therefore image-dependent and couples source allocation across queries. It is computed by the scaling procedure rather than learned as a separate parameter. This identity describes the idealized weights; the actual forward pass uses~\PAIQeqref{eq:app_aggregation}.

\paragraph{Relation to uniform-marginal transport.}
For finite $\mC$ and $\varepsilon>0$, entropy-regularized transport with uniform probability marginals is defined by~\citep{cuturi2013sinkhorn}
\begin{equation}
\begin{aligned}
\min_{\Gamma\geq0}\quad&
\langle\Gamma,\mC\rangle+\varepsilon\sum_{j,i}\Gamma_{ji}(\log\Gamma_{ji}-1),\\
\text{subject to}\quad&
\Gamma\mathbf1_m=\tfrac1n\mathbf1_n,\qquad
\Gamma^\top\mathbf1_n=\tfrac1m\mathbf1_m.
\end{aligned}
\label{eq:ot_reference}
\end{equation}
Both marginals have unit total mass, including for rectangular grids. The solution is obtained by positive row and column scaling of $\exp(-\mC/\varepsilon)$.

The unclipped unit-sum updates in~\PAIQeqref{eq:sinkhorn_updates} are related to these probability-marginal updates by a global scale. Let $\mathcal R$ and $\mathcal C$ normalize a positive matrix to unit row sums and unit column sums, respectively. For $c>0$, $\mathcal R(cM)=\mathcal R(M)$ and $\mathcal C(cM)=\mathcal C(M)$, so probability-marginal updates are $(1/n)\mathcal R$ and $(1/m)\mathcal C$. Starting from the same kernel, the first column update in the probability-scaled scheme yields $1/m$ times the corresponding unit-scaled matrix. The next row update cancels this factor, and the next column update restores it. By induction, the two matrices differ by a factor of $1/m$ after each complete iteration. Final row conditioning cancels that factor when $\delta=0$.

At convergence to the transport solution $\Gamma^\star$, the row-conditioned coefficients satisfy
\begin{equation}
\boldsymbol\Pi^\star=n\Gamma^\star,\qquad
\boldsymbol\Pi^\star\mathbf1_m=\mathbf1_n,\qquad
(\boldsymbol\Pi^\star)^\top\mathbf1_n=\tfrac nm\mathbf1_m.
\label{eq:ideal_marginals}
\end{equation}
The column marginal balances total source use while permitting a source to contribute to multiple queries. Balancing can discourage concentration on a small set of sources, but can also increase the weights of less relevant features. The task-level effect is therefore empirical. The finite, clipped forward pass with $\delta>0$ is a joint allocation rule motivated by this scaling construction; it need not satisfy the exact marginals in~\PAIQeqref{eq:ideal_marginals}.

\subsection{Cayley orthogonality and residual geometry}
\label{app:geometry}

The residuals in Section~\ref{sec:orthogonal_injection} are
\begin{equation}
\vr_j=\bar{\vs}_j-\vd_j,\qquad \mR=\bar{\mS}-\mD.
\label{eq:residual}
\end{equation}
We establish that the Cayley parameterization in~\PAIQeqref{eq:cayley} is well-defined for every trainable $W_Q$ in exact arithmetic and derive the update geometry in~\PAIQeqref{eq:residual_gram}.

\begin{APXProof}{Cayley orthogonality and residual geometry}
Write $B=\mA/2$, so $B^\top=-B$. For every nonzero $x$, skew-symmetry gives $x^\top Bx=0$ and hence
\begin{equation}
\|(\mI-B)x\|_2^2=\|x\|_2^2+\|Bx\|_2^2>0.
\label{eq:cayley_invertible}
\end{equation}
Thus $M=\mI-B$ is invertible; the same argument applies to $N=\mI+B$. Since $M^\top=N$ and $M,N$ commute, the Cayley transform satisfies
\begin{equation}
\PAIQrotate{\mQ^\top\mQ}
=(M^{-1}N)^\top M^{-1}N
=MN^{-1}M^{-1}N=\mI.
\label{eq:cayley_proof}
\end{equation}
The Cayley transform along $t\mA$, $0\leq t\leq1$, is continuous and orthogonal and equals $\mI$ at $t=0$. Its determinant therefore remains $+1$, establishing that the transform is a rotation in feature space.

For the update matrix $\boldsymbol\Delta=\mZ-\mD=\lambda\mR\mQ^\top$, orthogonality yields
\begin{equation}
\boldsymbol\Delta\boldsymbol\Delta^\top
=\lambda^2\mR\mQ^\top\mQ\mR^\top
=\lambda^2\mR\mR^\top.
\label{eq:gram_proof}
\end{equation}
Let $\Delta_j=\vz_j-\vd_j$ and take $\lambda>0$. The diagonal and off-diagonal entries of this identity give
\begin{equation}
\|\Delta_j\|_2=\lambda\|\vr_j\|_2,\qquad
\langle\Delta_j,\Delta_k\rangle=\lambda^2\langle\vr_j,\vr_k\rangle.
\label{eq:norm}
\end{equation}
It also follows that
\begin{equation}
\|\Delta_j-\Delta_k\|_2=\lambda\|\vr_j-\vr_k\|_2.
\label{eq:residual_pair_geometry}
\end{equation}
Dividing the inner-product identity by the corresponding norms proves angle preservation for nonzero residuals.
\end{APXProof}

\paragraph{Scope of residual geometry.}
The Gram identity constrains the updates for a given residual set. Adding them to the base gives
\begin{equation}
\|\vz_j\|_2^2=\|\vd_j\|_2^2+\lambda^2\|\vr_j\|_2^2
+2\lambda\vd_j^\top\PAIQrotate{\mQ}\vr_j.
\label{eq:output_cross_term}
\end{equation}
The cross term depends on the update direction, and the trainable content projections can change the scales of both the bases and the residuals. Consequently, $\lambda$ controls the scale of a given residual update, rather than bounding its size relative to the base or specifying the proportion of semantic and spatial information in the fused representation.

\subsection{Patch differences: proof and extensions}
\label{app:separation}

\paragraph{Rotation-aware separation.}
\label{app:rotation_bound}
The central observation is that the same rotation governing the residual geometry also determines how the base difference contributes to the fused distance. We prove the shared-aggregate identity and its extension to arbitrary aggregates in a common argument.

\begin{APXProof}{Proof of Lemma~\ref{lem:separation}}
Fix the projected base and aggregate features, let $x_{jk}=\vd_j-\vd_k$ and $e_{jk}=\bar{\vs}_j-\bar{\vs}_k$, and take $0<\lambda<1$ with $\mQ^\top\mQ=\mI$. Subtracting the updates gives~\PAIQeqref{eq:pair_decomposition}.

For any $x$, orthogonality implies $\|\mQ x\|_2=\|x\|_2$. Expanding the squared norm of the transformed base difference yields
\begin{align}
\|(\mI-\lambda\mQ)x\|_2^2
&=(1+\lambda^2)\|x\|_2^2-2\lambda x^\top\mQ x\nonumber\\
&=(1-\lambda)^2\|x\|_2^2
+\PAIQrotate{\lambda\|(\mI-\mQ)x\|_2^2}.
\label{eq:contrast_proof}
\end{align}
By the definition in~\PAIQeqref{eq:rotation_margin}, $g_Q(x)=\|(\mI-\lambda\mQ)x\|_2$.

\textbf{(i) Shared aggregates.}
If $e_{jk}=0$, the second term in~\PAIQeqref{eq:pair_decomposition} vanishes. Therefore $\|\vz_j-\vz_k\|_2^2=g_Q(x_{jk})^2$, proving~\PAIQeqref{eq:contrast_identity}.

\textbf{(ii) General aggregates.}
For arbitrary $e_{jk}$, the reverse triangle inequality applied to~\PAIQeqref{eq:pair_decomposition} gives
\begin{equation}
\|\vz_j-\vz_k\|_2
\geq \|(\mI-\lambda\mQ)x_{jk}\|_2-\lambda\|\mQ e_{jk}\|_2
=g_Q(x_{jk})-\lambda\|e_{jk}\|_2.
\label{eq:rotation_bound_proof}
\end{equation}
Taking the positive part, since the left-hand side is nonnegative, proves~\PAIQeqref{eq:rotation_aware_bound}.
\end{APXProof}

The identity~\PAIQeqref{eq:contrast_proof} separates the distance retained by direct interpolation from the contribution of rotation. If $\mQ x\ne x$, this contribution is strictly positive. For arbitrary aggregate differences, the lower bound guarantees distinct outputs whenever $x_{jk}\ne0$ and $\|e_{jk}\|_2<g_Q(x_{jk})/\lambda$. It characterizes a sufficient condition for fixed features and a given rotation; it does not require or imply that training realizes any particular rotation.

\paragraph{Rotation-independent bounds.}
Applying the triangle and reverse triangle inequalities to $(\mI-\lambda\mQ)x$ yields
\begin{equation}
(1-\lambda)\|x\|_2
\leq\|(\mI-\lambda\mQ)x\|_2
\leq(1+\lambda)\|x\|_2.
\label{eq:base_singular_bound}
\end{equation}
Combining this with~\PAIQeqref{eq:pair_decomposition}, we obtain the following two-sided bounds for arbitrary aggregates:
\begin{equation}
\begin{aligned}
\big[(1-\lambda)\|x_{jk}\|_2-\lambda\|e_{jk}\|_2\big]_+
&\leq\|\vz_j-\vz_k\|_2\\
&\leq(1+\lambda)\|x_{jk}\|_2+\lambda\|e_{jk}\|_2.
\end{aligned}
\label{eq:pair_bound}
\end{equation}
\begingroup
\makeatletter
\protected@edef\@currentlabel{\p@equation\theequation}
\label{eq:pair_lower}
\makeatother
\endgroup
The lower bound in~\PAIQeqref{eq:pair_bound} is independent of the learned rotation and never exceeds the rotation-aware bound in~\PAIQeqref{eq:rotation_aware_bound}. Under a shared aggregate, the bounds simplify to
\begin{equation}
(1-\lambda)\|\vd_j-\vd_k\|_2
\leq\|\vz_j-\vz_k\|_2
\leq(1+\lambda)\|\vd_j-\vd_k\|_2.
\label{eq:shared_semantics}
\end{equation}
More generally, for $x_{jk}\ne0$ and $\|e_{jk}\|_2\leq\kappa\|x_{jk}\|_2$ with $\kappa<(1-\lambda)/\lambda$, the lower bound is $(1-\lambda-\lambda\kappa)\|x_{jk}\|_2>0$.

\paragraph{Connection to feature allocation.}
The aggregate difference can also be expressed through the allocation weights. Let $M_s=\max_i\|\vs_i\|_2$ and $\boldsymbol\pi_j$ denote the $j$th row of $\boldsymbol\Pi$. By~\PAIQeqref{eq:barycentric} and the triangle inequality,
\begin{equation}
\|e_{jk}\|_2
=\Big\|\sum_i(\pi_{ji}-\pi_{ki})\vs_i\Big\|_2
\leq M_s\|\boldsymbol\pi_j-\boldsymbol\pi_k\|_1.
\label{eq:weight_to_feature}
\end{equation}
Substitution into~\PAIQeqref{eq:pair_bound} gives
\begin{equation}
\|\vz_j-\vz_k\|_2
\geq\Big[(1-\lambda)\|\vd_j-\vd_k\|_2
-\lambda M_s\|\boldsymbol\pi_j-\boldsymbol\pi_k\|_1\Big]_+.
\label{eq:matching_separation}
\end{equation}
This relates the separation condition directly to the similarity of the allocation rows. The argument uses linear aggregation and thus also applies to the stabilized weights of the implemented forward pass.

\paragraph{Role and scope of orthogonality.}
With fixed distinct bases and a shared aggregate, direct interpolation retains a fraction $1-\lambda$ of the original feature distance, and an orthogonal rotation can increase the distance through the second term of~\PAIQeqref{eq:contrast_proof}. In contrast, the unrestricted residual map $G=\lambda^{-1}\mI$ gives
\begin{equation}
\vd_j+\lambda G(\bar{\vs}_j-\vd_j)=\bar{\vs}_j,
\label{eq:unconstrained_example}
\end{equation}
eliminating all base differences under shared aggregates. Orthogonality excludes this specific collapse for $0<\lambda<1$. When aggregates differ, cancellation between the two terms in~\PAIQeqref{eq:pair_decomposition} remains possible. In particular, the rotation-aware lower bound exceeding the interpolation lower bound does not imply a larger realized distance for arbitrary aggregate differences. These statements concern distances between projected features: $1-\lambda=0.4$ in our implementation is a distance-bound coefficient, not a spatial-information retention rate. The unrestricted-map example likewise establishes a possibility within that parameterization, not a prediction of the outcome of unconstrained training.

\subsection{Equivalent parameterizations with free projections}
\label{app:equivalence}

The preceding analysis isolates the geometry of the residual update at fixed intermediate features. We now consider the function class of the full trainable fusion module, whose affine content projections and linear matching maps can adapt jointly with the rotation.

\begin{paiqtheorybox}
\begin{lemma}[Function-class equivalence to direct interpolation]
\label{lem:equivalence}
Assume unrestricted affine content projections and unrestricted linear matching maps $W_d,W_s$, with no intervening activation or normalization between them that prevents absorbing linear transformations into the projections. For a fixed $0<\lambda<1$ and the same matching and aggregation rule, \modelname{} and the model with $\mQ=\mI$ represent the same input--output function class in exact arithmetic.
\end{lemma}
\end{paiqtheorybox}

\begin{APXProof}{Proof of Lemma~\ref{lem:equivalence}}
For any parameter setting of \modelname{}, define
\begin{equation}
B_d=\frac{\mI-\lambda\mQ}{1-\lambda}.
\label{eq:fold_base}
\end{equation}
By~\PAIQeqref{eq:base_singular_bound}, $\mI-\lambda\mQ$ has a strictly positive lower norm bound for nonzero $x$, making $B_d$ invertible. Reparameterize the content projections and matching maps of the direct-interpolation model as
\begin{equation}
\begin{aligned}
P'_d&=B_dP_d, & b'_d&=B_db_d, & W'_d&=W_dB_d^{-1},\\
P'_s&=\mQ P_s, & b'_s&=\mQ b_s, & W'_s&=W_s\mQ^\top.
\end{aligned}
\label{eq:fold_parameters}
\end{equation}
The reparameterized features are $\vd'_j=B_d\vd_j$ and $\vs'_i=\mQ\vs_i$. Their matching vectors before normalization are unchanged:
\[
W'_d\vd'_j=W_d\vd_j,\qquad W'_s\vs'_i=W_s\vs_i.
\]
Consequently, the normalized queries and keys, costs, and all allocation coefficients are identical, including under finite iterations, clipping, and the stabilizer. Since aggregation is linear in the content features, $\bar{\vs}'_j=\mQ\bar{\vs}_j$. The direct-interpolation output then satisfies
\begin{equation}
(1-\lambda)\vd'_j+\lambda\bar{\vs}'_j
=(\mI-\lambda\mQ)\vd_j+\lambda\mQ\bar{\vs}_j=\vz_j.
\label{eq:fold_output}
\end{equation}
The reverse inclusion follows by setting $W_Q=0$ in \modelname{}, which gives $\mQ=\mI$. Both constructions supply identical fused visual inputs to the same frozen language model and therefore yield identical input--output functions.
\end{APXProof}

\paragraph{Implications for the parameterization.}
The equivalence shows that the explicit orthogonal layer specifies a geometry-constrained training parameterization rather than enlarging the function class under these assumptions. The construction changes the intermediate base and source features, so it does not contradict the fixed-feature separation result in Lemma~\ref{lem:separation}. Nor does equivalence of representable functions establish identical initialization, gradients, optimization trajectories, or performance under a finite training budget. Parameter folding is an exact-arithmetic identity and need not give bitwise-equivalent execution in floating-point arithmetic.

\section{Implementation and Internal Diagnostics}
\label{app:method_details}

\subsection{Feature extraction and execution}
\label{app:implementation}

\paragraph{Visual features.}
For a given image $I$, the frozen encoders produce
\begin{equation}
\mX_d=E_d(I)\in\mathbb R^{196\times1024},\qquad
\mX_s=E_s(I)\in\mathbb R^{576\times1024},
\label{eq:raw_features}
\end{equation}
where each $E$ includes image preprocessing and token selection. \DINO{} processes a $224\times224$ RGB input with patch size 16. We extract its final normalized patch features, discarding one CLS token and four register tokens. \SIGLIP{} processes a $384\times384$ input and supplies the final $24\times24$ patch features. For \DINO{}, the channel means and standard deviations are $(0.485,0.456,0.406)$ and $(0.229,0.224,0.225)$, respectively; for \SIGLIP{}, both are $(0.5,0.5,0.5)$. The two encoders differ in both pretraining and input resolution. Their token counts specify the feature grids, rather than determining their semantic or spatial capabilities.

\paragraph{Projection and initialization.}
The two content projections in~\PAIQeqref{eq:content_project} are independent affine maps with biases and no subsequent activation or normalization. The bias-free matching matrices $W_d,W_s$ are initialized to identity, and $W_Q$ is initialized to zero. Thus $\mQ=\mI$ at initialization and the fused sequence is $\mZ=0.4\mD+0.6\bar{\mS}$. Throughout training, the same $\mQ$ is shared across all positions and images. Both the matching and fusion computations depend only on the image and the learned parameters; the language model uses the resulting visual sequence in the context of its text instruction.

\paragraph{Numerical execution.}
The matching linear maps are applied under the outer precision context, before their outputs are converted to float32. The subsequent normalization, log-domain scaling, aggregation, and Cayley solve use float32. We obtain the rotation by solving the linear system
\begin{equation}
(\mI-\mA/2)\mQ=\mI+\mA/2,
\label{eq:cayley_solve}
\end{equation}
which implements~\PAIQeqref{eq:cayley} without explicitly forming a matrix inverse. The fusion computation $\mD+\lambda(\bar{\mS}-\mD)\mQ^\top$ is followed by a cast back to the input-feature precision. The exact-arithmetic identities in Appendix~\ref{app:theory} are consequently subject to numerical tolerances in the implementation. The reported execution reconstructs the dense Cayley system for each image, without assuming caching or parameter folding.

\paragraph{Optimization and language input.}
The $196$ fused vectors replace the image-placeholder embeddings in the language-model input. Both visual encoders and the language model, including its output head, are frozen, while $\theta=\{P_d,b_d,P_s,b_s,W_d,W_s,W_Q\}$ is optimized through~\PAIQeqref{eq:training}. The dependence of $p_\theta$ on $\theta$ is through $\mZ(I)$ alone. The input prompt is followed by the target answer and EOS; labels for the prompt, image placeholders, and padding are masked, leaving only the answer tokens and EOS in the loss. Gradients propagate through the frozen language model to the content projections, residual transform, aggregation, and unrolled normalization steps. The answer-generation objective supplies all training supervision, with no auxiliary correspondence, contrastive, transport, or orthogonality loss.

\subsection{Parameter counts and computational scope}
\label{app:parameters}

The two affine content projections, two linear matching maps, and dense matrix $W_Q$ contain
\begin{equation}
N_{\mathrm{train}}=2(1024d+d)+3d^2
\label{eq:parameter_count}
\end{equation}
stored trainable parameters. Table~\ref{tab:paiq_parameters} reports the count for each hidden dimension. Although the rotation depends only on the skew-symmetric part of $W_Q$, we count the full stored matrix rather than its effective degrees of freedom.
\begin{table}[t]
\centering
\caption{Trainable parameter counts derived from the layer shapes. All figures include both content projections; they exclude frozen visual and language parameters.}
\label{tab:paiq_parameters}
\normalsize
\setlength{\tabcolsep}{5pt}
\renewcommand{\arraystretch}{1.18}
\begin{tabular}{rrrrr}
\toprule
\textbf{$d$} & \textbf{Content maps} & \textbf{Matching maps} & \textbf{$W_Q$} & \textbf{Total}\\
\midrule
2048 & 4,198,400 & 8,388,608 & 4,194,304 & 16,781,312\\
4096 & 8,396,800 & 33,554,432 & 16,777,216 & 58,728,448\\
5120 & 10,496,000 & 52,428,800 & 26,214,400 & 89,139,200\\
\bottomrule
\end{tabular}
\end{table}

Excluding the frozen encoders and language model, the dense fusion computation per image scales as
\begin{equation}
\mathcal O\!\left((n+m)1024d+(2n+m)d^2+nmd+Lnm+d^3\right),\qquad L=5.
\label{eq:complexity}
\end{equation}
The terms account for the content projections, matching and residual maps, pairwise scores and value aggregation, scaling iterations, and dense Cayley solve, respectively. The fixed $196$-token sequence controls the visual prefix length presented to the language model. End-to-end computation, latency, and memory also depend on the costs of feature extraction and fusion. Appendix~\ref{app:compute} specifies the analytical compute proxy used in the experiments.

\subsection{Aggregation and residual diagnostics}
\label{app:diagnostics}

We examine the trained visual interface at three levels: how source features are allocated, how strongly the projected base features are updated, and how the language model's likelihood of a fixed answer responds to local residual removal. The diagnostics are defined on the forward tensors $\mD,\mS,\boldsymbol\Pi,\mQ,\mZ$ of a fixed model and are used only for post-training analysis.

\paragraph{Source concentration.}
For each output position $j$, the row $\boldsymbol\pi_j=(\pi_{j1},\ldots,\pi_{jm})$ specifies the linear mixing coefficients of contextualized \SIGLIP{} content features. We summarize their concentration by
\begin{equation}
H(\boldsymbol\pi_j)=-\sum_i\pi_{ji}\log(\pi_{ji}+\delta_H),\qquad \delta_H>0.
\label{eq:provenance_entropy}
\end{equation}
The stabilized allocation rows have sums below one (Appendix~\ref{app:matching}), and the logarithm contains a stabilizer. Accordingly, $H$ is a concentration statistic rather than exact Shannon entropy. It characterizes mixing of contextualized source features, not semantic uncertainty or attribution of an answer to input pixels.

\paragraph{Relative update magnitude.}
For any position with $\|\vd_j\|_2>0$, we measure the magnitude of its update relative to the projected base:
\begin{equation}
\rho_j=\frac{\|\vz_j-\vd_j\|_2}{\|\vd_j\|_2}
=\lambda\frac{\|\bar{\vs}_j-\vd_j\|_2}{\|\vd_j\|_2}.
\label{eq:rho}
\end{equation}
The second equality follows from the residual geometry in~\PAIQeqref{eq:residual_gram}. For fixed base and aggregate features, the magnitude is independent of the rotation direction. A $14\times14$ map of these ratios shows where the output representation receives relatively large updates. Ratios greater than one are possible, and a small base norm can produce a large ratio; at a zero base norm the ratio is undefined. The map describes update magnitudes at output positions rather than the spatial origins of the contextualized source information.

\paragraph{Fixed-answer likelihood sensitivity.}
To measure the effect of a local residual update on the language model's prediction, we partition the $14\times14$ output grid into 49 non-overlapping $2\times2$ blocks. For each block $B$, we remove the corresponding residual updates by restoring its features to the same model's projected \DINO{} bases:
\begin{equation}
\vz_j^{(-B)}=\begin{cases}
\vd_j,&j\in B,\\
\vz_j,&j\notin B.
\end{cases}
\label{eq:block_replacement}
\end{equation}
This replacement holds the matching weights, the remaining fused features, the model parameters, and the text fixed. In particular, it operates on the fused tokens, rather than masking image pixels or substituting features from an independently trained single-encoder model.

Let $y$ be the answer generated by the original model. We compare its log-likelihood under the original and modified visual sequences:
\begin{equation}
R(B)=\log p_\theta(y\mid T,\mZ)-\log p_\theta(y\mid T,\mZ^{(-B)}).
\label{eq:reliance}
\end{equation}
In both evaluations, the sequence log-likelihood is $\sum_t\log p_\theta(y_t\mid y_{<t},T,\cdot)$, computed by teacher forcing with the same answer and prefixes, without regeneration or length normalization. A positive $R(B)$ indicates that removing the block's residual updates lowers the likelihood of the fixed answer; a negative value indicates the opposite. The score measures an answer-level effect of the specified internal replacement, rather than phrase-level causality or attribution to a region of the input image.

\paragraph{Dependence on the parameterization.}
The equivalent construction in Lemma~\ref{lem:equivalence} preserves $\mZ$ and $\boldsymbol\Pi$ while potentially changing $\mD$. Thus both the relative update magnitude in~\PAIQeqref{eq:rho} and the replacement baseline in~\PAIQeqref{eq:block_replacement} depend on the intermediate feature parameterization. These diagnostics describe how a particular trained model constructs and uses its visual representation; they are not uniquely determined by its input--output function. Decoder-attention overlays constitute a separate measurement, described in Appendix~\ref{app:extended-cases}.

\section{Experimental Settings and Evaluation Scope}
\label{app:experiment_settings}

\subsection{Language backbones, training, and generation}

We attach the visual interface to five frozen language-side configurations. The backbone names refer to the language components of these models; in particular, the InternVL configuration uses its language weights and tokenizer with our visual interface in place of its original vision encoder and projector.

The reported training recipe uses 200K FLUX-Reason examples for two epochs, AdamW with a learning rate of $10^{-4}$ and zero weight decay, cosine decay with 5\% warmup, and seed 42. The archived 8B-scale runs use four-GPU distributed data parallel training, with a per-device batch size of 8 and four gradient-accumulation steps (global batch size 128). The 2B and 27B runs follow the same optimization recipe. The reported results use a single training seed.

We use greedy decoding. Generation is capped at 128 new tokens for Qwen3.5-2B/9B, InternVL3.5-8B, and Llama-3.1-8B. The 12,000 predictions in the Qwen3.5-27B archive use a limit of 512 new tokens, following an earlier diagnostic of the effects of the 128-token boundary. Comparisons between interfaces at a given scale therefore use a common generation limit, while comparisons across scales also reflect this difference in the decoding budget. The realized response lengths may vary within each setting.

\paragraph{Single-encoder interfaces and fusion comparators.}
The \DINO{}-only and \SIGLIP{}-only interfaces use independently trained projections and visual sequences of 196 and 576 tokens, respectively. The Granulon-style comparator uses fixed-$K$ clustering in the pairwise cosine-relation space of projected \DINO{} features. With $K=10$, appending the cluster means to the 196 patch features yields 206 visual tokens. We train only its two-layer multimodal projector (6,295,552 parameters) for two epochs and 3,126 optimizer steps, using four distributed workers and a global batch size of 128. This adaptation fixes the cluster count in place of the original text-conditioned granularity controller.

\paragraph{Fixed-rotation comparator.}
To examine the role of explicit residual rotation, the fixed-$\mQ=\mI$ comparator retains the two-stream matching and aggregation rule and uses direct interpolation throughout training. This comparator differs from the learnable unconstrained residual transform in Table~\ref{tab:independent_ablations}: the former fixes the transform to identity, whereas the latter learns a matrix without the orthogonal constraint.

\subsection{Evaluation slices}
\label{app:data_slices}

We use fixed evaluation slices of 1,000 records per cell, aligned across methods. Table~\ref{tab:paiq_data_slices} lists their inclusive source-index ranges. FLUX shares its source dataset with training and is used as an in-domain image-description diagnostic; overlap of individual records with the optimization set is unverified. Caption, SEED, and A-OKVQA provide cross-source evaluations, with potential pretraining overlap remaining unassessed.
\begin{table}[!htbp]
\centering
\caption{Source slices and the evaluation information used. FLUX is a source-overlap description diagnostic, not a held-out test of reasoning-answer correctness.}
\label{tab:paiq_data_slices}
\normalsize
\setlength{\tabcolsep}{4pt}
\renewcommand{\arraystretch}{1.18}
\begin{tabularx}{\linewidth}{l>{\raggedright\arraybackslash}X>{\raggedright\arraybackslash}X}
\toprule
\textbf{Dataset} & \textbf{Slice} & \textbf{Scoring context}\\
\midrule
FLUX-Reason & Training-source records 500--1499 & Image and generated description\\
Caption & CC3M training-slice rows 500--1499 & Image and generated description\\
SEED-Bench & Test records 500--1499 & Image, question, options, reference, response\\
A-OKVQA & Validation rows 500--1144, then 0--354 & Image, question, options, reference, response\\
\bottomrule
\end{tabularx}
\end{table}
The Caption slice comes from \path{strict_original_cc3m/chat.json}. Dataset identities record provenance and are withheld from the evaluator. On FLUX, the image and generated description define the evaluation task, so the resulting scores characterize description quality rather than correctness on the source reasoning questions.

\subsection{Scoring protocol and interpretation}
\label{app:evaluation}

Each prediction receives two independently elicited integer scores in $[0,100]$. \emph{Accuracy} measures task-appropriate response correctness, and \emph{Hallucination} measures the severity of unsupported or contradictory content. Higher Accuracy and lower Hallucination are preferred. The evaluator receives neither the other score nor its conversation history; the two metrics are independently defined and need not sum to 100. We interpret them as judge scores rather than percentages of correct answers or hallucination frequencies.

The two task families use different evidence for assessing a response (Table~\ref{tab:eval_families}): image descriptions are evaluated against visible image content, while VQA responses are evaluated against an image, a specific question, and its authoritative reference answer. Comparisons are made within a dataset, backbone, and evaluation version. Task-macro averages provide equal-weight descriptive summaries across tasks with different rubrics.
\begin{table}[!htbp]
\caption{Input context and score interpretation for the two evaluation families. Neither branch reports an official dataset metric.}
\label{tab:eval_families}
\centering
\normalsize
\setlength{\tabcolsep}{4pt}
\renewcommand{\arraystretch}{1.18}
\begin{tabularx}{\linewidth}{>{\raggedright\arraybackslash}p{0.23\textwidth}>{\raggedright\arraybackslash}X>{\raggedright\arraybackslash}X}
\toprule
\textbf{Property} & \textbf{FLUX / Caption}
& \textbf{SEED / A-OKVQA} \\
\midrule
Task family & Open-ended image description & Question-conditioned VQA \\
Factual authority & Provided image
& Provided image plus authoritative reference answer \\
Text supplied & Exact model response
& Question, all options, reference answer, and exact model response \\
Accuracy construct & Correspondence with visible image content
& Semantic correctness of the answer to the question \\
Hallucination construct & Unsupported or contradictory image claims
& False, contradictory, or unsupported answer content \\
Official dataset or exact-match metric & No & No \\
Requests per prediction & Two independent requests & Two independent requests \\
\bottomrule
\end{tabularx}
\end{table}

\paragraph{Image descriptions.}
Each evaluation request contains the image and the unedited model response. Accuracy assesses correspondence to the visible scene, whereas Hallucination assesses unsupported or contradictory visual claims. The original generation instruction, reference caption, dataset and model identities, competing responses, and earlier scores are withheld. The archived FLUX and Caption templates supply the response under \texttt{MODEL OUTPUT} (Appendix~\ref{app:description_prompts}).

\paragraph{Question-conditioned VQA.}
Each request provides the image, question, all answer options, authoritative reference answer, and unedited response, with method and model identities withheld. The options are used to interpret answer labels and resolve label--text conflicts; unselected distractors are excluded from the evidence and the candidate's claims. Correct labels, answer text, and semantic equivalents are accepted, including reasonable background knowledge connecting the image to the reference. Accuracy assesses question-answer correctness; unsupported additions affect Accuracy only if they contradict or undermine the selected answer and are assessed separately for Hallucination. Table~\ref{tab:vqa_boundaries} summarizes the scoring rules, and Appendix~\ref{app:vqa_prompts} reproduces the executed templates.

\begin{table}[H]
\caption{Scoring rules in the executed VQA prompts.}
\label{tab:vqa_boundaries}
\centering
\normalsize
\setlength{\tabcolsep}{3.5pt}
\renewcommand{\arraystretch}{1.18}
\begin{tabularx}{\linewidth}{>{\raggedright\arraybackslash}p{0.25\textwidth}>{\raggedright\arraybackslash}X>{\raggedright\arraybackslash}X}
\toprule
\textbf{Candidate response} & \textbf{Accuracy}
& \textbf{Hallucination} \\
\midrule
Correct label, text, or semantic equivalent & Exactly 100 when unambiguous & Exactly 0 \\
Correct answer plus grounded explanation & Exactly 100
& Exactly 0 \\
Correct answer plus fabricated explanation & Preserve correctness unless contradicted
& Greater than 0 according to unsupported explanation \\
Wrong answer only & Low or zero according to residual relevance
& Potentially near 100 because the only substantive claim is false \\
General image description without an answer & Reduce according to question-answering
value & Evaluate only asserted unsupported content \\
Label--text conflict & Evaluate the dominant or final answer
and degree of resolution & Score the false or contradictory portion \\
Unselected distractor & No effect & No effect \\
Empty or whitespace-only & Exactly 0
& Exactly 0; report empty coverage separately \\
\bottomrule
\end{tabularx}
\end{table}

\subsection{Closed-ended MCQ evaluation}
\label{app:mcq_evaluation}

To assess whether the VQA gains also appear independently of automated judging, we evaluate SEED-Bench and A-OKVQA under benchmark-aligned closed-ended answer-correctness protocols.

\paragraph{SEED-Bench.}
Following the official answer-ranking protocol~\citep{li2023seedbench}, we select the candidate option with the highest teacher-forced log-likelihood conditioned on the image and question. Let $I$ denote the image, $q$ the question, and $o_1,\ldots,o_4$ the four candidate options. After tokenization, option $o_k$ is represented by $(t_{k,1},\ldots,t_{k,L_k})$. Its score is
\begin{equation}
s_k=\sum_{j=1}^{L_k}\log p_\theta(t_{k,j}\mid I,q,t_{k,<j}).
\label{eq:seed_mcq_score}
\end{equation}
The predicted option is
\begin{equation}
\hat{k}=\arg\max_{k\in\{1,\ldots,4\}}s_k.
\label{eq:seed_mcq_prediction}
\end{equation}
Accuracy is computed as
\begin{equation}
\mathrm{Acc}_{\mathrm{SEED}}
=100\times\frac{N_{\mathrm{correct}}}{N},
\qquad N=1000,
\label{eq:seed_mcq_accuracy}
\end{equation}
where $N_{\mathrm{correct}}$ is the number of questions for which the predicted option matches the ground-truth answer.

\paragraph{A-OKVQA.}
We evaluate the selected option under the official A-OKVQA multiple-choice protocol~\citep{schwenk2022aokvqa}:
\begin{equation}
\mathrm{Acc}_{\mathrm{A\text{-}OKVQA}}
=100\times\frac{N_{\mathrm{correct}}}{N},
\qquad N=1000.
\label{eq:aokvqa_mcq_accuracy}
\end{equation}
Here, $N_{\mathrm{correct}}$ counts predictions matching the official \texttt{correct\_choice\_idx}. The random-choice accuracy is $25\%$.

\paragraph{Evaluation settings.}
We evaluate each Qwen3.5-2B interface using its own trained checkpoint. For each dataset, we use the same 1,000 examples as in Table~\ref{tab:main}; each prediction is produced by the corresponding model, without external candidate-answer replacement. On A-OKVQA, the model selects from the four supplied choices, and the selected option text is submitted to the official evaluation script for comparison with the ground truth.

\paragraph{Results.}
Table~\ref{tab:closed_ended_mcq} reports accuracy under these protocols. \modelname{} achieves 42.1\% on SEED-Bench and 43.6\% on A-OKVQA, improving on the strongest included comparators (DINOv3 and CoME-VL, respectively) by 1.5 and 1.3 percentage points. This comparison complements the response-quality evaluation with direct answer-correctness measurements on the same slices; its scope is the Qwen3.5-2B configurations listed in the table.
\begin{table}[!htbp]
\centering
\caption{Closed-ended multiple-choice accuracy (\%) on SEED-Bench and A-OKVQA using the Qwen3.5-2B backbone. No LLM judge is used.}
\label{tab:closed_ended_mcq}
\small
\setlength{\tabcolsep}{7pt}
\renewcommand{\arraystretch}{1.14}
\begin{tabular}{lcc}
\toprule
\textbf{Method} & \textbf{SEED-Bench (Acc.)} & \textbf{A-OKVQA (Acc.)} \\
\midrule
SigLIP & 34.4 & 37.1 \\
DINOv3 & 40.6 & 38.9 \\
Granulon & 39.3 & 37.7 \\
CoME-VL & 38.9 & 42.3 \\
LLaVA-Mini & 35.8 & 29.0 \\
\rowcolor{PAIQYellow}
\textbf{\modelname{}} & \textbf{42.1} & \textbf{43.6} \\
\bottomrule
\end{tabular}
\end{table}

\subsection{Evaluator routes and request validation}
\label{app:judge_routes}

\paragraph{Primary evaluation.}
We use \texttt{openai/gpt-4o} through OpenRouter for the primary VQA evaluation (Table~\ref{tab:primary_judge_settings}). Each request contains a single user message with the rendered prompt and one \texttt{image\_url} element at \texttt{high} detail, without a system message. Image-description requests primarily use \texttt{gpt-4o} through OpenLux. For each request, we retain the returned-model identifier and provider response ID.
\begin{table}[H]
\caption{Primary VQA evaluator settings. FLUX and Caption use a separate image-description route.}
\label{tab:primary_judge_settings}
\centering
\normalsize
\setlength{\tabcolsep}{4pt}
\renewcommand{\arraystretch}{1.18}
\begin{tabularx}{\linewidth}{>{\raggedright\arraybackslash}p{0.31\textwidth}>{\raggedright\arraybackslash}X}
\toprule
\textbf{Field} & \textbf{Frozen value} \\
\midrule
Provider / endpoint & OpenRouter; \url{https://openrouter.ai/api/v1/chat/completions} \\
Requested model & \texttt{openai/gpt-4o} \\
Accepted receipt & GPT-4o family identifier \\
Image detail / temperature & \texttt{high} / 0.1 \\
Maximum output tokens & 512 \\
Response format & \texttt{\{"type":"json\_object"\}} \\
System / user messages & No system message; one user message \\
Request contents & Image, question, options, reference answer, and candidate response \\
Connect / read timeout & 20 seconds / 120 seconds \\
Maximum cumulative attempts & Three per request key \\
Retry payload / in-provider fallback & Semantically identical payload / disabled \\
\bottomrule
\end{tabularx}
\end{table}

\paragraph{Recovery of missing evaluator responses.}
For SEED and A-OKVQA, we recover only audited request keys whose primary evaluation terminated with \texttt{empty\_response}, using the Qwen route in Table~\ref{tab:fallback_judge_settings}. Recovery is separate from provider-side automatic fallback, and valid GPT-4o scores are retained. Each recovery request matches the original rendered prompt, image, question, options, reference, and candidate response, and the recovered score retains its evaluator identity.
\begin{table}[!htbp]
\caption{Recovery route for the audited empty-response terminals of the VQA primary evaluator.}
\label{tab:fallback_judge_settings}
\centering
\normalsize
\setlength{\tabcolsep}{4pt}
\renewcommand{\arraystretch}{1.18}
\begin{tabularx}{\linewidth}{>{\raggedright\arraybackslash}p{0.31\textwidth}>{\raggedright\arraybackslash}X}
\toprule
\textbf{Field} & \textbf{Frozen value} \\
\midrule
Provider / endpoint & Paratera; \url{https://llmapi.paratera.com/v1/chat/completions} \\
Requested and accepted model
& \texttt{Qwen3-VL-235B-A22B-Instruct} only \\
Image detail / temperature & \texttt{high} / 0.1 \\
Maximum tokens / response format
& 512 / \texttt{\{"type":"json\_object"\}} \\
System message & None \\
Connect / read timeout & 20 seconds / 180 seconds \\
Maximum cumulative attempts & Three \\
Eligible scope & Exact audited GPT-4o
\texttt{empty\_response} terminal keys only \\
\bottomrule
\end{tabularx}
\end{table}

\paragraph{Validation and retries.}
An accepted result has HTTP status 200, an allowed returned-model identifier, one choice, \texttt{finish\_reason=stop}, nonempty content, and a nonempty provider response ID. The content must parse as one JSON object containing exactly the requested field, with an integer in $[0,100]$; Booleans, floats, and extra fields are rejected. The schemas are \texttt{\{"accuracy\_score": 0\}} and \texttt{\{"hallucination\_score": 0\}}, where zero illustrates the type rather than a default score.

Retryable events include HTTP 408, 409, 425, 429 and 5xx responses; timeouts and connection failures; empty content; invalid JSON or schema; and \texttt{finish\_reason=length}. Each retry preserves the semantic payload. Wrong-model, authentication, and quota errors fail closed. Attempts and terminal failures remain in append-only records. A failed evaluator request is not converted to zero, dropped from coverage, or imputed from another candidate output. Record validation establishes coverage and structural consistency, not semantic correctness of the evaluator's decisions.

\subsection{Evaluator-source coverage}
\label{app:judge_coverage}

Table~\ref{tab:judge_provenance} reports evaluator provenance for the finalized VQA archives. Each prediction contributes two scoring requests, one per metric. Of the 68,000 finalized VQA scores in this archive, 207 (0.3044\%) use the Qwen recovery route; the description archives contain five additional Qwen-recovered scores outside this denominator. The resulting evaluation therefore combines primary and recovery evaluator outputs.
\begin{table}[!htbp]
\caption{Evaluator-source counts in the finalized SEED/A-OKVQA archives. Counts refer to scoring axes, with two axes per prediction.}
\label{tab:judge_provenance}
\centering
\normalsize
\setlength{\tabcolsep}{3.8pt}
\renewcommand{\arraystretch}{1.18}
\begin{tabularx}{\linewidth}{>{\raggedright\arraybackslash}Xrrrr}
\toprule
\textbf{Final package} & \textbf{Total}
& \textbf{GPT-4o}
& \textbf{Qwen recovery}
& \textbf{Terminal} \\
\midrule
Qwen3.5-9B, InternVL3.5-8B, Llama-3.1-8B, Qwen3.5-27B
& 48,000 & 47,832 & 168 & 0 \\
Qwen3.5-2B & 20,000 & 19,961 & 39 & 0 \\
Combined finalized VQA scope & 68,000 & 67,793 & 207 & 0 \\
\bottomrule
\end{tabularx}
\end{table}
Matching prompts and output schemas preserves the request definition. Request-level records retain the provider, requested and returned model identifiers, response ID, source run, and prompt/context provenance. These coverage checks establish completeness and structural validity; inter-evaluator calibration and agreement with human judgments remain unverified.

\section{Additional Results and Statistical Analysis}
\label{app:additional_results}

\subsection{Qwen3.5-2B VQA results and recovery}
\label{app:qwen2b_final}

We examine the dedicated Qwen3.5-2B VQA archive, comprising five methods, two datasets, and 10,000 predictions (20,000 scoring requests). GPT-4o supplies 19,961 valid scores; the 39 terminal empty-response keys (34 Accuracy and five Hallucination) are recovered on the first attempt through the Qwen route. The resulting 20 method--dataset--metric groups each contain 1,000 records, with no missing or duplicate keys or terminal failures.

Table~\ref{tab:qwen2b_vqa_final} reports the mean scores and marginal 95\% bootstrap intervals for this five-method subset, including the fixed-rotation comparator and fixed-$10$ Granulon-style adaptation (Appendix~\ref{app:experiment_settings}).
\begin{table}[H]
\caption{Qwen3.5-2B VQA judge scores, reported as mean [95\% bootstrap CI]. Each cell contains 1,000 aligned records; the clustering comparator is the fixed-$10$ Granulon-style adaptation.}
\label{tab:qwen2b_vqa_final}
\centering
\normalsize
\setlength{\tabcolsep}{5pt}
\renewcommand{\arraystretch}{1.18}
\begin{tabularx}{\linewidth}{>{\raggedright\arraybackslash}Xcc}
\toprule
\textbf{Method} & \textbf{Accuracy $\uparrow$} & \textbf{Hallucination $\downarrow$} \\
\midrule
\multicolumn{3}{l}{\textbf{SEED-Bench}} \\
\DINO{} & 32.70 [29.89, 35.52] & 72.80 [70.85, 74.74] \\
\SIGLIP{} & 29.77 [27.00, 32.63] & 74.71 [72.55, 76.80] \\
Granulon-style ($K=10$) & 37.66 [34.74, 40.66] & 69.85 [67.85, 71.87] \\
Fixed $\mQ=\mI$ & 32.78 [29.92, 35.70] & 72.47 [70.38, 74.49] \\
\rowcolor{APXBlueTint} \textbf{\modelname{}} & \textbf{54.02 [50.94, 57.04]} & \textbf{52.63 [50.34, 54.86]} \\
\addlinespace[5pt]
\multicolumn{3}{l}{\textbf{A-OKVQA}} \\
\DINO{} & 28.29 [25.55, 30.98] & 77.94 [76.12, 79.80] \\
\SIGLIP{} & 27.68 [25.00, 30.49] & 78.86 [76.91, 80.76] \\
Granulon-style ($K=10$) & 34.26 [31.38, 37.22] & 73.85 [71.86, 75.83] \\
Fixed $\mQ=\mI$ & 32.27 [29.37, 35.18] & 75.86 [73.72, 77.95] \\
\rowcolor{APXBlueTint} \textbf{\modelname{}} & \textbf{48.85 [45.94, 51.92]} & \textbf{55.48 [53.05, 57.93]} \\
\bottomrule
\end{tabularx}
\end{table}
Across both datasets and both metrics, all 16 paired-difference intervals favor \modelname{} against the four comparators. In particular, relative to fixed interpolation, Accuracy increases by 21.24 points on SEED and 16.58 on A-OKVQA. The paired comparisons use aligned score differences; the marginal intervals in Table~\ref{tab:qwen2b_vqa_final} summarize each method separately. Both quantify sampling uncertainty conditional on the recorded outputs and evaluator scores.

\subsection{Aggregation and paired uncertainty}
\label{app:bootstrap}

For each backbone--method--dataset--metric setting, we estimate the mean score across the aligned evaluation records:
\begin{equation}
\bar s=\frac1N\sum_{i=1}^{N}s_i,\qquad N=1{,}000
\label{eq:score_mean}
\end{equation}
The estimate uses all records in a complete evaluation cell, including empty candidate responses. Accuracy and Hallucination are aggregated separately; any incomplete scoring set requires its actual denominator to be reported.

We use nonparametric percentile bootstrap intervals, drawing $N$ records with replacement per replicate and taking the 2.5th and 97.5th percentiles of the resulting means. The FLUX/Caption archive uses 10,000 replicates and seed base 20260810. The four-backbone VQA archive uses 2,000 replicates and the dedicated 2B VQA archive uses 10,000, with seeds derived deterministically from stable group labels in both VQA packages. The replicate counts differ in Monte Carlo precision at the same nominal confidence level.

For Table~\ref{tab:bootstrap}, the comparator in each backbone--dataset cell is the single encoder with higher mean Accuracy, fixed before paired resampling. For the same aligned sample $i$, the displayed differences are
\begin{equation}
\Delta_i^{\mathrm{acc}}=s_i^{\mathrm{acc}}(\text{\modelname{}})-s_i^{\mathrm{acc}}(b),\qquad
\Delta_i^{\mathrm{hall}}=s_i^{\mathrm{hall}}(b)-s_i^{\mathrm{hall}}(\text{\modelname{}}).
\label{eq:paired_differences}
\end{equation}
Positive differences favor \modelname{} on either metric. We resample the paired-difference vectors directly, preserving within-record correspondence. The raw archive uses the opposite sign for Hallucination differences; we reverse its interval endpoints and signs for display.
\begin{table}[H]
\caption{Paired-bootstrap 95\% confidence intervals over aligned sample-level Judge scores. FLUX/Caption use 10,000 resamples. SEED/A-OKVQA use 2,000 resamples for the four-backbone hybrid archive and 10,000 for the dedicated Qwen3.5-2B archive.}
\label{tab:bootstrap}
\centering
\normalsize
\setlength{\tabcolsep}{4pt}
\renewcommand{\arraystretch}{1.18}
\begin{tabularx}{\linewidth}{ll>{\centering\arraybackslash}X>{\centering\arraybackslash}X}
\toprule
\textbf{Backbone} & \textbf{Dataset}
& \textbf{$\Delta$Accuracy}
& \textbf{$\Delta$Hallucination} \\
\midrule
Qwen3.5-2B & FLUX & $[47.475,51.289]$ & $[46.807,50.219]$ \\
Qwen3.5-2B & Caption & $[52.523,56.385]$ & $[46.597,50.109]$ \\
Qwen3.5-2B & SEED & $[18.026,24.692]$ & $[17.823,22.596]$ \\
Qwen3.5-2B & A-OKVQA & $[17.109,24.063]$ & $[19.951,24.944]$ \\
\midrule
Qwen3.5-9B & FLUX & $[11.309,14.086]$ & $[14.014,17.097]$ \\
Qwen3.5-9B & Caption & $[23.801,28.979]$ & $[23.108,28.111]$ \\
Qwen3.5-9B & SEED & $[4.511,10.166]$ & $[5.798,10.240]$ \\
Qwen3.5-9B & A-OKVQA & $[7.904,14.036]$ & $[9.279,14.888]$ \\
\midrule
Qwen3.5-27B & FLUX & $[2.636,4.331]$ & $[3.596,5.853]$ \\
Qwen3.5-27B & Caption & $[1.204,4.992]$ & $[4.816,8.687]$ \\
Qwen3.5-27B & SEED & $[3.226,8.959]$ & $[8.922,14.083]$ \\
Qwen3.5-27B & A-OKVQA & $[3.101,8.049]$ & $[7.636,12.357]$ \\
\midrule
InternVL3.5-8B & FLUX & $[2.149,4.382]$ & $[2.816,5.470]$ \\
InternVL3.5-8B & Caption & $[6.808,10.044]$ & $[8.843,12.346]$ \\
InternVL3.5-8B & SEED & $[-0.778,3.721]$ & $[0.454,4.727]$ \\
InternVL3.5-8B & A-OKVQA & $[-2.677,2.164]$ & $[-0.888,3.697]$ \\
\midrule
Llama-3.1-8B & FLUX & $[44.950,48.596]$ & $[43.139,46.428]$ \\
Llama-3.1-8B & Caption & $[54.981,58.869]$ & $[44.433,48.234]$ \\
Llama-3.1-8B & SEED & $[6.279,13.637]$ & $[5.379,10.638]$ \\
Llama-3.1-8B & A-OKVQA & $[3.831,11.014]$ & $[4.882,10.176]$ \\
\bottomrule
\end{tabularx}
\end{table}

\paragraph{Interpretation.}
In 18 of the 20 backbone--dataset settings, both paired intervals in Table~\ref{tab:bootstrap} lie above zero. For InternVL/SEED, the Accuracy interval includes zero, and for InternVL/A-OKVQA, both intervals include zero. The intervals quantify record-level sampling uncertainty conditional on the selected comparator, fixed evaluator records, and a single training seed; they do not account for variation across training seeds, repeated evaluations, prompt changes, or multiple-comparison adjustment. The four 1,000-record slices contain 999, 1,000, 908, and 985 distinct images, respectively, so the record-level resampling also leaves within-image clustering unmodeled. Baseline selection is held fixed across bootstrap replicates.

\subsection{Post-hoc best-of-two reference}
\label{app:oracle}

To compare feature-level fusion with selection at the response level, we construct a post-hoc reference from the two independently generated single-encoder outputs. Let $J_d^{\mathrm{acc}}(x)$ and $J_s^{\mathrm{acc}}(x)$ denote their judged Accuracy for example $x$. We select
\begin{equation}
m^*(x)=\operatorname*{arg\,max}_{m\in\{d,s\}}J_m^{\mathrm{acc}}(x),
\label{eq:oracle_choice}
\end{equation}
assigning exact ties to \DINO{}. We then retain both scores of that selected output, $J_{m^*(x)}^{\mathrm{acc}}(x)$ and $J_{m^*(x)}^{\mathrm{hall}}(x)$. The archived comparison covers the five FLUX backbone settings. The reference upper-bounds Accuracy attainable by choosing between the two observed responses for each example; it neither independently optimizes Hallucination nor bounds the performance of a representation that combines both feature streams. Selection uses the completed outputs and their judged scores, so it is a post-hoc reference rather than a deployable routing policy.

\section{Analytical Compute Accounting}
\label{app:compute}

The analytical compute proxy in Figure~\ref{fig:oracle_efficiency} and Table~\ref{tab:efficiency} characterizes the cost of the selected visual-encoder and language-model terms using an approximate $2PT$ accounting, where $P$ is the parameter count and $T$ is the token budget. We use 201 \DINO{} tower tokens and 577 \SIGLIP{} tower tokens, including special tokens, with parameter counts of 303M and 316M, respectively. The language model receives 196 or 576 visual tokens depending on the interface, with a fixed text budget of 64 tokens. \modelname{} retains a 196-token visual prefix while evaluating both encoders.

The average-efficiency coordinate includes both visual towers and the mean cost of the two language branches (196 and 576 visual tokens). It serves as an analytical reference for plotting the post-hoc best-of-two Accuracy in Appendix~\ref{app:oracle}, rather than the total cost of generating and selecting between both responses. The accounting excludes the fusion operations, including transport normalization and the dense Cayley solve, whose asymptotic cost is given in~\PAIQeqref{eq:complexity}. Thus the proxy characterizes the selected computation under specified token budgets, rather than complete end-to-end FLOPs, measured latency, or memory.

For the archived Qwen3.5-27B language-side parameter count $P=26{,}895{,}998{,}464$, the corresponding estimates are 14,107.7, 34,791.5, 24,692.9, and 14,472.4 GFLOPs for \DINO{}-only, \SIGLIP{}-only, average efficiency, and \modelname{}, respectively. The Llama ledger uses the untied configuration
$(d,d_{\mathrm{ff}},L,h,h_{\mathrm{KV}},V)=(4096,14336,32,32,8,128256)$,
where $L$ is decoder depth. Device-matched latency and memory measurements, as well as the potential effects of caching $\mQ$, are outside the scope of this accounting.

\section{Qualitative Diagnostics and Case Interpretation}
\label{app:case_record}

\subsection{Local Residual Updates and Fixed-Answer Sensitivity}
\label{app:reliance_trace}

We illustrate the distinction between relative update magnitude and fixed-answer likelihood sensitivity in Figure~\ref{fig:ablation_reliance}\subref{fig:reliance_trace} using FLUX sample 501 from the Qwen3.5-2B \modelname{} archive, stored as a $1024\times1024$ JPEG with image and array provenance. Its generated response begins: ``Two individuals, a woman in a flowing floral dress and a man in a sharp navy suit, walk hand-in-hand ...''.

\paragraph{Blockwise residual intervention.}
We divide the $14\times14$ fused grid into 49 disjoint $2\times2$ blocks. For a block $B$, let $m_j(B)=0$ for $j\in B$ and $m_j(B)=1$ otherwise. The counterfactual token is
\begin{equation}
\vz_j^{(-B)}=\vd_j+m_j(B)\lambda\mQ(\bar{\vs}_j-\vd_j),
\label{eq:block_ablation}
\end{equation}
which restores the base feature only inside $B$ and is equivalent to~\PAIQeqref{eq:block_replacement}. We teacher-force the original answer $y$ under both $\mZ$ and $\mZ^{(-B)}$, holding the text, answer prefixes, and all other features fixed. The difference in summed token log-likelihoods, $R(B)$ in~\PAIQeqref{eq:reliance}, is positive when removing the residual lowers the answer's likelihood. No answer is regenerated. Relative update magnitude $\rho_j$ is defined in~\PAIQeqref{eq:rho}.

\paragraph{Construction and reliance maps.}
The update-magnitude map in panel (b2) shows relatively large updates over a broad part of the output grid. In panel (b3), the largest positive likelihood changes are more localized, near positions corresponding to the woman's hair and floral dress, both of which appear in the generated description. We evaluate all 49 blocks against the same answer and select the highlighted blocks by $R(B)$. This contrast illustrates the distinction between the magnitude of an update and the sensitivity of the fixed answer to its removal.

\paragraph{Interpretation.}
The visualization clips negative $R(B)$ values to zero, omitting increases in the fixed answer's likelihood under residual removal. The spatial labels ``hair'' and ``floral dress'' link the highlighted boxes to phrases in the description for inspection. Since $R(B)$ sums log-likelihood changes over the entire answer, the displayed correspondences do not establish phrase-level causality. As discussed in Appendix~\ref{app:diagnostics}, the measurements depend on the chosen intermediate feature parameterization. This selected example illustrates the diagnostic; it does not estimate its prevalence or reliability across the dataset.

\subsection{Scope of the qualitative comparison}
Appendix~\ref{app:extended-cases} presents six selected comparisons with decoder-attention overlays. Those overlays summarize attention to visual-token positions, whereas $\boldsymbol\Pi$ describes cross-encoder aggregation and $R(B)$ measures a specific residual-removal effect. The case captions distinguish recognition successes from unsupported details in the same response. Neither selected examples nor attention concentration estimate population-level accuracy or establish phrase-level causality.

\clearpage
\input{appendices/qualitative_cases.tex}

\clearpage
\section{Evaluator Prompts}
\label{app:prompts}

The VQA prompts below are the executed axis-specific templates. Each request contains one candidate response and its task evidence, with no competing output or score from the other axis. The two image-description tasks use a different archived template family, identified in Appendix~\ref{app:description_prompts}; a description rubric is not substituted for the VQA rubric or vice versa.

\subsection{Question-conditioned VQA prompts}
\label{app:vqa_prompts}

For both axes, replace the question, options, reference, and prediction placeholders with the exact record fields and attach the corresponding image. The requests are independent; no score or conversation state is carried from one to the other.

\paragraph{Accuracy.}
\begin{APXRawPrompt}[listing options={style=APXLiteral,breakautoindent=false,breakindent=0pt}]{VQA Accuracy evaluator}{prm:vqa-accuracy}
You are an objective evaluator of a question-conditioned visual question answering
response. Use the PROVIDED IMAGE together with the QUESTION, OPTIONS, and REFERENCE
ANSWER. Assign one ACCURACY score from 0 to 100.

ROLE OF THE OPTIONS:
The OPTIONS are supplied only to interpret option-letter references, understand
candidate wording, and identify contradictions between a label and written answer.
Their presence does not mean the MODEL RESPONSE selected them, and the distractors are
not factual evidence. The REFERENCE ANSWER is authoritative. This is a continuous
semantic VQA evaluation, not an exact-match multiple-choice accuracy calculation.

ACCURACY DEFINITION:
Accuracy measures how correctly the MODEL RESPONSE answers the specific question. A
response may give the answer using an option letter, option text, or a semantically
equivalent phrase. Do not require exact wording. If duplicate options carry the same
reference text, treat any matching label as correct.

Some questions may require commonsense or outside knowledge in addition to visual
evidence. Do not reject the authoritative reference merely because it cannot be derived
from pixels alone.

A response that clearly gives the correct option or answer receives exactly 100 even
when it is brief. Additional unsupported explanation is evaluated separately under
Hallucination and must not lower Accuracy unless it contradicts, changes, or undermines
the selected answer. Do not reward a general image description that fails to answer the
question. Accuracy and Hallucination are independent and must not sum to 100.

If an option letter conflicts with the written answer or final answer, interpret both
using the supplied option mapping. Evaluate the response as a whole, considering which
answer is dominant or final, how clearly the response resolves the inconsistency, and
how the inconsistency affects overall answer correctness.

Select the band that best represents question-answer correctness, then use the ones
digit to express the position within that band.

ACCURACY SCORE    LEVEL
100     Fully correct. The response clearly gives the authoritative reference answer,
        by correct option, answer text, or semantic equivalent, with no contradiction
        that changes or undermines the answer.
91-99   Excellent. The core answer is correct, but slight hedging, ambiguity, or a
        trivial internal inconsistency prevents an unqualified 100.
81-90   Very high. The intended answer is dominant and essentially correct, but the
        response is qualified, incomplete, or mildly inconsistent.
71-80   High. The response identifies the correct concept or option direction, but an
        important qualification or ambiguity remains.
61-70   Moderately high. The response shows the correct line of interpretation but does
        not clearly commit to or fully express the authoritative answer.
51-60   Moderate. Meaningful question-relevant understanding is present, but the core
        answer remains unresolved or is mixed with a competing answer.
41-50   Moderately low. Some relevant content is correct, but the response does not
        provide a reliable answer to the question.
31-40   Low. The response is largely incorrect but contains limited question-relevant
        evidence, interpretation, or partial understanding.
21-30   Very low. The response offers little question-answering value and may mainly
        describe the image or provide generic information without resolving the answer.
11-20   Minimal. Only isolated question-relevant content is present.
0-10    Negligible. The response is empty, unrelated, or provides no meaningful answer.
        An empty response receives exactly 0.

QUESTION:
<<QUESTION>>

OPTIONS:
<<OPTIONS>>

REFERENCE ANSWER:
<<REFERENCE>>

MODEL RESPONSE:
<<PREDICTION>>

Return ONLY a valid JSON object with exactly one integer field:
{"accuracy_score": 0}
\end{APXRawPrompt}

\paragraph{Hallucination.}
\begin{APXRawPrompt}[listing options={style=APXLiteral,breakautoindent=false,breakindent=0pt}]{VQA Hallucination evaluator}{prm:vqa-hallucination}
You are an objective evaluator of hallucination in a question-conditioned visual
question answering response. Use the PROVIDED IMAGE together with the QUESTION,
OPTIONS, and REFERENCE ANSWER. Assign one HALLUCINATION score from 0 to 100.

ROLE OF THE OPTIONS:
The OPTIONS are supplied only to interpret option-letter references, understand
candidate wording, and identify contradictions between a label and written answer.
Do not treat any option as a claim made by the MODEL RESPONSE merely because it appears
in the OPTIONS block. The distractors are not factual evidence. The REFERENCE ANSWER is
authoritative. This is not an exact-match multiple-choice accuracy calculation.

HALLUCINATION DEFINITION:
Hallucination measures false, contradictory, or unsupported content asserted by the
MODEL RESPONSE. Evaluate both its selected answer and any explanation. Treat the
authoritative REFERENCE ANSWER as supported for this benchmark.

Some questions may require commonsense or outside knowledge. Treat reasonable knowledge
needed to connect visible evidence to the accepted answer as supported. Do not label a
claim hallucinated merely because it is not directly visible in the pixels. However,
invented named entities, exact numbers, locations, events, causes, motives, or knowledge
claims that are not supported by the image, question, reference answer, or a reasonable
inference are hallucinations.

A correct answer alone, expressed by option letter, option text, or semantic equivalent,
receives exactly 0 Hallucination. A correct answer followed by fabricated reasoning may
receive nonzero Hallucination. Do not penalize omissions, brevity, or failure to describe
the entire image.

A wrong answer is a false answer claim and contributes Hallucination, but do not compute
Hallucination as 100 minus Accuracy. Weigh the severity, prominence, and fraction of
unsupported content. A short response whose only substantive claim is a wrong answer
may be almost entirely hallucinated, while a longer response may contain substantial
grounded analysis despite reaching a wrong final answer.

If an option letter conflicts with the written answer or final answer, interpret both
using the supplied option mapping and score the false or contradictory portion. Do not
attribute any unselected option text to the model.

Select the best severity band, then use the ones digit within that band.

HALLUCINATION SCORE    SEVERITY
0       No hallucination. The answer and all substantive explanation are supported by
        the task evidence, accepted reference, and reasonable inference.
1-10    Negligible hallucination. A trivial unsupported modifier or imprecision does
        not affect the answer.
11-20   Very low hallucination. A small unsupported secondary statement is present.
21-30   Low hallucination. Some noticeable unsupported content appears, while the core
        answer and most reasoning remain grounded.
31-40   Moderate hallucination. A clear unsupported or contradictory statement affects
        factual reliability, but substantial grounded content remains.
41-50   Moderately high hallucination. False or unsupported content is prominent, or
        the response materially conflicts between correct and incorrect answers.
51-60   High hallucination. A large portion of the answer or explanation is unsupported.
61-70   Very high hallucination. The response is predominantly unsupported,
        substantially contradicts the accepted answer, or relies on materially false
        outside knowledge.
71-80   Severe hallucination. Most substantive claims are unsupported or describe a
        materially wrong answer or scene.
81-90   Very severe hallucination. Nearly the entire substantive response is false,
        contradictory, or based on invented visual or knowledge evidence.
91-100  Extreme hallucination. The only substantive answer is wrong, or the response is
        wholly unrelated to the question and image. Use exactly 100 only when
        essentially no substantive claim is supported.

An empty response receives exactly 0 Hallucination because omission is not an asserted
false claim. Empty-output coverage must be recorded and reported separately.

QUESTION:
<<QUESTION>>

OPTIONS:
<<OPTIONS>>

REFERENCE ANSWER:
<<REFERENCE>>

MODEL RESPONSE:
<<PREDICTION>>

Return ONLY a valid JSON object with exactly one integer field:
{"hallucination_score": 0}
\end{APXRawPrompt}

\Needspace{24\baselineskip}
\subsection{Image-description protocol artifacts}
\label{app:description_prompts}

The FLUX and Caption runs use image-description templates, with a separate template for the fixed-$\mQ=\mI$ ablation. The description-score differences in that comparison should therefore be interpreted in light of this template distinction. The rendered requests contain the source image and the exact candidate text under \texttt{MODEL OUTPUT}, without the generation instruction, reference caption, question, or answer options. Accuracy and Hallucination are requested separately as integer scores in $[0,100]$. The historical templates do not prescribe an explicit score for an empty candidate.

Exact reproduction depends on the serialized templates and their rendered requests; their full text is not reproduced here. The descriptive protocol summary above does not introduce additional scoring rules. In particular, the VQA empty-output rule is not retrospectively applied to the historical description scores. 

\paragraph{Reproduction records.}
The evaluation archive associates each score with its sample ID, generated response, evaluator route, rendered prompt, provider response ID, and attempt history. Training manifests, checkpoint identifiers, decoding limits, and compute assumptions define the remaining execution conditions. Numerical summaries use the recorded scores and fixed sample sets; syntactic validation and complete request coverage are distinct from the semantic validity of an automated judgment.

%% file: appendices/related_work.tex
\section{Related Work}
\label{app:related_work}

\paragraph{Complementary visual representations.}
Visual pretraining objectives give rise to different representational strengths. CLIP~\citep{CLIP} and SigLIP~\citep{SigLIP} learn language-aligned features through image--text supervision, while DINOv2~\citep{DINOv2} and DINOv3~\citep{DINOv3} learn self-supervised representations with strong local structure for dense prediction. These strengths overlap: SigLIP~2~\citep{SigLIP2} improves localization and dense representations, and dino.txt~\citep{DINOText} aligns DINOv2 features with text. In the context of multimodal large language models (MLLMs), Eyes Wide Shut~\citep{EyesWideShut} identifies visual blind spots in CLIP that persist in downstream models and mitigates them with complementary self-supervised features. Granulon~\citep{Granulon} explores a complementary direction within a single DINOv3 stream, using a text-conditioned granularity controller to guide pooling and relation-aware clustering, yielding multi-granularity semantic tokens.

\paragraph{Compact multi-encoder fusion.}
Integrating heterogeneous visual features requires establishing cross-encoder correspondence while controlling the visual sequence length. Common-grid approaches fuse features through channel concatenation: COMM~\citep{COMM} aggregates multilayer CLIP and DINOv2 features on a shared patch grid, and Eagle~\citep{Eagle} resamples features from multiple vision experts to a common resolution. MERV~\citep{MERV} extends common-grid fusion to video through spatiotemporal pooling and input-dependent encoder-level mixture weights. Query-based interfaces offer another way to control the token budget. BRAVE's MEQ-Former~\citep{BRAVE} aggregates multi-encoder features into a fixed-length sequence, while Cambrian-1~\citep{Cambrian1} restricts spatially organized queries to corresponding regions across encoders. CoME-VL~\citep{CoMEVL} uses SigLIP~2 tokens to query DINOv3 features across heterogeneous grids and applies orthogonal projections during multilayer aggregation within each encoder. LLaVA-Mini~\citep{LLaVAMini} reduces the explicit visual sequence through query-based compression and image--text pre-fusion, also conveying visual information through conditioned text states. While CoME-VL uses orthogonal projections within each encoder, \modelname{} applies a shared orthogonal transform to the residual between each base feature and its cross-encoder aggregate, explicitly constraining the geometry of local updates while retaining the base-token budget.

\paragraph{Visual evidence use and intervention-based analysis.}
The contribution of visual features to answer prediction can be examined through interventions on the model's internal computation. Attention weights alone provide limited evidence of predictive dependence~\citep{AttentionNotExplanation}. \citet{CrossModalFlow} trace visual information flow by blocking image-to-text attention pathways at different depths and measuring the resulting changes in answer prediction. HalluTrace~\citep{HalluTrace} uses component-level interventions to distinguish visual grounding failure, language-prior dominance, and cross-modal conflict as sources of hallucination. Following this line of analysis, we examine the sensitivity of a fixed answer's likelihood to local residual removal, complementing the characterization of how fused visual representations are constructed.

%% file: appendices/qualitative_cases.tex
\section{Extended Qualitative Cases with Decoder Attention}
\label{app:extended-cases}

\begingroup
\colorlet{PAIQCaseBaselineBG}{black!3}
\ifPAIQcolor
  \colorlet{PAIQCaseMethodBG}{PAIQTheoryFill}
\else
  \colorlet{PAIQCaseMethodBG}{black!3}
\fi

We examine six selected Qwen3.5-2B examples: three FLUX descriptions and three Caption descriptions. They illustrate object and attribute errors rather than estimate their frequency. The displayed responses retain their original wording, including errors; an ellipsis marks a response that reached the 128-token generation limit. Selected \PAIQcasegood{image-supported phrases} and \PAIQcasebad{unsupported or contradicted phrases} are annotated in green and red, respectively. Uncolored text is unassessed, ambiguous, or stylistic; it is not an endorsement of the remaining response.

Each overlay averages decoder self-attention from the teacher-forced answer-token queries to visual-token keys over all heads at zero-based layers 11, 15, 19, and 23. The visual grids are $14\times14$ for \DINO{} and \modelname{}, and $24\times24$ for \SIGLIP{}. Each map is normalized independently: color intensity describes allocation within that map, not comparable total visual reliance across models. These maps are neither the cross-encoder weights $\boldsymbol\Pi$ nor the residual-removal score $R(B)$, and do not establish object localization or phrase-level causality.

\subsection{Case 1: FLUX 549---Watercolor grape cluster}
\label{app:case-flux-549}

\modelname{} identifies the purple grapes, whereas \DINO{} describes a strawberry and \SIGLIP{} describes a flower bouquet. The description is not uniformly correct: \modelname{} calls the elongated cluster \emph{spherical}.

  \noindent\begin{minipage}{\linewidth}
  \begin{minipage}[t]{0.238\linewidth}\vspace{0pt}\centering
    \includegraphics[width=\linewidth]{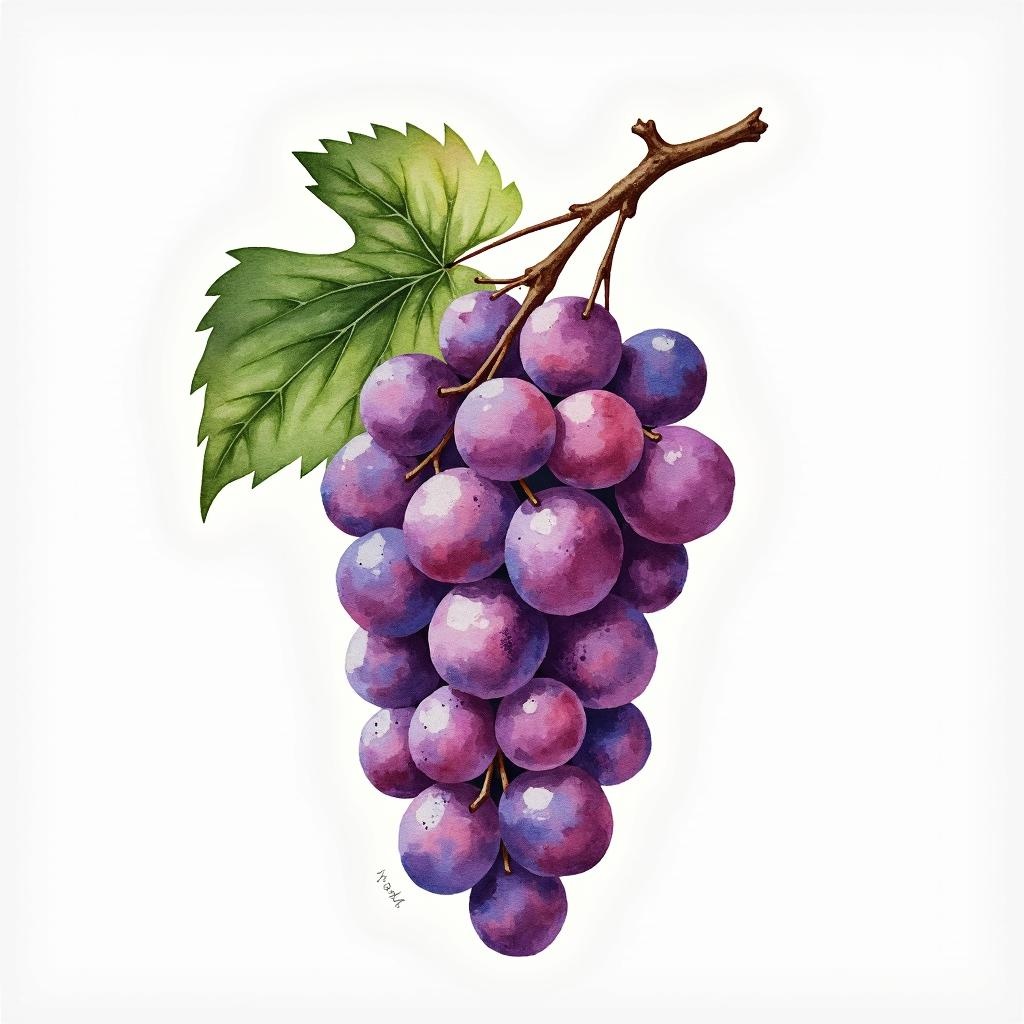}\par
    {\footnotesize\textbf{(a) Input}}
  \end{minipage}\hfill
  \begin{minipage}[t]{0.238\linewidth}\vspace{0pt}\centering
    \includegraphics[width=\linewidth]{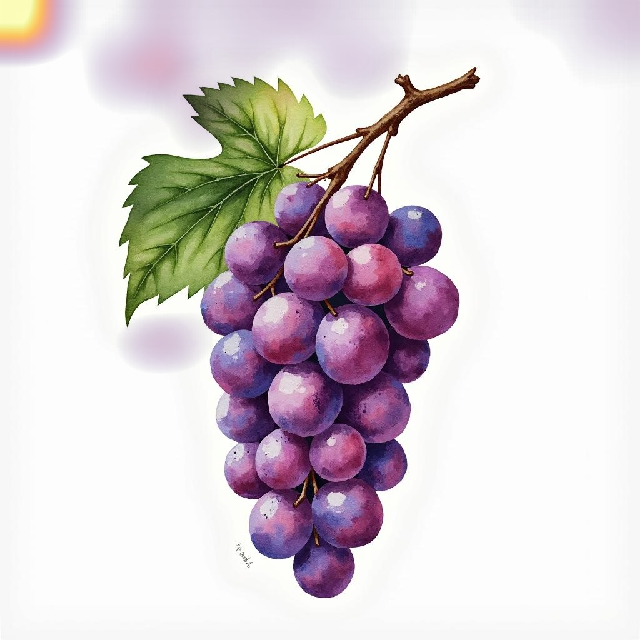}\par
    {\footnotesize\textbf{(b) \DINO{} attention}}
  \end{minipage}\hfill
  \begin{minipage}[t]{0.238\linewidth}\vspace{0pt}\centering
    \includegraphics[width=\linewidth]{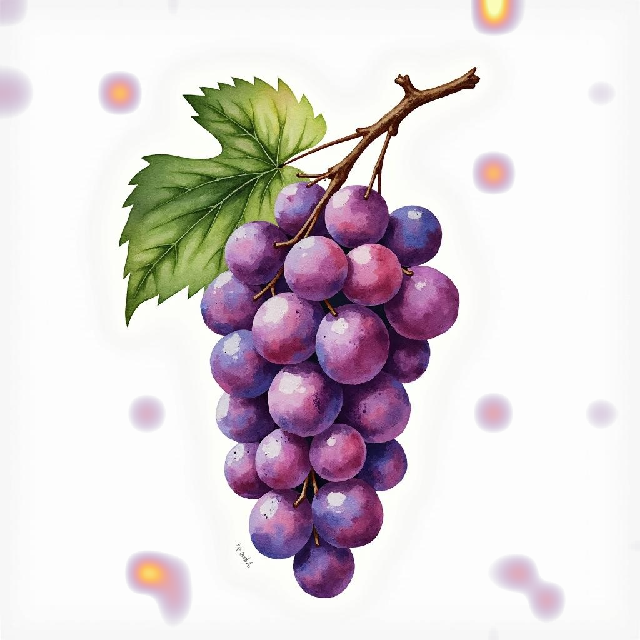}\par
    {\footnotesize\textbf{(c) \SIGLIP{} attention}}
  \end{minipage}\hfill
  \begin{minipage}[t]{0.238\linewidth}\vspace{0pt}\centering
    \includegraphics[width=\linewidth]{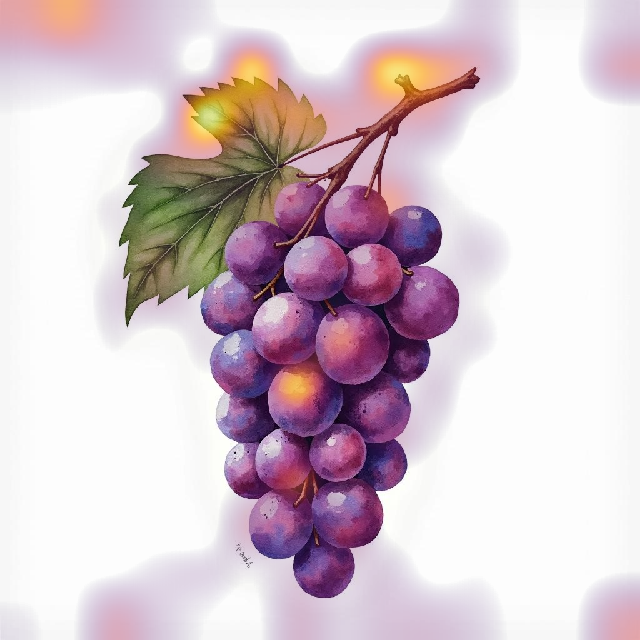}\par
    {\footnotesize\textbf{(d) \modelname{} attention}}
  \end{minipage}
  \end{minipage}\par\nopagebreak[4]
\begin{paiqcaseoutput}{PAIQCaseBaselineBG}
\textbf{\DINO{}-only output.}\enspace Vibrant and meticulously crafted, this \PAIQcasegood{digital illustration} of \PAIQcasebad{a vibrant red strawberry} stands out with its rich, glossy texture and symmetrical design. The strawberry is depicted in a side profile, showcasing its \PAIQcasebad{deep crimson hue} and \PAIQcasegood{a single, prominent green leaf} at the top, which adds a touch of natural elegance. The fruit's surface is adorned with \PAIQcasebad{numerous small, white seeds}, creating a textured appearance that enhances its realism. \PAIQcasegood{The background is a clean, off-white canvas}, allowing the strawberry to be the sole focal point, emphasizing its vivid colors and intricate details. The composition is simple yet striking, with the strawberry positioned slightly off-center to the left\ldots{}
\end{paiqcaseoutput}
\begin{paiqcaseoutput}{PAIQCaseBaselineBG}
\textbf{\SIGLIP{}-only output.}\enspace Vibrant and meticulously crafted, this image showcases \PAIQcasebad{a bouquet of fresh flowers}, including \PAIQcasebad{vibrant red roses, delicate white lilies}, and lush green leaves, arranged in a \PAIQcasebad{simple yet elegant white vase}. The composition is centered, with the flowers filling the frame, creating a balanced and harmonious visual effect. The roses, with their rich red petals and subtle yellow centers, stand out prominently against the \PAIQcasegood{clean white background}, while the lilies, with their pristine white petals and green stems, add a touch of purity and contrast. The green leaves, varying in size and shape, provide a natural frame around the flowers, enhancing the overall aesthetic appeal\ldots{}
\end{paiqcaseoutput}
\begin{paiqcaseoutput}{PAIQCaseMethodBG}
\textbf{\modelname{} output.}\enspace Vibrant and meticulously rendered, \PAIQcasegood{this watercolor illustration} captures the essence of \PAIQcasegood{a cluster of purple grapes}, each berry glistening with a rich, deep hue that suggests freshness and ripeness. \PAIQcasegood{The grapes are densely packed}, forming a lush, \PAIQcasebad{spherical cluster} that exudes a sense of abundance and natural beauty. Each grape is depicted with \PAIQcasegood{subtle shading and highlights}, enhancing their three-dimensional appearance and inviting the viewer to appreciate their intricate details. \PAIQcasegood{A slender brown vine extends from the top right}, adorned with \PAIQcasegood{a single, vibrant green leaf featuring serrated edges} and a glossy texture, adding a touch of organic elegance. The composition is centered around the grape cluster\ldots{}
\end{paiqcaseoutput}

\clearpage

\subsection{Case 2: FLUX 768---Skull forearm tattoo}
\label{app:case-flux-768}

\modelname{} identifies the skull tattoo and its placement on a forearm; the single-encoder outputs instead describe a three-dimensional head and a baseball cap. Its statement that the forearm is partly covered by clothing remains unsupported.

  \noindent\begin{minipage}{\linewidth}
  \begin{minipage}[t]{0.238\linewidth}\vspace{0pt}\centering
    \includegraphics[width=\linewidth]{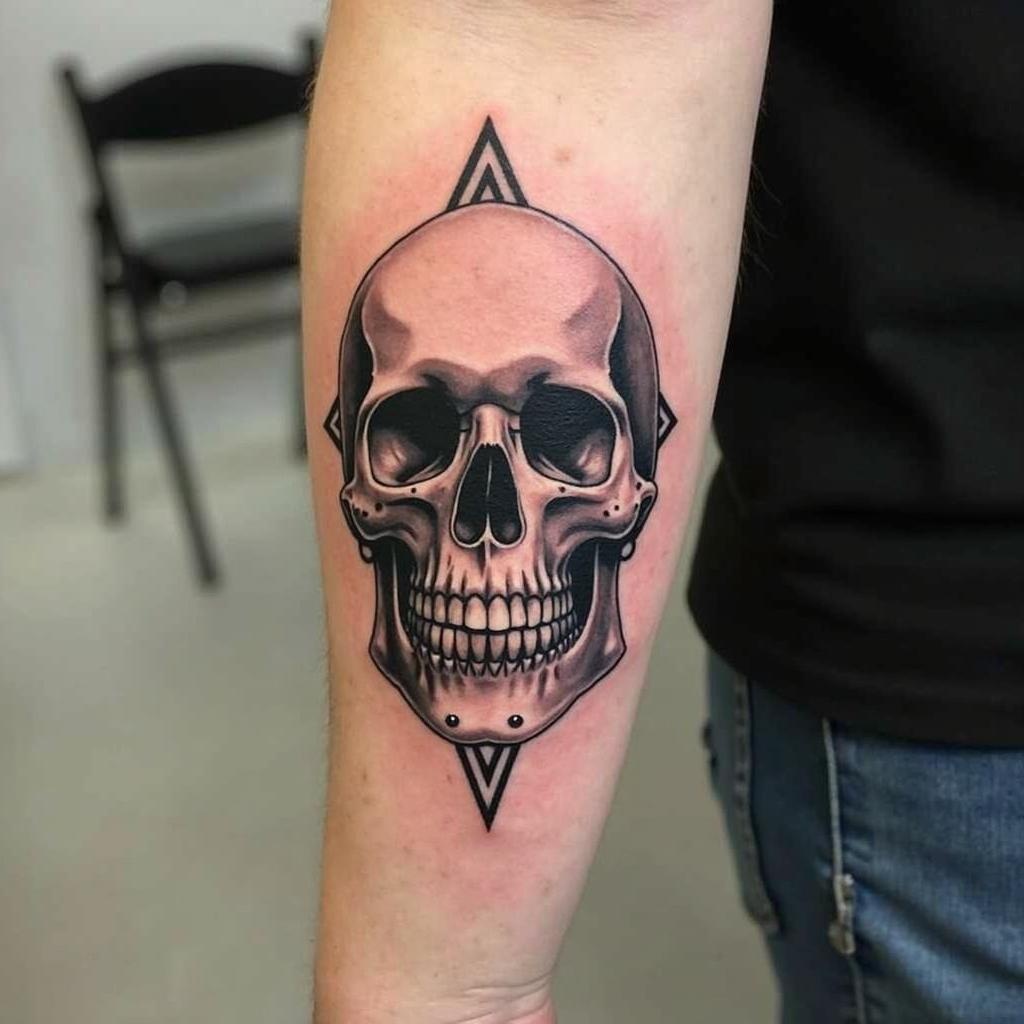}\par
    {\footnotesize\textbf{(a) Input}}
  \end{minipage}\hfill
  \begin{minipage}[t]{0.238\linewidth}\vspace{0pt}\centering
    \includegraphics[width=\linewidth]{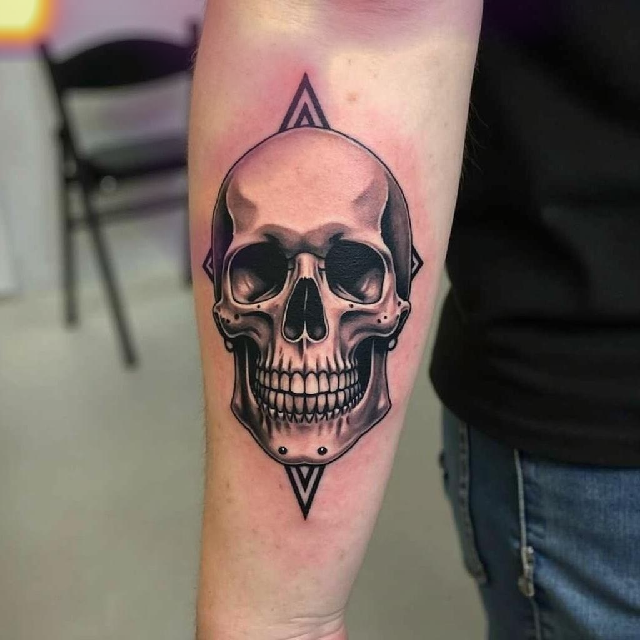}\par
    {\footnotesize\textbf{(b) \DINO{} attention}}
  \end{minipage}\hfill
  \begin{minipage}[t]{0.238\linewidth}\vspace{0pt}\centering
    \includegraphics[width=\linewidth]{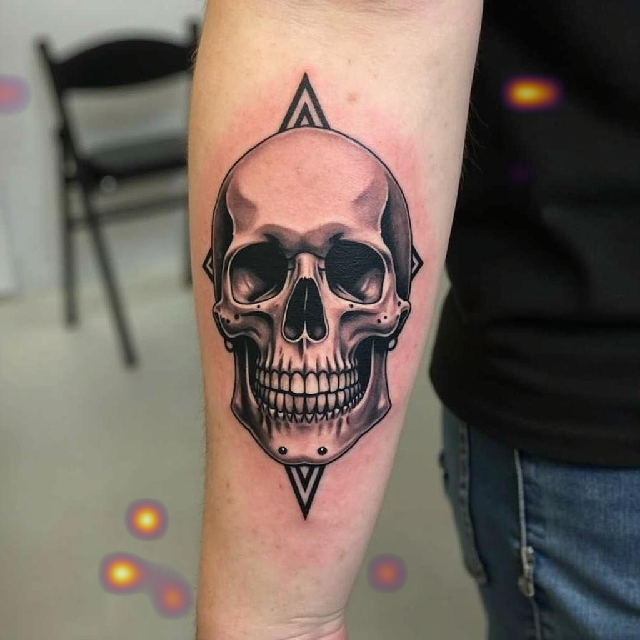}\par
    {\footnotesize\textbf{(c) \SIGLIP{} attention}}
  \end{minipage}\hfill
  \begin{minipage}[t]{0.238\linewidth}\vspace{0pt}\centering
    \includegraphics[width=\linewidth]{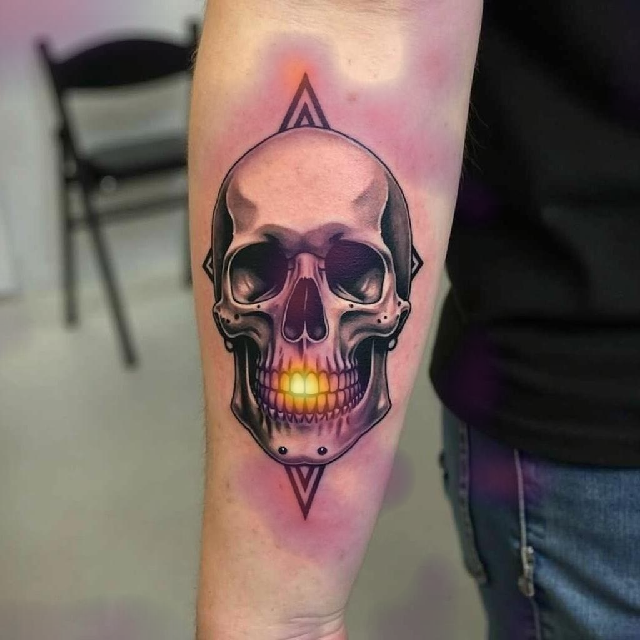}\par
    {\footnotesize\textbf{(d) \modelname{} attention}}
  \end{minipage}
  \end{minipage}\par\nopagebreak[4]
\begin{paiqcaseoutput}{PAIQCaseBaselineBG}
\textbf{\DINO{}-only output.}\enspace A meticulously crafted \PAIQcasebad{3D-printed model of a human head}, rendered in a photorealistic style, is displayed \PAIQcasebad{against a plain white background}. The head features \PAIQcasebad{a detailed, realistic face} with a prominent nose, expressive eyes, and a full mouth, all rendered with high precision. The skin texture is smooth yet detailed, capturing subtle variations in tone and shading that give the model a lifelike appearance. \PAIQcasebad{The hair is styled in a short, dark cut}, framing the face and adding to the overall realism. The model is positioned centrally, with its face turned slightly to the side, allowing a clear view of its features. The lighting\ldots{}
\end{paiqcaseoutput}
\begin{paiqcaseoutput}{PAIQCaseBaselineBG}
\textbf{\SIGLIP{}-only output.}\enspace Close-up portrait captures a meticulously crafted \PAIQcasebad{leather baseball cap}, showcasing its intricate details and craftsmanship. The cap features a \PAIQcasebad{prominent embroidered logo} on the front, depicting a stylized figure in a dynamic pose, rendered in bold black and white threads against a textured brown leather background. The cap's design includes a \PAIQcasebad{structured crown with a slightly curved brim}, secured by a visible button at the top, and \PAIQcasebad{a strap with a metal buckle} on the side, adding a touch of functionality and style. The leather material exhibits a rich, warm tone with subtle stitching and texture, emphasizing the artisanal quality of the piece. The composition focuses tightly on the cap,\ldots{}
\end{paiqcaseoutput}
\begin{paiqcaseoutput}{PAIQCaseMethodBG}
\textbf{\modelname{} output.}\enspace Dominating the frame is a meticulously crafted \PAIQcasegood{black-and-white tattoo of a skull}, \PAIQcasegood{positioned prominently on a person's forearm}. The skull, rendered with sharp, symmetrical precision, features \PAIQcasegood{hollow eye sockets}, a prominent nose, and a row of sharp teeth, all accentuated by bold, clean lines. The design is \PAIQcasegood{framed by two triangular shapes, one above and one below the skull}, adding a geometric contrast to the organic form. The tattoo is set against the smooth skin of the forearm, which is \PAIQcasebad{partially covered by the individual's dark sleeve and light blue jeans}, suggesting a casual yet deliberate artistic choice. The background is minimal, with a\ldots{}
\end{paiqcaseoutput}

\clearpage

\subsection{Case 3: FLUX 1453---Rendered virus particle}
\label{app:case-flux-1453}

\modelname{} describes the rendered virus particle and the smaller particles behind it. \DINO{} and \SIGLIP{} instead describe a snowman and a rose. The comparison concerns recognition of the depicted subject, not identification of a biological species.

  \noindent\begin{minipage}{\linewidth}
  \begin{minipage}[t]{0.238\linewidth}\vspace{0pt}\centering
    \includegraphics[width=\linewidth]{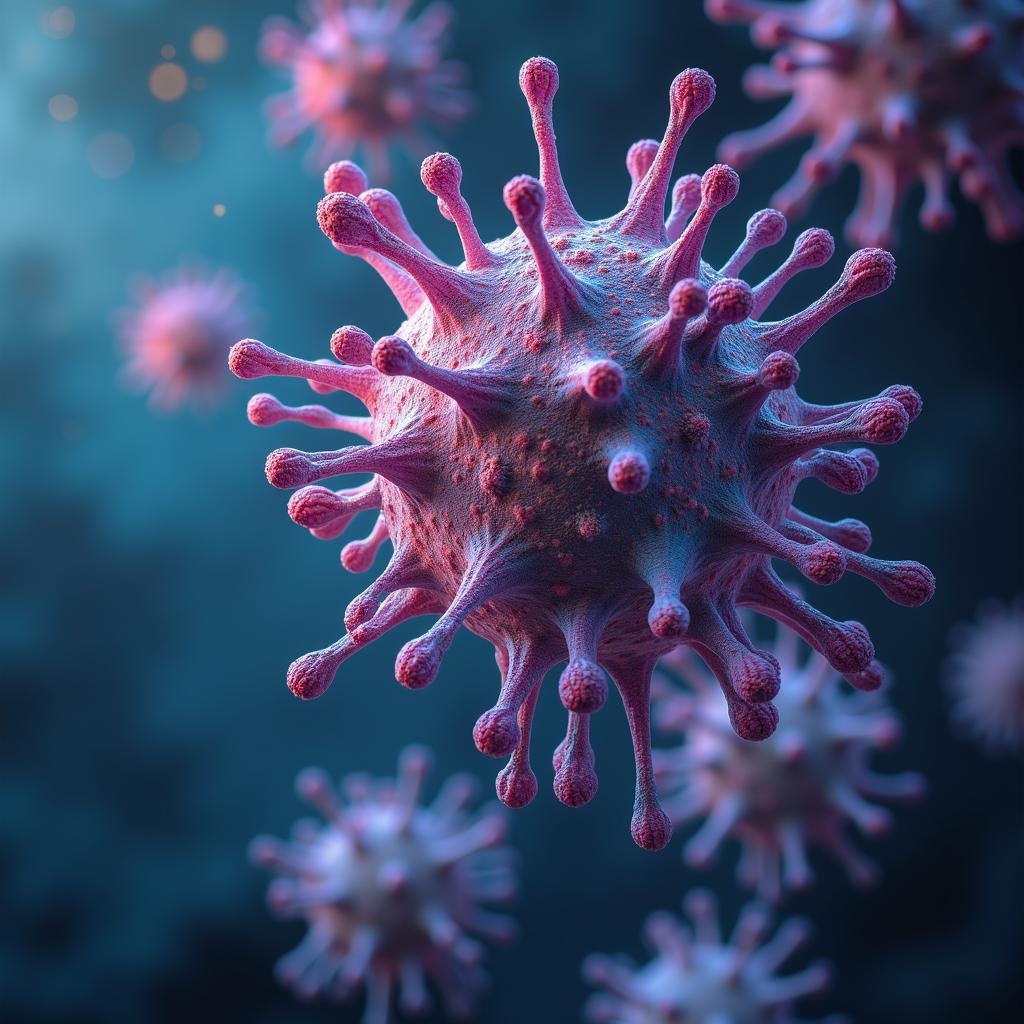}\par
    {\footnotesize\textbf{(a) Input}}
  \end{minipage}\hfill
  \begin{minipage}[t]{0.238\linewidth}\vspace{0pt}\centering
    \includegraphics[width=\linewidth]{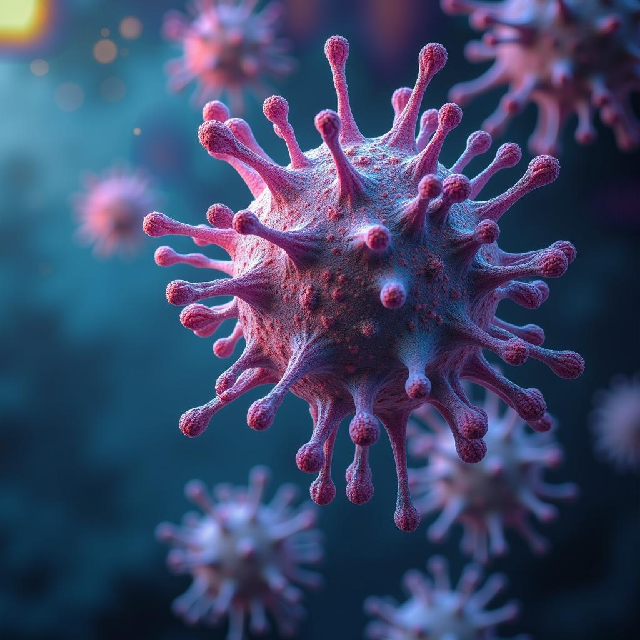}\par
    {\footnotesize\textbf{(b) \DINO{} attention}}
  \end{minipage}\hfill
  \begin{minipage}[t]{0.238\linewidth}\vspace{0pt}\centering
    \includegraphics[width=\linewidth]{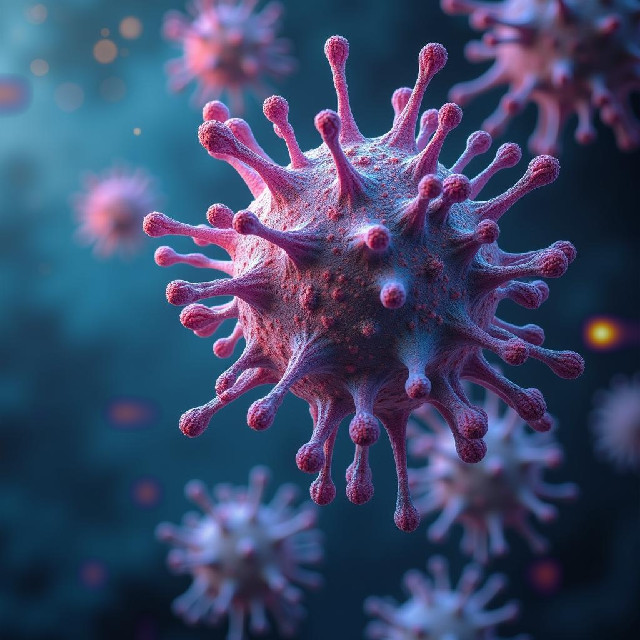}\par
    {\footnotesize\textbf{(c) \SIGLIP{} attention}}
  \end{minipage}\hfill
  \begin{minipage}[t]{0.238\linewidth}\vspace{0pt}\centering
    \includegraphics[width=\linewidth]{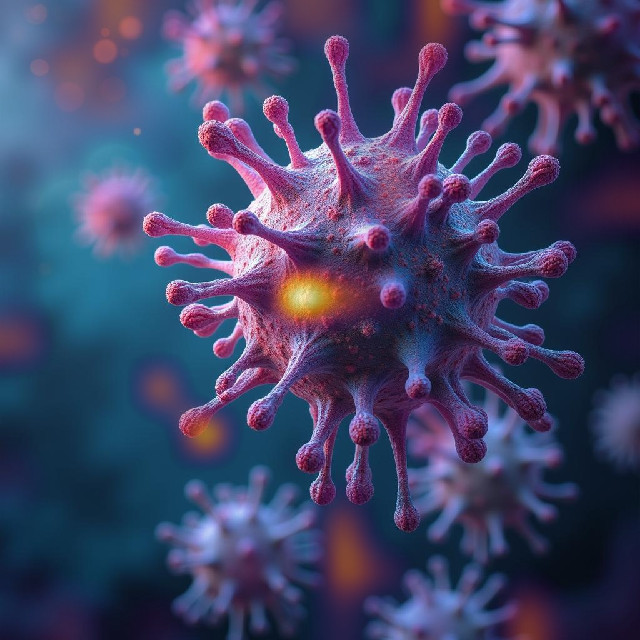}\par
    {\footnotesize\textbf{(d) \modelname{} attention}}
  \end{minipage}
  \end{minipage}\par\nopagebreak[4]
\begin{paiqcaseoutput}{PAIQCaseBaselineBG}
\textbf{\DINO{}-only output.}\enspace A vividly detailed, \PAIQcasegood{high-contrast illustration} of \PAIQcasebad{a snowman} stands prominently against a dark, starry night sky, creating a striking visual contrast that emphasizes the snowman's bright white form. The snowman is composed of \PAIQcasebad{three distinct sections}: a round head adorned with \PAIQcasebad{a cheerful red hat and a black scarf}, a square body with a black belt, and a round base with a red scarf. Each section is meticulously rendered with smooth, rounded edges and subtle shading to convey depth and texture. The background features a deep blue sky dotted with \PAIQcasebad{twinkling stars}, enhancing the wintry atmosphere. The composition is centered, drawing the viewer's attention\ldots{}
\end{paiqcaseoutput}
\begin{paiqcaseoutput}{PAIQCaseBaselineBG}
\textbf{\SIGLIP{}-only output.}\enspace Vibrant and meticulously crafted, \PAIQcasebad{a single red rose} stands prominently against a \PAIQcasegood{soft, blurred background}, its petals radiating with a rich, deep crimson hue that contrasts beautifully with the \PAIQcasebad{delicate white stamen and green stem}. The rose is positioned slightly off-center, drawing the viewer's eye to its intricate details, including the \PAIQcasebad{tightly curled petals} and the slender, vibrant green stem extending downward. The composition is simple yet striking, with the rose's symmetrical arrangement and the subtle play of light and shadow enhancing its lifelike appearance. The background, rendered in soft, muted tones, ensures the rose remains the focal point, emphasizing its natural beauty\ldots{}
\end{paiqcaseoutput}
\begin{paiqcaseoutput}{PAIQCaseMethodBG}
\textbf{\modelname{} output.}\enspace Vibrant and vividly detailed, \PAIQcasegood{a 3D-rendered virus dominates the frame}, \PAIQcasegood{its spiky, bulbous structure} radiating with a menacing yet intricate design. The virus, \PAIQcasegood{rendered in shades of pink and purple}, features \PAIQcasegood{numerous protruding spikes and rounded lobes}, creating a sense of dynamic movement and complexity. Its surface is textured with a mix of smooth and rough areas, enhancing its three-dimensional appearance. \PAIQcasegood{The background is a deep, dark blue}, which contrasts sharply with the virus's bright colors, making it stand out prominently. Scattered throughout the image are \PAIQcasegood{smaller, similarly styled viruses}, some partially obscured by the main subject\ldots{}
\end{paiqcaseoutput}

\clearpage

\subsection{Case 4: Caption 929---Navy zipper with a gold-colored pull}
\label{app:case-caption-929}

\modelname{} identifies the zipper, fabric, and gold-colored pull, while the single-encoder outputs describe an e-bike and a comb. It nevertheless describes the zipper as partly open and gives an unsupported direction for the diagonal track.

  \noindent\begin{minipage}{\linewidth}
  \begin{minipage}[t]{0.238\linewidth}\vspace{0pt}\centering
    \includegraphics[width=\linewidth]{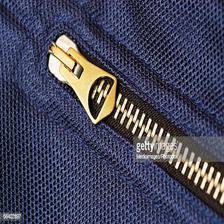}\par
    {\footnotesize\textbf{(a) Input}}
  \end{minipage}\hfill
  \begin{minipage}[t]{0.238\linewidth}\vspace{0pt}\centering
    \includegraphics[width=\linewidth]{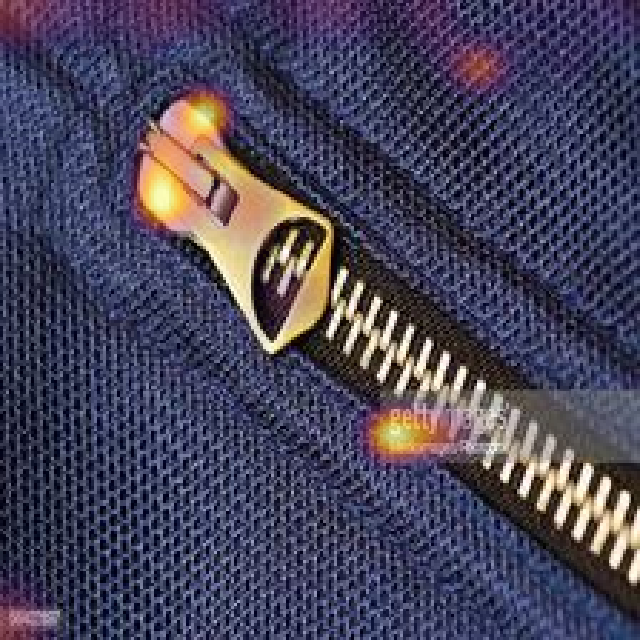}\par
    {\footnotesize\textbf{(b) \DINO{} attention}}
  \end{minipage}\hfill
  \begin{minipage}[t]{0.238\linewidth}\vspace{0pt}\centering
    \includegraphics[width=\linewidth]{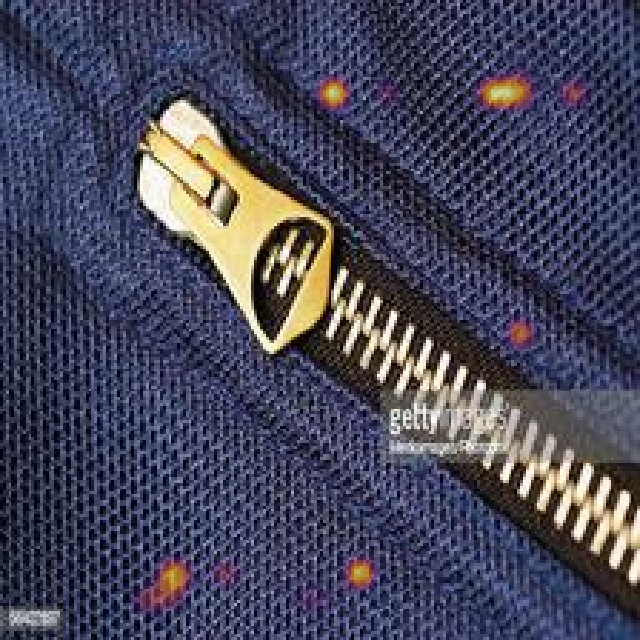}\par
    {\footnotesize\textbf{(c) \SIGLIP{} attention}}
  \end{minipage}\hfill
  \begin{minipage}[t]{0.238\linewidth}\vspace{0pt}\centering
    \includegraphics[width=\linewidth]{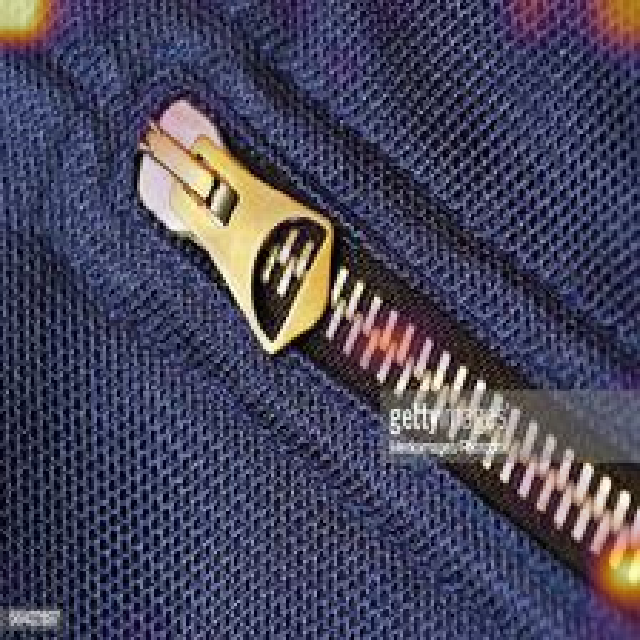}\par
    {\footnotesize\textbf{(d) \modelname{} attention}}
  \end{minipage}
  \end{minipage}\par\nopagebreak[4]
\begin{paiqcaseoutput}{PAIQCaseBaselineBG}
\textbf{\DINO{}-only output.}\enspace The image displays a sleek, modern \PAIQcasebad{electric bicycle (e-bike)} with a minimalist design, featuring \PAIQcasebad{a black frame and a compact, integrated front suspension fork}. The bike is equipped with a large, \PAIQcasebad{high-capacity battery pack} mounted on the rear rack, emphasizing its range and efficiency. The \PAIQcasebad{handlebars} are equipped with a digital display, likely showing speed, battery level, and other metrics, while the \PAIQcasebad{front wheel} is fitted with a durable tire suitable for varied terrains. The overall aesthetic is clean and functional, highlighting the bike's technological advancements and eco-friendly appeal.
\end{paiqcaseoutput}
\begin{paiqcaseoutput}{PAIQCaseBaselineBG}
\textbf{\SIGLIP{}-only output.}\enspace The image displays a meticulously crafted \PAIQcasebad{wooden comb}, featuring a \PAIQcasebad{natural wood grain texture} that adds a rustic and organic aesthetic. The comb has a \PAIQcasebad{rectangular shape with rounded edges}, and its teeth are evenly spaced, creating a symmetrical design. The teeth appear to be made of a lighter-colored material, possibly bone or plastic, contrasting subtly against the darker wood. \PAIQcasebad{The handle of the comb} is smooth and polished, suggesting a high-quality finish. The overall composition is simple yet elegant, emphasizing the craftsmanship and natural beauty of the object. The background is neutral, ensuring that the focus remains entirely on the comb.
\end{paiqcaseoutput}
\begin{paiqcaseoutput}{PAIQCaseMethodBG}
\textbf{\modelname{} output.}\enspace \PAIQcasegood{Close-up view of a navy blue fabric zipper}, featuring \PAIQcasegood{a metallic gold pull tab} with a sleek, modern design. \PAIQcasegood{The zipper teeth are evenly spaced} and appear to be made of a durable material, likely metal or plastic, reflecting light subtly. The zipper is \PAIQcasebad{partially open}, revealing \PAIQcasegood{the underlying texture of the fabric, which has a fine, woven pattern}. \PAIQcasegood{The zipper track runs diagonally across the frame}, leading the eye \PAIQcasebad{toward the top right corner}. The lighting highlights \PAIQcasegood{the contrast between the deep blue fabric and the shiny gold pull tab}, emphasizing the quality and craftsmanship of the zipper. The image is sharp and focused, capturing the intricate details of the\ldots{}
\end{paiqcaseoutput}

\clearpage

\subsection{Case 5: Caption 1083---Climbing wall with two people}
\label{app:case-caption-1083}

\modelname{} recovers the climbing scene instead of a globe or ball game. It also describes the person below as climbing, although the image only clearly shows the upper person on the wall. Correct scene recognition does not make every action claim supported.

  \noindent\begin{minipage}{\linewidth}
  \begin{minipage}[t]{0.238\linewidth}\vspace{0pt}\centering
    \includegraphics[width=\linewidth]{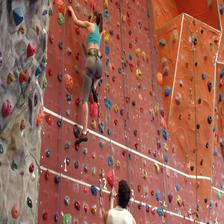}\par
    {\footnotesize\textbf{(a) Input}}
  \end{minipage}\hfill
  \begin{minipage}[t]{0.238\linewidth}\vspace{0pt}\centering
    \includegraphics[width=\linewidth]{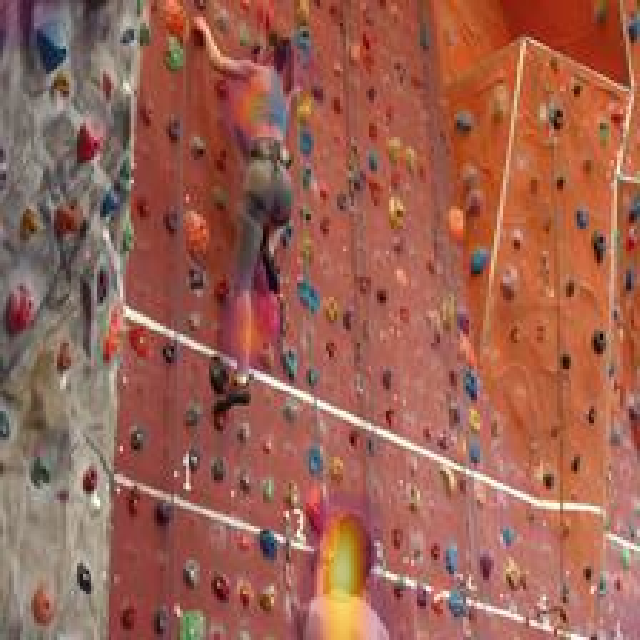}\par
    {\footnotesize\textbf{(b) \DINO{} attention}}
  \end{minipage}\hfill
  \begin{minipage}[t]{0.238\linewidth}\vspace{0pt}\centering
    \includegraphics[width=\linewidth]{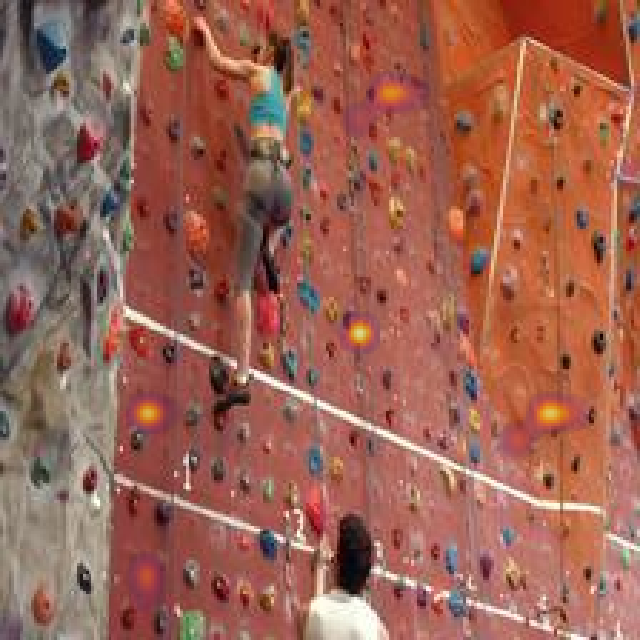}\par
    {\footnotesize\textbf{(c) \SIGLIP{} attention}}
  \end{minipage}\hfill
  \begin{minipage}[t]{0.238\linewidth}\vspace{0pt}\centering
    \includegraphics[width=\linewidth]{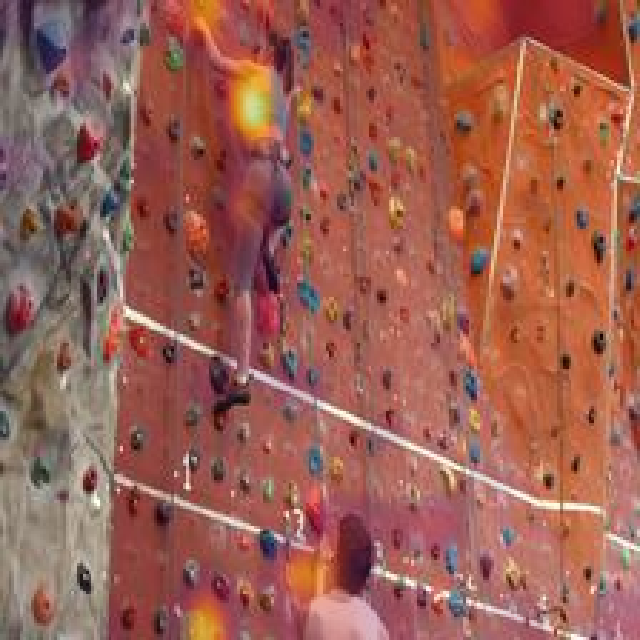}\par
    {\footnotesize\textbf{(d) \modelname{} attention}}
  \end{minipage}
  \end{minipage}\par\nopagebreak[4]
\begin{paiqcaseoutput}{PAIQCaseBaselineBG}
\textbf{\DINO{}-only output.}\enspace The image displays a close-up of a \PAIQcasebad{person's hands holding a small, round object} that appears to be \PAIQcasebad{a miniature globe or a model of the Earth}. The hands are positioned in the foreground, with \PAIQcasebad{fingers gently grasping the object}, suggesting a sense of care or curiosity. The background is blurred, drawing attention to the hands and the object they are holding. The lighting is soft, highlighting the texture of the skin and the reflective surface of the globe, which may show continents or oceans. The overall composition conveys a sense of exploration, discovery, or appreciation for the world, emphasizing the tactile connection between the human hand and the miniature representation\ldots{}
\end{paiqcaseoutput}
\begin{paiqcaseoutput}{PAIQCaseBaselineBG}
\textbf{\SIGLIP{}-only output.}\enspace In a vibrant and bustling arena, a group of enthusiastic participants clad in \PAIQcasebad{matching red and white uniforms} gather for a lively \PAIQcasebad{game of "Catch the Ball,"} a dynamic and energetic activity designed to foster teamwork and fun. The scene is set under a bright blue sky, with a large, open-air structure providing ample space for the game. Participants are actively engaged, \PAIQcasebad{some holding red balls}, while others are ready to catch them, creating a sense of motion and excitement. The atmosphere is charged with energy, as players move swiftly across the field, their movements synchronized and purposeful. The background reveals \PAIQcasebad{a crowd of spectators}, adding to the communal and festive\ldots{}
\end{paiqcaseoutput}
\begin{paiqcaseoutput}{PAIQCaseMethodBG}
\textbf{\modelname{} output.}\enspace \PAIQcasegood{Two individuals} are engaged in a dynamic activity on \PAIQcasegood{a vibrant, multi-colored rock climbing wall}. The person in the foreground, wearing a blue top and grey shorts, \PAIQcasegood{is actively climbing}, with their body positioned mid-stride, suggesting movement and effort. Below them, \PAIQcasebad{another climber in a white shirt is also scaling the wall}, adding to the sense of shared activity and community. The wall itself is a striking feature, composed of numerous colorful rocks in shades of red, orange, yellow, green, and blue, arranged in a structured pattern that guides the climbers. The scene is well-lit, highlighting the vivid colors and the physical engagement of\ldots{}
\end{paiqcaseoutput}

\clearpage

\subsection{Case 6: Caption 1242---Kitchen and open shelving}
\label{app:case-caption-1242}

Both \DINO{} and \modelname{} recognize a kitchen; \SIGLIP{} focuses on the shelving. \modelname{} mentions the faucet and cabinetry, but also calls the light countertop black and adds unsupported details. The attention overlays do not establish that these objects have been localized by the model.

  \noindent\begin{minipage}{\linewidth}
  \begin{minipage}[t]{0.238\linewidth}\vspace{0pt}\centering
    \includegraphics[width=\linewidth]{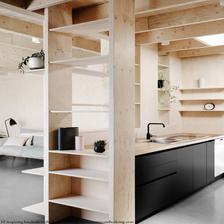}\par
    {\footnotesize\textbf{(a) Input}}
  \end{minipage}\hfill
  \begin{minipage}[t]{0.238\linewidth}\vspace{0pt}\centering
    \includegraphics[width=\linewidth]{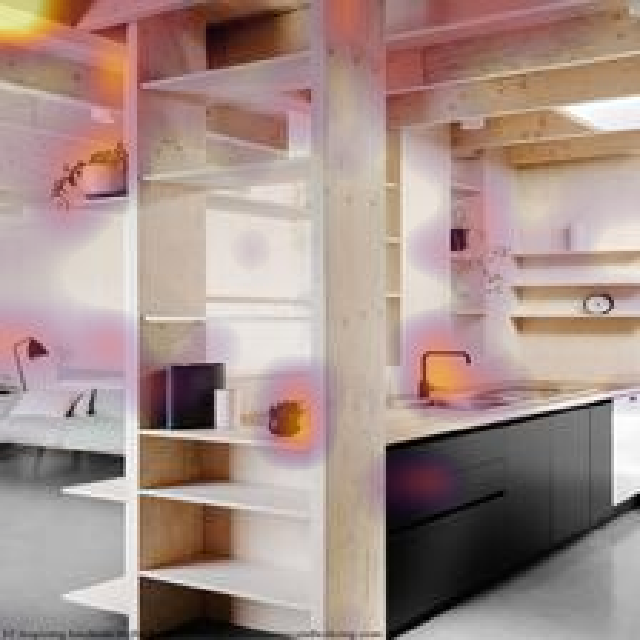}\par
    {\footnotesize\textbf{(b) \DINO{} attention}}
  \end{minipage}\hfill
  \begin{minipage}[t]{0.238\linewidth}\vspace{0pt}\centering
    \includegraphics[width=\linewidth]{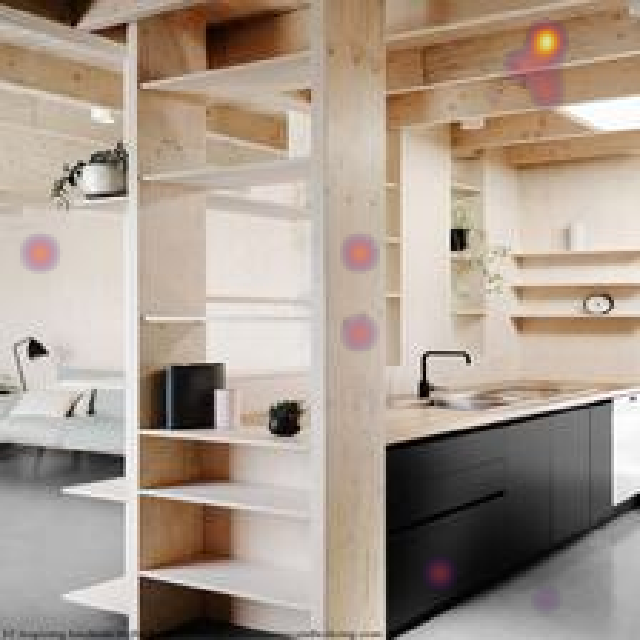}\par
    {\footnotesize\textbf{(c) \SIGLIP{} attention}}
  \end{minipage}\hfill
  \begin{minipage}[t]{0.238\linewidth}\vspace{0pt}\centering
    \includegraphics[width=\linewidth]{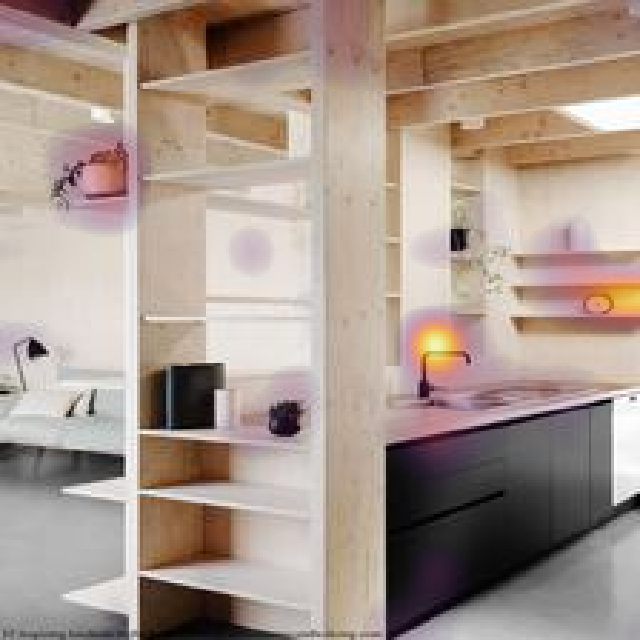}\par
    {\footnotesize\textbf{(d) \modelname{} attention}}
  \end{minipage}
  \end{minipage}\par\nopagebreak[4]
\begin{paiqcaseoutput}{PAIQCaseBaselineBG}
\textbf{\DINO{}-only output.}\enspace Nestled within a \PAIQcasegood{rustic, wood-paneled interior}, this compact living space exudes a cozy and functional aesthetic. The focal point is a sleek, modern \PAIQcasebad{kitchen island} with a dark countertop, complemented by \PAIQcasebad{a built-in oven and microwave}, seamlessly integrated into the cabinetry. Above the counter, \PAIQcasebad{a large window} allows natural light to flood the room, enhancing the warm, earthy tones of the wood and stone. To the left, \PAIQcasegood{a small potted plant} adds a touch of greenery, while \PAIQcasebad{a wooden chair sits invitingly near the window}, suggesting a space designed for both relaxation and practical use. The overall design emphasizes simplicity\ldots{}
\end{paiqcaseoutput}
\begin{paiqcaseoutput}{PAIQCaseBaselineBG}
\textbf{\SIGLIP{}-only output.}\enspace Minimalist \PAIQcasegood{wooden shelving unit} with a sleek, modern design, featuring three tiers of open shelves and \PAIQcasebad{a lower closed compartment}, all crafted from \PAIQcasegood{light-toned wood with visible grain}. The unit is positioned against a neutral wall, complemented by \PAIQcasebad{a large window on the right} that allows natural light to flood the space, enhancing the airy and open feel. \PAIQcasegood{A small potted plant} sits atop the top shelf, adding a touch of greenery, while a few books and decorative items are neatly arranged on the lower shelves, showcasing a blend of functionality and aesthetic appeal. The overall composition emphasizes simplicity and elegance, ideal for a contemporary interior setting.
\end{paiqcaseoutput}
\begin{paiqcaseoutput}{PAIQCaseMethodBG}
\textbf{\modelname{} output.}\enspace Compact and minimalist design, \PAIQcasegood{this modern kitchen} exemplifies the essence of Scandinavian architecture, characterized by its clean lines, natural wood accents, and functional layout. The space features a sleek \PAIQcasebad{black countertop} paired with a \PAIQcasegood{dark cabinetry unit}, creating a striking contrast against the \PAIQcasegood{light wooden beams that form the ceiling and walls}. \PAIQcasegood{Open shelving units, constructed from light wood}, are strategically placed to maximize storage while maintaining an airy feel, with decorative elements such as \PAIQcasegood{potted plants} and \PAIQcasebad{small vases} adding a touch of warmth and personality. \PAIQcasebad{A black sink} with \PAIQcasegood{a modern faucet} sits atop the counter, complemented by a small \PAIQcasebad{wall-mounted} clock and \PAIQcasebad{a framed picture}\ldots{}
\end{paiqcaseoutput}

\endgroup